# Sharp Computational-Statistical Phase Transitions via Oracle Computational Model

Zhaoran Wang*  Quanquan Gu[†]  Han Liu*

**Abstract**

We study the fundamental tradeoffs between computational tractability and statistical accuracy for a general family of hypothesis testing problems with combinatorial structures. Based upon an oracle model of computation, which captures the interactions between algorithms and data, we establish a general lower bound that explicitly connects the minimum testing risk under computational budget constraints with the intrinsic probabilistic and combinatorial structures of statistical problems. This lower bound mirrors the classical statistical lower bound by Le Cam (1986) and allows us to quantify the optimal statistical performance achievable given limited computational budgets in a systematic fashion. Under this unified framework, we sharply characterize the statistical-computational phase transition for two testing problems, namely, normal mean detection and sparse principal component detection. For normal mean detection, we consider two combinatorial structures, namely, sparse set and perfect matching. For these problems we identify significant gaps between the optimal statistical accuracy that is achievable under computational tractability constraints and the classical statistical lower bounds. Compared with existing works on computational lower bounds for statistical problems, which consider general polynomial-time algorithms on Turing machines, and rely on computational hardness hypotheses on problems like planted clique detection, we focus on the oracle computational model, which covers a broad range of popular algorithms, and do not rely on unproven hypotheses. Moreover, our result provides an intuitive and concrete interpretation for the intrinsic computational intractability of high-dimensional statistical problems. One byproduct of our result is a lower bound for a strict generalization of the matrix permanent problem, which is of independent interest.

## 1 Introduction

Statistical inference on massive datasets with high dimensionality and large sample size gives rise to the following key question: What is the optimal statistical performance we can achieve under limited computational budgets? In this paper, we aim to establish a better understanding of this question by studying the problem of combinatorial structure detection. In detail, let $\boldsymbol{X} \in \mathbb{R}^d$ be a random vector, $\mathbb{P}_{\mathcal{S}}$ be the distribution which corresponds to $\mathcal{S} \subseteq [d]$, and $\mathcal{C}$ be the family of all possible index sets $\mathcal{S}$ that are of interest. Given $n$ independent realizations of $\boldsymbol{X}$, we study the general hypothesis testing problem, where the alternative hypothesis is that there exists an $\mathcal{S} \in \mathcal{C}$ such that $\boldsymbol{X} \sim \mathbb{P}_{\mathcal{S}}$, while the null hypothesis is that there exits no such $\mathcal{S}$, which is denoted by $\boldsymbol{X} \sim \mathbb{P}_0$. This testing problem is ubiquitous in high-dimensional statistics. As will be illustrated in §2, it covers, among others, sparse normal mean detection and sparse principal component detection as special cases. See, e.g., Addario-Berry et al. (2010); Verzelen and Arias-Castro (2014); Arias-Castro et al. (2015a); Donoho and Jin (2015) for more examples. Besides, this testing problem fully captures the difficulty of the associated estimation and support recovery problems, in the sense that, if we can solve the latter two problems, we can use the estimator or recovered support to construct a test that solves the testing problem.

---
*Department of Operations Research and Financial Engineering, Princeton University, Princeton, NJ 08544, USA; Email: {zhaoran, hanliu}@princeton.edu

[†]Department of Systems and Information Engineering, University of Virginia, Charlottesville, VA 22904, USA; e-mail: qg5w@virginia.edu

Following the pioneer works by Berthet and Rigollet (2013a,b), a recent line of research quantifies the optimal statistical performance achievable using polynomial-time algorithms (Ma and Wu, 2014; Arias-Castro and Verzelen, 2014; Zhang et al., 2014; Hajek et al., 2014; Gao et al., 2014; Chen and Xu, 2014; Wang et al., 2014a; Chen, 2015; Krauthgamer et al., 2015; Cai et al., 2015). These works are mostly built upon a computational hardness hypothesis for planted clique detection. Respectively, their proofs are based upon randomized polynomial-time reductions from the planted clique detection problem to statistical problems. This reduction-based approach has several drawbacks.

- Most of existing computational hardness hypotheses are on the worst-case complexity. Apart from the planted clique hypothesis, there are very few average-case complexity hypotheses. Nevertheless, statistical problems in general involve probability distributions on the input of algorithms, and are naturally associated with average-case complexity. Furthermore, even if we have more average-case complexity hypotheses, we lack a systematic method to connect a statistical problem with a proper average-case complexity hypothesis.

- For certain statistical problems, even if we manage to find a corresponding hypothesis on average-case complexity, it is possible that there lacks consensus on the correctness of the hypothesis. One such example is the Random 3SAT hypothesis (Feige, 2002). In other words, relying on average-case complexity hypotheses can sometimes be risky.

- Most hypotheses on computational hardness assert whether a problem can be solved in polynomial time, which is prohibiting for quantifying computational complexity in a more fine-grained fashion. For example, we may be interested in whether a problem can be solved in $O(p^2)$ time, where $p$ is the input size. For massive datasets, this fine-grained characterization of computational complexity is crucial. For example, when $p$ is large, even $O(p^3)$ time can be a too expensive overhead that is undesirable in large-scale settings.

In this paper, we take a different path from the previous reduction-based approach. Rather than relating the statistical problem to a problem that is conjectured to be hard to compute, we establish a direct connection between the optimal statistical performance achievable with limited computational budgets and the intrinsic probabilistic and combinatorial structures of statistical problems. In detail, without conditioning on any unproven hypothesis, we establish a tight lower bound on the risk of any test that is obtained under computational budget constraints. As will be specified in a moment, this lower bound has an explicit dependency on the structure class $\mathcal{C}$, null distribution $\mathbb{P}_0$, and alternative distribution $\mathbb{P}_{\mathcal{S}}$. This lower bound under computational constraints mirrors the classical lower bound for testing two hypotheses (Le Cam, 1986), where the minimum testing risk explicitly depends on the distance between the null and alternative distributions. In particular, for the aforementioned testing problem, the classical lower bound depends on the total variation distance between $\mathbb{P}_0$ and a certain mixture of $\{\mathbb{P}_\mathcal{S} : \mathcal{S} \in \mathcal{C}\}$. In contrast, as will be shown in a moment, our lower bound depends on the difference between $\mathbb{P}_0$ and the elements of $\{\mathbb{P}_\mathcal{S} : \mathcal{S} \in \bar{\mathcal{C}}\}$, where $\bar{\mathcal{C}}$ is a subset of $\mathcal{C}$. That is to say, our lower bound can be viewed as an extension of Le Cam's lower bound, which quantifies the localized difference between null and alternative distributions.

Our general lower bound on testing risk is obtained under a computational model named oracle model, which is also known as the statistical query model (Kearns, 1993). The high-level intuition for this model is as follows. To solve statistical problems, algorithms need to interact with data. Hence, the total number of rounds of interactions with data is a good proxy for quantifying the algorithmic complexity of statistical problems. Based on this intuition, the oracle model is defined by specifying a protocol on the way algorithms interact with data. In detail, at each round, the algorithm sends a query function $q : \mathcal{X} \to \mathbb{R}$ to an oracle, where $\mathcal{X}$ is the domain of the random vector $\boldsymbol{X}$; The oracle responds the algorithm with a random variable concentrated around $\mathbb{E}[q(\boldsymbol{X})]$. See Definition 3.1 for a formal definition. Despite its restriction on the way algorithms interact with data, the oracle model captures a broad range of popular algorithms for statistical problems, including convex optimization methods for $M$-estimation such as first-order methods and coordinate descent, matrix decomposition algorithms for principal component analysis or factor models such as power method and QR method, expectation-maximization algorithms for latent variable model estimation, and sampling algorithms like Markov chain Monte Carlo. See §3 for a detailed discussion on the generality of the oracle model.

We are motivated to study the oracle computational model on statistical problems by the success of the black-box model in convex optimization (Nemirovski and Yudin, 1983). In detail, the black-box model allows an optimization algorithm, which targets at minimizing an unknown objective function $f : \mathcal{X} \to \mathbb{R}$, to interact with the zeroth-order or first-order oracle, which provides the objective value $f(\mathbf{x})$ or subgradient $\partial f(\mathbf{x})$ at $\mathbf{x} \in \mathcal{X}$. The black-box model restricts the way optimization algorithms access the objective function, while capturing a broad family of algorithms (Nesterov, 2004; Bubeck, 2015). By restricting the computational model without losing generality, the black-box model enables a fine-grained characterization of the intrinsic iteration complexity of optimization problems without relying on reductions to computational hardness hypotheses. Following the same thinking, we resort to the oracle model to bypass the unproven computational hardness hypotheses. Instead of accessing the objective function via objective values or subgradients as in the black-box model of optimization, here algorithms interact with data using any query function to solve a statistical problem. Following the convention in convex optimization literature, we name the total number of rounds of interactions with the oracle as the oracle complexity.

Under the aforementioned oracle computational model, our lower bound on the testing risk takes the following form. For any test $\phi$ obtained by an algorithm, which interacts with an oracle no more than $T$ rounds, roughly speaking, we have

$$\overline{R}(\phi) \geq 1 - T \cdot \sup_{q \in \mathcal{Q}} |\mathcal{C}(q)|/|\mathcal{C}|. \tag{1.1}$$

Here $\overline{R}$ is risk measure under the uniform prior over the alternative hypotheses, which will be defined in (4.2), while $\mathcal{C}$ is the aforementioned structure class, and $|\mathcal{C}|$ is its cardinality. For a query function $q$, $\mathcal{C}(q)$ is the class of $\mathcal{S}$'s for which $\mathbb{P}_{\mathcal{S}}$ can be distinguished from $\mathbb{P}_0$ by $q$, i.e., the difference between $\mathbb{E}_{\boldsymbol{X} \sim \mathbb{P}_{\mathcal{S}}}[q(\boldsymbol{X})]$ and $\mathbb{E}_{\boldsymbol{X} \sim \mathbb{P}_0}[q(\boldsymbol{X})]$ is sufficiently large for $\mathcal{S} \in \mathcal{C}(q)$. In Definition 4.1, we will formally define $\mathcal{C}(q)$. Meanwhile, $\mathcal{Q}$ is the set of all possible query functions. This main result will be formally presented in Theorem 4.2. The intuition behind (1.1) is as follows. At each round of interaction with the oracle, the algorithm can use the collected information to distinguish between $\mathbb{P}_0$ and a subset of $\{\mathbb{P}_{\mathcal{S}} : \mathcal{S} \in \mathcal{C}\}$. Using a single query $q$, intuitively the algorithm can separate $\mathbb{P}_0$ with $\{\mathbb{P}_{\mathcal{S}} : \mathcal{S} \in \mathcal{C}(q)\}$. Thus, using $T$ queries, the algorithm can separate $\mathbb{P}_0$ with a subset of $\{\mathbb{P}_{\mathcal{S}} : \mathcal{S} \in \mathcal{C}\}$ with cardinality at most $T \cdot \sup_{q \in \mathcal{Q}} |\mathcal{C}(q)|$. Respectively, within $\{\mathbb{P}_{\mathcal{S}} : \mathcal{S} \in \mathcal{C}\}$, there exist at least $|\mathcal{C}| - T \cdot \sup_{q \in \mathcal{Q}} |\mathcal{C}(q)|$ elements that can not be distinguished from $\mathbb{P}_0$ by the algorithm, which corresponds to the minimum testing risk on the right-hand side of (1.1). This general result enables us to characterize the optimal statistical performance achievable under constraints on computational budgets in a more systematic manner. In detail, since it is easy to calculate $|\mathcal{C}|$ for specific testing problems, it remains to establish the upper bound of $\sup_{q \in \mathcal{Q}} |\mathcal{C}(q)|$. As will be shown in §4.2, under certain regularity conditions, which are satisfied by a broad range of testing problems, the upper bound of $\sup_{q \in \mathcal{Q}} |\mathcal{C}(q)|$ is connected to the following quantity,

$$\mathbb{E}_{\mathcal{S}' \sim \mathbb{Q}_{\bar{\mathcal{C}}}}[h(|\mathcal{S} \cap \mathcal{S}'|)], \quad \text{where} \quad |\bar{\mathcal{C}}| = \sup_{q \in \mathcal{Q}} |\mathcal{C}(q)|. \tag{1.2}$$

Here $\mathbb{Q}_{\bar{\mathcal{C}}}$ is the uniform distribution over $\bar{\mathcal{C}} \subseteq \mathcal{C}$, $h$ is a nondecreasing function, which only depends on the null and alternative distributions, and $\mathcal{S} \in \mathcal{C}$. In Theorem 4.4 we will prove that, for a particular $\bar{\mathcal{C}}$, (1.2) does not depend on $\mathcal{S}$ and is nonincreasing in $|\bar{\mathcal{C}}| = \sup_{q \in \mathcal{Q}} |\mathcal{C}(q)|$. Using information-theoretical techniques, we establish a lower bound of (1.2), which implies an upper bound of $\sup_{q \in \mathcal{Q}} |\mathcal{C}(q)|$. Note that for a specific statistical model, which corresponds to a particular $h$ function, (1.2) is a quantity that only depends on the combinatorial property of $\mathcal{C}$, i.e., the overlaps of $\mathcal{S}$ and $\mathcal{S}'$ with $\mathcal{S}'$ uniformly drawn from a subset of $\mathcal{C}$. In fact, Addario-Berry et al. (2010) show that this combinatorial property of $\mathcal{C}$ is crucial to the information-theoretical lower bounds for a broad range of testing problems that involve combinatorial structures. Here our result shows that it also plays a key role in computational lower bounds. The above unified framework can be viewed from a perspective on the duality between spaces of measures and functions, which reveals its connection to the notions of Wasserstein distance, Radon distance, and total variation distance. See §4.1 for a detailed discussion and Figures 1 and 2 for an illustration of the intuition behind (1.1) and (1.2) respectively.

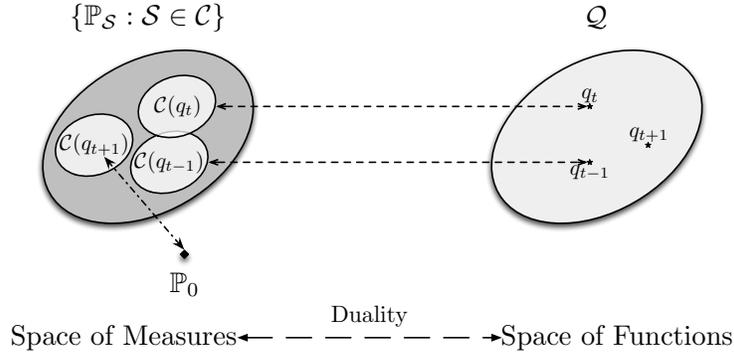

Figure 1: An illustration of the intuition behind (1.1). Here we illustrate three queries, namely, $q_{t-1}$, $q_t$, and $q_{t+1}$, among the $T$ queries used by the algorithm. Each query $q \in \mathcal{Q}$ on the right corresponds to a subset $\mathcal{C}(q)$ of $\mathcal{C}$, which denotes that $\mathbb{P}_{\mathcal{S}}$ with $\mathcal{S} \in \mathcal{C}(q)$ can be distinguished from $\mathbb{P}_0$ by query $q$. The dark area on the left consists of the elements of $\mathcal{C}$ for which $\mathbb{P}_{\mathcal{S}}$ can not be distinguished from $\mathbb{P}_0$ by any of the $T$ queries. The proportion of the dark area is lower bounded by the right-hand side of (1.1), which is the lower bound of the testing risk. On the left, $\mathbb{P}_{\mathcal{S}}$ ($\mathcal{S} \in \mathcal{C}$) and $\mathbb{P}_0$ lie in the space of probability measures, while on the right, $q$ lies in the space of functions. As we will show in §4.1, the notion of $\mathcal{C}(q)$ is connected with the pairings between the elements from the spaces of functions and measures, which are dual to each other.

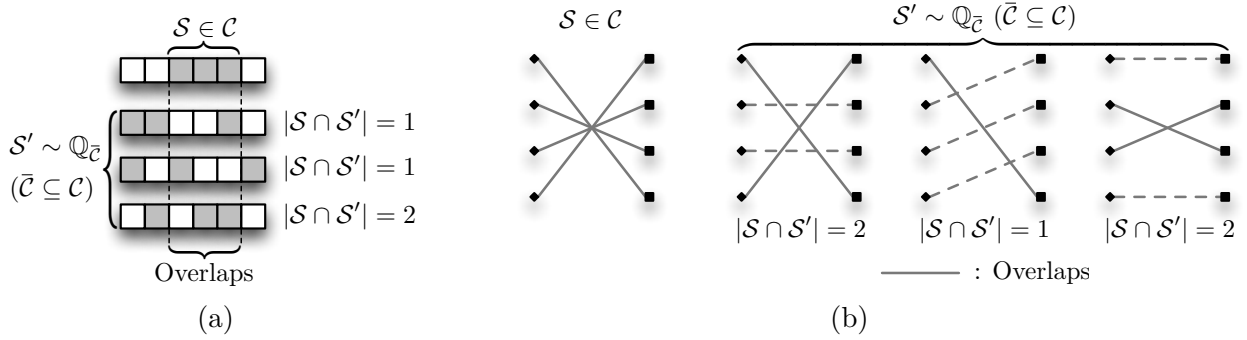

Figure 2: An illustration of the intuition behind (1.2). In (a) we illustrate the case where $\mathcal{C}$ consists of sparse sets, i.e., $\mathcal{C} = \{\mathcal{S} \subseteq [d] : |\mathcal{S}| = s^*\}$. Here we take $s^* = 3$ and $d = 6$. In (b) we illustrate the case where $\mathcal{C}$ consists of perfect matchings, which will be defined in §2. Here $\mathcal{S} \in \mathcal{C}$ is fixed and $\mathcal{S}'$ comes from the uniform distribution $\mathbb{Q}_{\bar{\mathcal{C}}}$ over $\bar{\mathcal{C}} \subseteq \mathcal{C}$ with $|\bar{\mathcal{C}}| = \sup_{q \in \mathcal{Q}} |\mathcal{C}(q)|$. The overlaps between $\mathcal{S}$ and $\mathcal{S}'$ and the $h$ function in (1.2) together determine $|\bar{\mathcal{C}}| = \sup_{q \in \mathcal{Q}} |\mathcal{C}(q)|$. The overlaps of $\mathcal{S}$ and $\mathcal{S}'$ depend on the combinatorial structure of $\mathcal{C}$, while $h$ depends on the probabilistic structure of the statistical problem. See §4.2 for more details.

Under this unified framework, we consider two statistical models, namely, shifted mean detection and sparse principal component detection. In detail, for shifted mean detection, we consider testing $H_0 : \boldsymbol{X} \sim N(\boldsymbol{0}, \mathbf{I}_d)$ against $H_1 : \boldsymbol{X} \sim (1-\alpha) N(\boldsymbol{0}, \mathbf{I}_d) + \alpha N(\boldsymbol{\theta}, \mathbf{I}_d)$, where $\text{supp}(\boldsymbol{\theta}) = \mathcal{S} \in \mathcal{C}$. We study the detection of two structures: (i) $\mathcal{C} = \{\mathcal{S} \subseteq [d] : |\mathcal{S}| = s^*\}$; (ii) $\mathcal{C}$ consists of all perfect matchings of a complete balanced bipartite graph with $2\sqrt{d}$ nodes. For sparse principal component detection, we consider testing $H_0 : N(\boldsymbol{0}, \mathbf{I}_d)$ against $H_1 : N(0, \mathbf{I}_d + \beta^* \mathbf{v}^* \mathbf{v}^{*\top})$, where $\|\mathbf{v}^*\|_2 = 1$ and $\text{supp}(\mathbf{v}^*) = \mathcal{S}$ with $\mathcal{S} \in \mathcal{C} = \{\mathcal{S} \subseteq [d] : |\mathcal{S}| = s^*\}$. See §2 for more details. These three examples cover two statistical models, which determine $h$ in (1.2), and two structure classes $\mathcal{C}$ with distinct combinatorial properties on overlapping pairs. For these examples, we sharply characterize the computational and statistical phase transition over all parameter configurations. For instance, for shifted mean detection where $\mathcal{C}$ is the class of all sparse sets, i.e., $\mathcal{C} = \{\mathcal{S} \subseteq [d] : |\mathcal{S}| = s^*\}$, we quantify the phase transition over the sparsity level $s^*$, dimension $d$, sample size $n$, minimum signal strength $\beta^* = \min_{j \in \mathcal{S}} |\theta_j|$, and mixture level $\alpha$. In detail, such a phase transition is shown in Figure 3, where we say a test is asymptotically

powerful (resp., powerless) if its risk converges to zero (resp., one), and use the following notation,

$$p_{s^*} = \log s^*/\log d, \quad p_{\beta^*} = \log(1/\beta^*)/\log d, \quad p_n = \log n/\log d, \quad p_\alpha = \log(1/\alpha)/\log d. \quad (1.3)$$

Our result captures the fundamental gap between the classical information-theoretical lower bound, which can be achieved by an algorithm with oracle complexity exponential in $d$, and the lower bound under computational tractability constraints, which can be achieved by an algorithm that has oracle complexity polynomial in $d$. As shown in Figure 4, the perfect matching detection problem exhibits a similar tradeoff between computational tractability and statistical accuracy. In addition, our result captures the same phenomenon for the sparse principal component detection problem. In particular, we recover the result of Berthet and Rigollet (2013a,b) under the oracle model without relying on the planted clique hypothesis. See §5 and §6 for a detailed description of the computational lower bounds, information-theoretical lower bounds, and corresponding upper bounds for each statistical problem. Notably, a byproduct of our result for perfect matching detection is a computational lower bound for a generalized matrix permanent problem. An implication of this lower bound is, although the Markov chain Monte Carlo algorithm can approximate the permanent of a matrix with nonnegative elements up to a arbitrarily small relative error in polynomial time (Jerrum et al., 2004), it would however fail to achieve the same accuracy in polynomial time for a strict generalization of the matrix permanent problem. See §5.2.4 for more details.

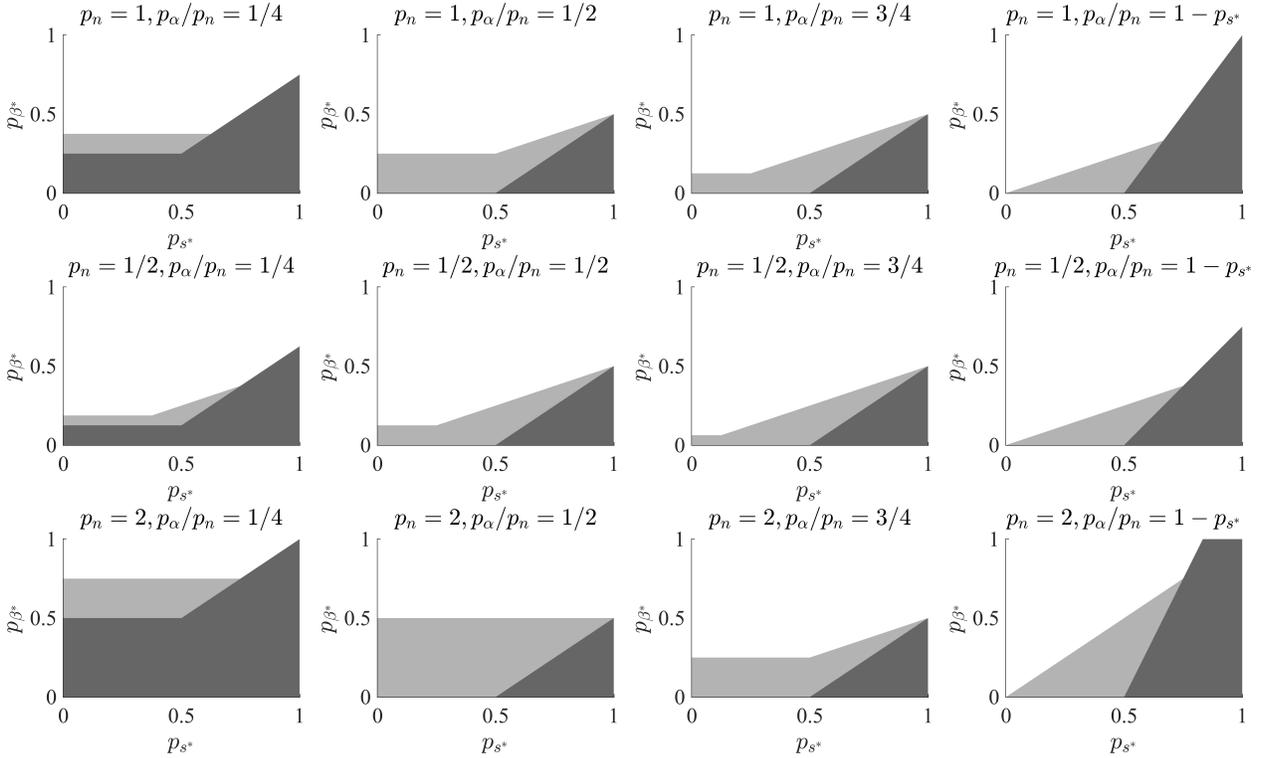

Figure 3: The computational and statistical phase transition of sparse set detection for the shifted mean model. Here $p_{s^*}$, $p_{\beta^*}$, $p_n$, and $p_\alpha$ are defined in (1.3). The blank area denotes the statistically impossible regime, where any test is asymptotically powerless. The light-colored area is the computationally intractable regime, where any test based on an algorithm with polynomial oracle complexity is asymptotically powerless. The dark-colored area is the computationally tractable regime in which there exists an asymptotically powerful test that is based upon an algorithm with polynomial oracle complexity. See §5.1.4 for a detailed discussion.

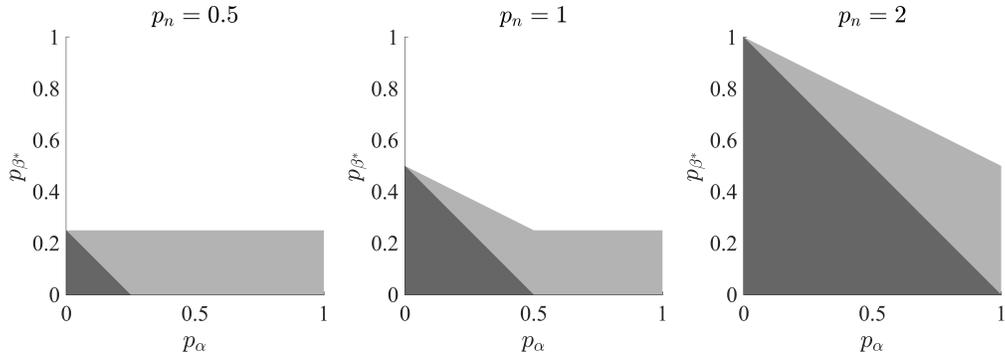

Figure 4: The computational and statistical phase transition of perfect matching detection for the shifted mean model. Here $p_{\beta^*}$, $p_n$, $p_\alpha$ are defined in (1.3). The blank area represents the statistically impossible regime, where any test is asymptotically powerless. The light-colored area is the computationally intractable regime, where any test based on an algorithm with polynomial oracle complexity is asymptotically powerless. The dark-colored area is the computationally tractable regime in which there exists an asymptotically powerful test that is based upon an algorithm with polynomial oracle complexity. See §5.2.4 for a detailed discussion.

## 1.1 Related Works

The oracle computational model studied in this paper extends the statistical query model proposed by Kearns (1993, 1998), which is subsequently studied by Blum et al. (1994, 1998); Servedio (1999); Yang (2001, 2005); Jackson (2003); Szörényi (2009); Feldman (2012); Feldman and Kanade (2012); Feldman et al. (2013, 2015). In particular, our oracle model is based on the VSTAT oracle proposed in the seminal works by Feldman et al. (2013, 2015) with the following extensions.

- In our model, the query functions are real-valued, while the VSTAT oracle by Feldman et al. (2013, 2015) only allows queries with discrete values, e.g., Boolean-valued queries. Meanwhile, statistical problems naturally involve real-valued queries. To translate the real-valued queries used in practice to Boolean values, we often have to pay extra factors in the oracle complexity due to discretization, which leads to a discrepancy between lower and upper bounds. By allowing real-valued queries, our model avoids such a discrepancy and better captures real-world algorithms.

  However, since the algorithm can access more information at each round using real-valued queries, the lower bounds on testing risk are more difficult to establish under our oracle model. To address this issue, we establish a new worst-case construction in the proof of (1.1), which differs from the one employed by Szörényi (2009); Feldman (2012); Feldman et al. (2013, 2015). The power of this construction comes from that it allows a more fine-grained characterization of $\mathcal{C}(q)$ in (1.1), which will be detailedly described in §4.2. Using this construction, we establish a stronger computational lower bound in comparison with Feldman et al. (2013, 2015).

- We explicitly incorporate the notions of query space capacity and tail probability to characterize the uniform deviation of the responses for queries. In detail, the oracle responds query $q$ with a random variable $Z_q$ that concentrates around $\mathbb{E}[q(\boldsymbol{X})]$. Roughly speaking, we require $Z_q$ to satisfy

$$\mathbb{P}\{\sup_{q \in \overline{\mathcal{Q}}} |Z_q - \mathbb{E}[q(\boldsymbol{X})]| \leq \tau\} \geq 1 - 2\xi, \tag{1.4}$$

where $\xi$ is the tail probability, $\overline{\mathcal{Q}} \subseteq \mathcal{Q}$ is query space, which depends on specific algorithms, while $\tau$ is the tolerance parameter, which depends on $\xi$ and $\overline{\mathcal{Q}}$. Apart from the aforementioned difference between real-valued and discrete-valued queries, the VSTAT oracle by Feldman et al. (2013, 2015) is a special case of (1.4) with $|\overline{\mathcal{Q}}| = 1$ and $\xi = 0$. As we will illustrate in §3, this restriction is too optimistic in the sense that it often requires more than $n$ data points to simulate the VSTAT oracle practically, which results in a contradiction with the information-theoretical lower bound. We will illustrate this phenomenon with an example in §5.1.3.

As previously discussed, our work is related to a recent line of research on the fundamental tradeoff between computational tractability and statistical accuracy (Berthet and Rigollet, 2013a,b; Ma and Wu, 2014; Zhang et al., 2014; Hajek et al., 2014; Chen and Xu, 2014; Wang et al., 2014a; Gao et al., 2014; Arias-Castro and Verzelen, 2014; Chen, 2015; Krauthgamer et al., 2015; Cai et al., 2015). This line of research uses reductions from a problem that is hypothetically intractable to compute to the statistical problem of interest. In contrast, we focus on the oracle computational model, which covers a broad range of widely used algorithms, and establish unconditional lower bounds that do not rely on any unproven hardness hypothesis. Moreover, we directly connects the algorithmic complexity of a statistical problem with its intrinsic probabilistic and combinatorial structures, which allows us to establish computational lower bounds in a more systematic fashion. Another related line of research considers the limits of convex relaxation for statistical problems (Chandrasekaran and Jordan, 2013; Deshpande and Montanari, 2015; Ma and Wigderson, 2015; Wang et al., 2015). In particular, they establish unconditional computational lower bounds for convex relaxation hierarchies. In comparison, our lower bounds cover a broad range of algorithms that fall in the oracle model.

It is worth noting that Feldman et al. (2013) establish a computational lower bound for a variant of the planted clique problem under the oracle model, and Berthet and Rigollet (2013a) establish the reduction from the planted clique problem to sparse principal component detection. One may suggest that we can combine these two results to establish the unconditional lower bound for sparse principal component detection in this paper. Nevertheless, this approach only yields a weaker lower bound for discrete-valued queries. More importantly, our goal is to illustrate how the intrinsic probabilistic and combinatorial structures of sparse principal component detection affect its algorithmic complexity by a unified approach. Such a unified approach is widely applicable to the statistical problems for which the reduction based upon planted clique is unclear or does not exist, e.g., the shifted-mean detection problem with combinatorial structures considered in this paper. More such examples include mixture of Gaussian, phase retrieval, mixed regression, detection of positive correlation and Markov random field, tensor principal component analysis, hidden Markov model, and sufficient dimension reduction, which will be covered in our upcoming followup paper based on an extension of the unified approach.

## 2 Background

In the following, we describe two hypothesis testing problems, which are special cases of the general detection problem introduced in §1.

**Shifted Mean Detection:** Recall $\boldsymbol{X}$ is a $d$-dimensional random vector. Under the null hypothesis $H_0$, $\boldsymbol{X}$ follows $N(\mathbf{0}, \mathbf{I}_d)$, while under the alternative hypothesis $H_1$, $\boldsymbol{X}$ follows the mixture distribution $(1-\alpha)N(\mathbf{0}, \mathbf{I}_d) + \alpha N(\boldsymbol{\theta}, \mathbf{I}_d)$, where $\alpha \in (0,1]$ and $\boldsymbol{\theta}$ is specified as follows. Let $\mathcal{C} = \{\mathcal{S}_1, \ldots, \mathcal{S}_m\}$ be the family of index sets in which $\mathcal{S}_\ell \subseteq [d]$ and $|\mathcal{S}_\ell| = s^*$ for any $\ell \in [m]$. We assume that there exists an index set $\mathcal{S} \in \mathcal{C}$ such that $\theta_j = \beta^*$ for $j \in \mathcal{S}$, while $\theta_j = 0$ for $j \notin \mathcal{S}$. Let $n$ be the total number of observations of $\boldsymbol{X}$. This hypothesis testing problem encompasses several examples.

  (i) For $\alpha = 1$ and $\mathcal{C} = \{\mathcal{S} \subseteq [d] : |\mathcal{S}| = s^*\}$, it is the detection of sparse normal mean problem. See, e.g., Donoho and Jin (2015) and the references therein. Moreover, for general $\alpha$, this problem is studied by Cai and Wu (2014) among others.

 (ii) For $\alpha = s^*/n$ and $\mathcal{C} = \{\mathcal{S} \subseteq [d] : |\mathcal{S}| = s^*\}$, it is a variant of submatrix detection problem. To see this, note that in expectation there are $s^*$ rows with shifted mean in the $n \times d$ data matrix. In other words, in expectation there is an $s^* \times s^*$ hidden submatrix in the data matrix. If $n$ is of the same order as $d$, it becomes a variant of the problem considered by Butucea and Ingster (2013); Ma and Wu (2014) among others.

(iii) For $\alpha = 1$ and $\mathcal{C}$ being the set of all perfect matchings in a complete balanced bipartite graph with $2\sqrt{d}$ nodes, it is known as the perfect matching detection problem (Addario-Berry et al., 2010). Here each entry of $\boldsymbol{X}$ represents an edge in the bipartite graph and we assume that $\sqrt{d}$ is an integer without any loss of generality. Particularly, for $j = (k-1)\sqrt{d} + k'$, $X_j$ corresponds to the edge between nodes $k$ and $k'$ on each partition of the graph. As we will show later, the perfect matching detection problem is connected with the matrix permanent problem through the likelihood ratio test (Addario-Berry et al., 2010).

See Addario-Berry et al. (2010) for more hypothesis testing problems with combinatorial structures. Note that $\alpha n$ is the expected number of data points that come from $N(\boldsymbol{\theta}, \mathbf{I}_d)$ under $H_1$. Throughout this paper, we focus on the setting with $\alpha n \to \infty$ and $\beta^* = o(1)$ for shifted mean detection.

**Principal Component Detection:** Let $\boldsymbol{X} \in \mathbb{R}^d$ be a Gaussian random vector. For the detection of sparse principal component, the null hypothesis $H_0$ is $\boldsymbol{X} \sim N(0, \mathbf{I}_d)$, and the alternative hypothesis $H_1$ is $\boldsymbol{X} \sim N(0, \mathbf{I}_d + \beta^* \mathbf{v}^* \mathbf{v}^{*\top})$ in which $\mathbf{v}^* \in \mathcal{B}_0(s^*)$. Here $\mathcal{B}_0(s^*) = \{\mathbf{v} \in \mathbb{R}^d : \|\mathbf{v}\|_2 = 1, \|\mathbf{v}\|_0 \leq s^*\}$. The model under $H_1$ is known as the spiked covariance model. See, e.g., Berthet and Rigollet (2013b) for details. Let $\mathrm{supp}(\mathbf{v}^*) = \mathcal{S} \in \mathcal{C} = \{\mathcal{S} \subseteq [d] : |\mathcal{S}| = s^*\}$. We assume $v_j^* = 1/\sqrt{s^*}$ if $j \in \mathcal{S}$ and $v_j^* = 0$ otherwise. Throughout this paper, we assume that $\beta^* = o(1)$ for principal component detection. It is worth noting that for shifted mean and principal component detection, we focus on the cases in which the nonzero entries of the signal vector take the same value, since they represent the most challenging settings. Our results can be easily extended to the cases where $\beta^* = \min_{j \in [d]} \theta_j$ or $\beta^* = \min_{j \in [d]} v_j^*$.

For the detection problems defined above, let $\mathbb{P}_0$ be the null distribution, and $\mathbb{P}_\mathcal{S}$ be the alternative distribution corresponding to $\mathcal{S} \in \mathcal{C}$. In the following, we define the risk measure for hypothesis tests.

**Risk Measure:** A hypothesis test $\phi$ maps the $n$ observations of $\boldsymbol{X} \in \mathbb{R}^d$ to $\{0, 1\}$. It takes zero if the null hypothesis $H_0$ is accepted and one otherwise. Throughout, we use the following risk measure for a test $\phi$,

$$R(\phi) = \mathbb{P}_0[\phi(\mathbf{X}) = 1] + \frac{1}{|\mathcal{C}|} \sum_{\mathcal{S} \in \mathcal{C}} \mathbb{P}_\mathcal{S}[\phi(\mathbf{X}) = 0], \tag{2.1}$$

where $\mathbf{X} \in \mathbb{R}^{n \times d}$ is the data matrix. The first and second terms on the right-hand side of (2.1) are the type-I and type-II errors correspondingly. It is worth noting that in our risk measure, the type-II error is averaged uniformly over the sets in $\mathcal{C}$. This risk measure is employed by previous works such as Arias-Castro et al. (2008); Addario-Berry et al. (2010). Lower bounds under the risk measure in (2.1) imply the lower bounds under the following risk measure,

$$R'(\phi) = \mathbb{P}_0[\phi(\mathbf{X}) = 1] + \max_{\mathcal{S} \in \mathcal{C}} \mathbb{P}_\mathcal{S}[\phi(\mathbf{X}) = 0],$$

which is also commonly used. Hereafter, we say a test is asymptotically powerful (resp., powerless) if its risk converges to zero (resp., one). It is worth mentioning that, for the simplicity of presentation, throughout this paper we assume the existence of limit for any sequence as $s^*, d, n$ go to infinity.

## 3 Oracle Computational Model

In the following, we introduce the oracle computational model. Let $\mathcal{X}$ be the domain of the random vector $\boldsymbol{X}$ and $\mathscr{A}$ be an algorithm.

**Definition 3.1** (Oracle Model). Under the oracle model, $\mathscr{A}$ can interact with an oracle $T$ rounds. Let $\mathcal{Q}_\mathscr{A}$ be the query space of $\mathscr{A}$. At each round, $\mathscr{A}$ uses a query function $q \in \mathcal{Q}_\mathscr{A} : \mathcal{X} \to [-b, b]$ to query an oracle $r$. The oracle returns a realization of $Z_q \in \mathbb{R}$, which satisfies

$$\mathbb{P}\big(\bigcap_{q \in \mathcal{Q}_\mathscr{A}} \{|Z_q - \mathbb{E}[q(\boldsymbol{X})]| \leq \tau_q\}\big) \geq 1 - 2\xi, \tag{3.1}$$

where $\tau_q = \max\{2b/3 \cdot [\eta(\mathcal{Q}_\mathscr{A}) + \log(1/\xi)]/n, \sqrt{2\operatorname{Var}[q(\boldsymbol{X})] \cdot [\eta(\mathcal{Q}_\mathscr{A}) + \log(1/\xi)]/n}\}$.

Here $\xi \in [0, 1/4)$, $\tau_q$ is the tolerance parameter and $\eta(\mathcal{Q}_\mathscr{A}) \geq 0$ measures the capacity of query space, e.g., if $\mathcal{Q}_\mathscr{A}$ is countable, we have $\eta(\mathcal{Q}_\mathscr{A}) = \log(|\mathcal{Q}_\mathscr{A}|)$. Hereafter $T$ is named as the oracle complexity. We denote by $\mathcal{R}[\xi, n, \eta(\mathcal{Q}_\mathscr{A})]$ the set of valid oracles that satisfy (3.1).

To understand the intuition behind Definition 3.1, we consider the ideal case where $Z_q = \mathbb{E}[q(\boldsymbol{X})]$ almost surely, i.e., the limiting case of our model with $\xi \to 0$, $[\eta(\mathcal{Q}_\mathscr{A}) + \log(1/\xi)]/n \to 0$. In this case, the algorithm can directly access information of the population distribution. For example, we obtain $\mathbb{E}(X_j)$ using $q(\mathbf{x}) = x_j$ ($j \in [d]$). Furthermore, this model allows adaptive choices of query functions based upon previous responses. For example, we consider the gradient ascent algorithm for maximum likelihood estimation. Let $f(\mathbf{x}; \boldsymbol{\theta})$ be the density function for the statistical model with parameter $\boldsymbol{\theta}$. At the $t$-th iteration, the gradient ascent algorithm employs $q(\mathbf{x}) = \partial \log[f(\mathbf{x}; \boldsymbol{\theta}^{t-1})]/\partial \theta_j$ ($j \in [d]$) to access the $j$-th coordinate of the gradient at $\boldsymbol{\theta}^{t-1}$ and updates $\boldsymbol{\theta}^t$ respectively. In the aforementioned ideal setting, under certain regularity conditions the algorithm converges to $\operatorname{argmax}_{\boldsymbol{\theta}} \mathbb{E}\{\log[f(\boldsymbol{X}; \boldsymbol{\theta})]\}$, which is the true parameter.

The ideal case shows $\mathscr{A}$ can only interact with data in a restrictive way under the oracle model. In particular, $\mathscr{A}$ only has access to the data distribution, but not to those individual data points. In other words, $\mathscr{A}$ makes decision based on global statistical properties. The restriction on the way $\mathscr{A}$ interacts with data captures the fundamental behavior of a broad range of algorithms for statistical problems, including convex optimization algorithms for $M$-estimation such as first-order methods or the coordinate descent algorithm, matrix decomposition algorithms for principal component analysis or factor models such as the power method or QR method, expectation-maximization algorithms for latent variable models, and sampling algorithms such as Markov chain Monte Carlo. See, e.g., Blum et al. (2005); Chu et al. (2007) for more details.

In contrast to the ideal case, with finite sample size $\mathscr{A}$ can only access the empirical distribution instead of the population distribution. Respectively, a common practice is to use sample average to replace expectation, which incurs a statistical error that decays with sample size. For a given query function with bounded value, this error is sharply characterized by Bernstein's inequality. Hereafter, we focus on bounded query functions for simplicity. In a follow-up paper, we will present extensions to unbounded queries. For query space with $|\mathcal{Q}_{\mathscr{A}}| > 1$, the uniform deviation of sample average from their expectation can be quantified using upper bounds for suprema of empirical processes. For the simplest example with countable $\mathcal{Q}_{\mathscr{A}}$, by combining Bernstein's inequality and union bound we have

$$\mathbb{P}\bigl(\bigcap_{q \in \mathcal{Q}_{\mathscr{A}}} \bigl\{\bigl|1/n \cdot \sum_{i=1}^n q(\mathbf{x}_i) - \mathbb{E}[q(\boldsymbol{X})]\bigr|\bigr\} \leq \tau_q\bigr) \geq 1 - 2\xi, \tag{3.2}$$

where $\tau_q$ is as defined in (3.1) with $\eta(\mathcal{Q}_{\mathscr{A}}) = \log(|\mathcal{Q}_{\mathscr{A}}|)$. For uncountable query space, we can obtain results similar to (3.2) by setting $\eta(\mathcal{Q}_{\mathscr{A}})$ in (3.1) to be other capacity measures in logarithmic scale, such as Vapnik-Chervonenkis dimension, metric entropy (with or without bracketing), as well as the corresponding Dudley's entropy integral. Respectively, those constants in $\tau_q$ may become larger and $\operatorname{Var}[q(\boldsymbol{X})]$ may be replaced with $\max_{q \in \mathcal{Q}_{\mathscr{A}}}\{\operatorname{Var}[q(\boldsymbol{X})]\}$. See, e.g., van der Vaart and Wellner (1996); Bousquet et al. (2004) for details. Later we will show that our computational lower bound holds for any $\mathcal{Q}_{\mathscr{A}}$ with $\eta(\mathcal{Q}_{\mathscr{A}}) \geq 0$, and thus implies lower bounds for both countable and uncountable query spaces.

For finite sample size, the motivation to study the oracle model is based upon a key observation: The success of the aforementioned algorithms only requires the bounded deviation of $1/n \cdot \sum_{i=1}^n q(\mathbf{x}_i)$ from $\mathbb{E}[q(\boldsymbol{X})]$ as in (3.2). In other words, if we replace $\mathscr{A}$'s access to $1/n \cdot \sum_{i=1}^n q(\mathbf{x}_i)$ with a random variable $Z_q$ that has the same deviation behavior, then $\mathscr{A}$ can achieve the same desired guarantees. In fact, such an observation forms the basis of the algorithmic analysis of a broad range of statistical problems, including $M$-estimation (Agarwal et al., 2012b; Xiao and Zhang, 2013; Wang et al., 2014c), principal component analysis (Yuan and Zhang, 2013; Wang et al., 2014d) and latent variable model estimation (Balakrishnan et al., 2014; Wang et al., 2014b). In detail, their analysis is deterministic conditioning on high-probability events characterizing the deviation of $1/n \cdot \sum_{i=1}^{n} q(\mathbf{x}_i)$. By replacing $1/n \cdot \sum_{i=1}^{n} q(\mathbf{x}_i)$ with the random variable $Z_q$, which has the same deviation behavior, their analysis yields the same guarantees. This key observation suggests that the oracle model in Definition 3.1 is a good proxy for studying lower bounds on the algorithmic complexity of statistical problems. In detail, if the success of $\mathscr{A}$ with access to data implies the success of $\mathscr{A}$ given access to the respective oracle, then the lower bound on the oracle complexity of $\mathscr{A}$ implies a lower bound on the number of rounds $\mathscr{A}$ needs to interact with data.

Definition 3.1 is an extension of the statistical query model proposed by Kearns (1998), which is improved by Blum et al. (1994, 1998); Servedio (1999); Yang (2001, 2005); Jackson (2003); Szörényi (2009); Feldman (2012); Feldman and Kanade (2012); Feldman et al. (2013, 2015). Our oracle model is based on the VSTAT oracle proposed by Feldman et al. (2013, 2015) with the following extensions.

- We explicitly incorporate the capacity of query space $\eta(\mathcal{Q}_{\mathscr{A}})$ and tail probability $\xi$ to quantify the uniform deviation of the responses for queries. In fact, the VSTAT oracle by Feldman et al. (2013, 2015) is a special case of Definition 3.1 with $\eta(\mathcal{Q}_{\mathscr{A}}) = 0$ and $\xi = 0$. For establishing upper bounds, such an assumption is too optimistic. More specifically, it may require more than $n$ data points to practically simulate the oracle with $\eta(\mathcal{Q}_{\mathscr{A}}) = 0$ and $\xi = 0$, i.e., to guarantee the same deviation of the responses for queries as in (3.1). As a result, even if $\mathscr{A}$ can achieve desired statistical accuracy by querying an oracle with $\eta(\mathcal{Q}_{\mathscr{A}}) = 0$ and $\xi = 0$, it can not achieve the same accuracy when its access to oracle is replaced with access to data, e.g., $1/n \cdot \sum_{i=1}^{n} q(\mathbf{x}_i)$. In other words, with access to an oracle that has $\eta(\mathcal{Q}_{\mathscr{A}}) = 0$ and $\xi = 0$, $\mathscr{A}$ may achieve a statistical accuracy that contradicts the information-theoretical lower bound. This phenomenon will be illustrated in §5.1.3. As we will show in Theorem 4.2, our general lower bound explicitly quantifies the effects of $\eta(\mathcal{Q}_{\mathscr{A}})$ and $\xi$.

- In our model, the queries are real-valued, while the VSTAT oracle by Feldman et al. (2013, 2015) only allows queries with discrete values, e.g., Boolean-valued queries. To translate the real-valued queries commonly used by practical algorithms to Boolean values, we have to pay an extra factor of $\log(b/\epsilon)$ in the oracle complexity for discretization, where $\epsilon$ is desired numerical accuracy and $b$ is defined in Definition 3.1. If $b$ and $1/\epsilon$ are large and increase with sample size or dimension, this extra factor induces gaps between lower and upper bounds. Since $\mathscr{A}$ can access more information at each iteration using real-valued queries, the lower bound on oracle complexity is more difficult to establish for our oracle model. To address this issue, we develop a new worst-case construction in the proof of Theorem 4.2, which differs from the one used by Szörényi (2009); Feldman (2012); Feldman et al. (2013, 2015). The power of this construction is from the fact that it allows a more fine-grained characterization of distinguishable distribution set, which will be defined in §4.1. Also, such a construction can be further extended to establish lower bounds for queries with unbounded values, which will be covered in a follow-up paper.

Definition 3.1 is also a generalization of the black-box model of convex optimization (Nemirovski and Yudin, 1983; Nesterov, 2004). In particular, given an unknown objective function, the algorithm can only query the zeroth-order optimization oracle, i.e., the objective value at a chosen location, or the first-order optimization oracle, i.e., the subgradient at a chosen location. Our model generalizes the black-box model by allowing more general query functions. Furthermore, our definition of oracle explicitly takes data into account and incorporates the notion of statistical error. In contrast, in the black-box model the oracles only implicitly depend on data through the objective functions. Besides, the techniques for establishing lower bounds in the black-box model are based upon pure worst-case construction, while for our model it is necessary to combine construction and information-theoretical techniques. It is worth noting that for stochastic convex optimization, the lower bounds are obtained via information-theoretical techniques (Shapiro and Nemirovski, 2005; Raginsky and Rakhlin, 2011; Agarwal et al., 2012a). However, their setting is different from ours, since they assume that the data points are accessed sequentially while we consider the batch setting.

## 4 Computational Lower Bound

In the following, we establish a general framework for characterizing the computational lower bound, which consists of two parts. First, in §4.1 we establish a general theorem, which connects the oracle complexity, testing risk, cardinality of $\mathcal{C}$, and the cardinality of distinguishable distribution set, which will be defined in Definition 4.1. In §4.2, we characterize the distinguishable distribution set by a key result involving the overlaps of $\mathcal{S}$ and $\mathcal{S}'$ from $\mathcal{C}$. For testing problems that satisfy certain regularity conditions, this lemma serves as a general interface for establishing upper bounds on the cardinality of distinguishable distribution set, which is used in §5 and §6 for specific examples.

### 4.1 General Theory

Before presenting the main theorem, we first define the distributions that can be distinguished from the null distribution $\mathbb{P}_0$ by a given query function. Then we introduce some notation.

**Definition 4.1** (Distinguishable Distribution). For a query function $q$, we define

$$\mathcal{C}(q) = \{\mathcal{S} : |\mathbb{E}_{\mathbb{P}_\mathcal{S}}[q(\boldsymbol{X})] - \mathbb{E}_{\mathbb{P}_0}[q(\boldsymbol{X})]| > \bar{\tau}_q, \ \mathcal{S} \in \mathcal{C}\}, \tag{4.1}$$

in which $\bar{\tau}_q$ denotes that, in the definition of $\tau_q$ in (3.1), $\text{Var}[q(\boldsymbol{X})]$ is taken under $\mathbb{P}_0$ and $\eta(\mathcal{Q}_\mathscr{A}) = 0$. We name $\{\mathbb{P}_\mathcal{S} : \mathcal{S} \in \mathcal{C}(q)\}$ as the set of distinguishable distributions.

For the oracle model defined in Definition 3.1, let $\mathcal{R}[\xi, n, \eta(\mathcal{Q}_\mathscr{A})]$ be the set of valid oracles that answer the queries of $\mathscr{A}$. Let $\mathcal{H}(\mathscr{A}, r)$ be the family of hypothesis tests that deterministically depend on $\mathscr{A}$'s queries to the oracle $r \in \mathcal{R}[\xi, n, \eta(\mathcal{Q}_\mathscr{A})]$ and its responses. We denote by $\mathcal{A}(T)$ the family of $\mathscr{A}$'s that interact with an oracle no more than $T$ times. Also, we define $\bar{\mathbb{P}}_0$ to be the distribution of the random variables returned by the oracle under the null hypothesis and define $\bar{\mathbb{P}}_\mathcal{S}$ correspondingly. We focus on the following risk measure

$$\overline{R}(\phi) = \bar{\mathbb{P}}_0(\phi = 1) + \frac{1}{|\mathcal{C}|} \sum_{\mathcal{S} \in \mathcal{C}} \bar{\mathbb{P}}_\mathcal{S}(\phi = 0), \tag{4.2}$$

which corresponds to (2.1). Recall that $\mathcal{Q}$ is the class of all possible queries.

**Theorem 4.2.** Under the oracle computational model in Definition 3.1, for any algorithm $\mathscr{A} \in \mathcal{A}(T)$, there exists an oracle $r \in \mathcal{R}[\xi, n, \eta(\mathcal{Q}_\mathscr{A})]$ such that

$$\inf_{\phi \in \mathcal{H}(\mathscr{A}, r)} \overline{R}(\phi) \geq \min\left\{1 - \frac{T \cdot \sup_{q \in \mathcal{Q}} |\mathcal{C}(q)|}{|\mathcal{C}|} + \min\left\{2\xi, T/|\mathcal{C}|, \sup_{q \in \mathcal{Q}} |\mathcal{C}(q)|/|\mathcal{C}|\right\}, T/|\mathcal{C}| + 1 - 2\xi, 1\right\}. \tag{4.3}$$

*Proof.* The proof idea is to construct a worst-case oracle $r$ for an arbitrary algorithm $\mathscr{A}$ to ensure any test $\phi \in \mathcal{H}(\mathscr{A}, r)$ fails to distinguish between $\mathbb{P}_0$ and $\mathbb{P}_\mathcal{S}$ for a large number of $\mathcal{S}$'s within $\mathcal{C}$. For any $\mathscr{A} \in \mathcal{A}(T)$ that makes queries $\{q_t\}_{t=1}^T \in \mathcal{Q}^T$ to $r$ with $\mathcal{Q}^T$ being the $T$-th cartesian power of $\mathcal{Q}$, we show that any $\phi$ must have large testing error in accepting or rejecting $\mathbb{P}_\mathcal{S}$ for $\mathcal{S} \in \mathcal{C} \setminus \bigcup_{t \in [T]} \mathcal{C}(q_t)$, which yields the $T \cdot \sup_{q \in \mathcal{Q}} |\mathcal{C}(q)|/|\mathcal{C}|$ term in (4.3) by further calculation. To prove a stronger lower bound on $\overline{R}(\phi)$, we consider two more refined settings on the mutual intersections of $\{\mathcal{C}(q_t)\}_{t=1}^T$, i.e., whether there exists a sequence $\{\mathcal{S}^t\}_{t=1}^T$ that satisfies $\mathcal{S}^t \in \mathcal{C}(q_t)$ and $\mathcal{S}^t \notin \bigcup_{t' \neq t} \mathcal{C}(q_{t'})$ for all $t \in [T]$. This allows us to obtain the rest terms on the right-hand side of (4.3), in particular the dependency on the tail probability $\xi$. See §7.1 for a detailed proof. ∎

Theorem 4.2 can be understood as follows. Let $\xi = 0$ for simplicity. If $T \cdot \sup_{q \in \mathcal{Q}} |\mathcal{C}(q)|/|\mathcal{C}| = o(1)$, then the testing error is at least lower bounded by $1 - o(1)$, i.e., any test $\phi \in \mathcal{H}(\mathscr{A}, r)$ is asymptotically powerless. In other words, the oracle complexity $T$ needs to be at least of the order $|\mathcal{C}|/\sup_{q \in \mathcal{Q}} |\mathcal{C}(q)|$, which relates oracle complexity lower bound to the upper bound of $\sup_{q \in \mathcal{Q}} |\mathcal{C}(q)|$, i.e., the cardinality of the largest distinguishable distribution set. The intuition is that, if the queries are effective in the sense that each query can distinguish $\mathbb{P}_\mathcal{S}$ from $\mathbb{P}_0$ for a large number of $\mathcal{S}$'s in $\mathcal{C}$, then we only need a small number of queries to distinguish $H_1$ from $H_0$. Recall that for the testing problems in §2, $|\mathcal{C}|$ is generally large, e.g., $|\mathcal{C}|$ can be exponential in $d$. To prove that $T$ needs to be at least exponential in $d$ to ensure small testing error, it only remains to prove $\sup_{q \in \mathcal{Q}} |\mathcal{C}(q)|$ is small. Furthermore, (4.3) quantifies the continuous transition from the feasible regime to the infeasible regime as one decreases the computational budget $T$, which is not captured by previous works that build on computational hardness assumptions, e.g., Berthet and Rigollet (2013a,b).

Note that (4.3) in Theorem 4.2 captures the effect of the tail probability $\xi$ in our oracle model in Definition 3.1. In particular, we are interested in the setting where both $T$ and $\sup_{q \in \mathcal{Q}} |\mathcal{C}(q)|$ are large with respect to $|\mathcal{C}|$, since otherwise the testing error is already very large for any $\xi$ according to our previous discussion. In this setting, (4.3) shows that the lower bound on testing error increases with $\xi$, since the oracle returns a random variable with larger tail probability.

Theorem 4.2 characterizes the complexity of a wide range of algorithms for statistical problems. As discussed in §3, for a broad family of algorithms, their access to data can be replaced with access to any oracle that satisfies Definition 3.1. In other words, if $\mathscr{A}$ succeeds with $T$ queries to data, then $\mathscr{A}$ will also succeed using $T$ queries to any oracle satisfying Definition 3.1. By Theorem 4.2, there is an oracle satisfying Definition 3.1 such that any $\mathscr{A} \in \mathcal{A}(T)$ fails under certain conditions. Therefore, under the same conditions, any $\mathscr{A}$ captured by the oracle model also fails if $\mathscr{A}$ interacts with data no more than $T$ rounds. Therefore, the total number of operations required by $\mathscr{A}$ is at least $T$.

Theorem 4.2 builds upon the notion of distinguishable distribution, which has a deep connection to the duality between the spaces of measures and functions. In particular, $\mathbb{E}_{\mathbb{P}_0}[q(\boldsymbol{X})]$ and $\mathbb{E}_{\mathbb{P}_\mathcal{S}}[q(\boldsymbol{X})]$

in (4.1) are indeed the pairings between the query function $q$, which lies in the function space, with $\mathbb{P}_0$ and $\mathbb{P}_\mathcal{S}$, which lie in the space of measures, i.e., the dual of the function space. The intuition behind Theorem 4.2 is that, the algorithms captured by the oracle model are essentially employing elements in the function space to separate a subset $\{\mathbb{P}_\mathcal{S} : \mathcal{S} \in \mathcal{C}\}$ with an element $\mathbb{P}_0$ in the space of measures. Theorem 4.2 establishes a fundamental connection between oracle complexity and the complexity of the space of measures, which is quantified using elements in its dual space. Furthermore, this duality point of view is connected to two closely related fields.

- Wasserstein distance, Radon distance and total variation distance in information theory can also be viewed from the above duality perspective. More specifically, they use a subset of the function space to pair with the measures. Particularly, Wasserstein distance uses Lipschitz functions, Radon distance uses bounded functions, and total variation distance uses indicator functions.

- Under the black-box model for convex optimization, algorithms can query the subgradient of the objective function. The space of subgradients is dual to the space of the decision variables of the objective function. For example, the definition of Bregman divergence involves the pairing between the subgradient and difference between two decision variables.

Particularly, the notion of distinguishable distribution can be viewed as a refinement of the level set under Radon distance if we restrict $\mathcal{Q}$ to continuous queries. In detail, the Radon distance between $\mathbb{P}$ and $\mathbb{P}'$ is defined as

$$\rho(\mathbb{P}, \mathbb{P}') = \sup\{\mathbb{E}_\mathbb{P}[q(\boldsymbol{X})] - \mathbb{E}_{\mathbb{P}'}[q(\boldsymbol{X})] \mid \text{continuous } q : \mathcal{X} \to [-1, 1]\}.$$

For simplicity, we replace $\bar{\tau}_q$ in Definition 4.1 with $\bar{\tau}$, which is a quantity that does not depend on $q$, and set $b = 1$ in Definition 3.1. For any $\mathbb{P}_\mathcal{S} \in \mathcal{C}(q)$, it holds that $\rho(\mathbb{P}_\mathcal{S}, \mathbb{P}_0) \leq \bar{\tau}$, i.e., $\mathcal{C}(q)$ is a subset of the lower level set of $\rho(\cdot, \mathbb{P}_0)$. In other words, we can view $\sup_{q \in \mathcal{Q}} |\mathcal{C}(q)|$ as a lower approximation of the cardinality of this level set.

The high-level proof idea of Theorem 4.2 is from Feldman et al. (2013, 2015), which build upon Szörényi (2009); Feldman (2012). Compared with Feldman et al. (2013, 2015), which consider query functions with discrete values, we allow query functions with real values in Definition 3.1. In order to establish lower bound under this more powerful oracle model, we propose a new construction of the worst-case oracle. Roughly speaking, our worst-case oracle responds query $q$ with a random variable that follows the uniform distribution over $\{\mathbb{E}_{\mathbb{P}_\mathcal{S}}[q(\boldsymbol{X})] : \mathcal{S} \in \widetilde{\mathcal{C}}\}$ under $H_0$, where $\widetilde{\mathcal{C}}$ is a certain subset of $\mathcal{C}$. Under $H_1$, the constructed oracle faithfully responds query $q$ using $\mathbb{E}_{\mathbb{P}_\mathcal{S}}[q(\boldsymbol{X})]$. In contrast, the worst-case oracle in Feldman et al. (2013, 2015) responds query $q$ with $\mathbb{E}_{\mathbb{P}_0}[q(\boldsymbol{X})]$ both under $H_0$ and for specific $\mathbb{P}_\mathcal{S}$'s under $H_1$. Our new construction leads to a more refined characterization of $\mathcal{C}(q)$. In particular, because of our new construction, the $\text{Var}[q(\boldsymbol{X})]$ term within $\bar{\tau}_q$ in Definition 4.1 is taken under $\mathbb{P}_0$, rather than $\mathbb{P}_\mathcal{S}$ as in Feldman et al. (2013, 2015). As will be shown in §4.2, this leads to a tighter upper bound on $|\mathcal{C}(q)|$ uniformly for any $q \in \mathcal{Q}$ when the queries are real-valued. Besides, as discussed in §3, we explicitly incorporate the tail probability $\xi$ in our proof using a characterization of the mutual intersections of $\{\mathcal{C}(q_t)\}_{t=1}^T$. See §7.1 for more details.

## 4.2 Characterization of Distinguishable Distributions

In this section, we characterize the size of the distinguishable distribution set $\mathcal{C}(q)$, which is defined in Definition 4.1. Once we have an upper bound on $\sup_{q\in\mathcal{Q}}|\mathcal{C}(q)|$, Theorem 4.2 yields a lower bound on the testing error under computational constraints. In the following we impose a regularity condition, which significantly simplifies our presentation. Then we will present a key theorem connecting $|\mathcal{C}(q)|$ with the overlaps of a fixed $\mathcal{S}$ with $\mathcal{S}'$ uniformly drawn from $\mathcal{C}$. Based on this theorem, establishing the upper bound of $|\mathcal{C}(q)|$ reduces to quantifying the level set of a specific function that only depends on $\mathbb{P}_0$, $\mathbb{P}_\mathcal{S}$ ($\mathcal{S} \in \mathcal{C}$), and the structure of $\mathcal{C}$, which will be further illustrated by examples in §5 and §6.

**Condition 4.3.** We assume that the following two conditions hold.

- For $\mathcal{S}'$ drawn uniformly at random from $\mathcal{C}$ and $\mathcal{S} \in \mathcal{C}$ fixed, the distribution of $|\mathcal{S} \cap \mathcal{S}'|$ is the same for all $\mathcal{S}$.

- For any $\mathcal{S}_1, \mathcal{S}_2 \in \mathcal{C}$, it holds that

$$\mathbb{E}_{\mathbb{P}_0}\left[\frac{\mathrm{d}\mathbb{P}_{\mathcal{S}_1}}{\mathrm{d}\mathbb{P}_0}\frac{\mathrm{d}\mathbb{P}_{\mathcal{S}_2}}{\mathrm{d}\mathbb{P}_0}(\boldsymbol{X})\right] = h(|\mathcal{S}_1 \cap \mathcal{S}_2|), \tag{4.4}$$

where $h$ is a nondecreasing function with $h(0) \geq 1$.

The first condition characterizes the symmetricity of $\mathcal{C}$. In detail, it states that $\mathcal{C}$ is homogeneous in terms of the intersection between its elements. This condition is previously used by Addario-Berry et al. (2010) to simplify the information-theoretical lower bounds for combinatorial testing problems. See Proposition 3.3 therein. As illustrated by Addario-Berry et al. (2010), a broad range of structures are symmetric, e.g., disjoint sets, sparse sets, perfect matchings, stars, and spanning trees. Here we focus on $\mathcal{C}$ consisting of sparse sets and perfect matchings, which are defined in §2.

The second condition is on $\mathbb{P}_0$ and $\mathbb{P}_\mathcal{S}$ ($\mathcal{S} \in \mathcal{C}$), which depends on the statistical model and testing problem. The left-hand side of (4.4) is the expected product of likelihood ratios, which plays a vital role in information-theoretical lower bounds. In detail, note that establishing information-theoretical lower bounds via Le Cam's method often involves

$$\chi^2(\mathbb{P}_{\mathcal{S}\in\mathcal{C}}, \mathbb{P}_0) = \mathbb{E}_{\mathbb{P}_0}\left\{\left[\frac{\mathrm{d}\mathbb{P}_{\mathcal{S}\in\mathcal{C}}}{\mathrm{d}\mathbb{P}_0}(\boldsymbol{X}) - 1\right]^2\right\} = \frac{1}{|\mathcal{C}|^2}\sum_{\mathcal{S}_1,\mathcal{S}_2\in\mathcal{C}}\left\{\mathbb{E}_{\mathbb{P}_0}\left[\frac{\mathrm{d}\mathbb{P}_{\mathcal{S}_1}}{\mathrm{d}\mathbb{P}_0}\frac{\mathrm{d}\mathbb{P}_{\mathcal{S}_2}}{\mathrm{d}\mathbb{P}_0}(\boldsymbol{X})\right] - 1\right\},$$

where $\mathbb{P}_{\mathcal{S}\in\mathcal{C}}$ is the uniform mixture of all $\mathbb{P}_\mathcal{S}$'s with $\mathcal{S} \in \mathcal{C}$. That is to say, the left-hand side of (4.4) is the cross term within the $\chi^2$-divergence between $\mathbb{P}_{\mathcal{S}\in\mathcal{C}}$ and $\mathbb{P}_0$. The condition states that the cross terms only depend on $\mathcal{S}_1$ and $\mathcal{S}_2$ via $|\mathcal{S}_1 \cap \mathcal{S}_2|$, and the statistical model via the monotone $h$ function. As will be shown in Lemmas 5.1 and 6.1, such a condition holds for the problems in §2. Furthermore, this condition also holds for more testing problems, e.g., the detection of Gaussian mixture (Azizyan et al., 2013; Verzelen and Arias-Castro, 2014), positive correlation (Arias-Castro et al., 2012, 2015a), and Markov random field (Arias-Castro et al., 2015b). Finally, it is worth noting that Condition 4.3 is used to simplify the characterization of $|\mathcal{C}(q)|$ and unify the analysis for specific testing problems. Our high-level proof strategy will also work for problems for which Condition 4.3 does not hold, but may require case-by-case treatments.

The following theorem connects the upper bound of $|\mathcal{C}(q)|$ with a combinatorial quantity involving the overlaps of a fixed $\mathcal{S}$ with $\mathcal{S}'$ uniformly drawn from $\mathcal{C}$. For $\mathcal{S} \in \mathcal{C}(q)$, we define $\bar{\mathcal{C}}(q, \mathcal{S}) \subseteq \mathcal{C}$ as

$$\bar{\mathcal{C}}(q, \mathcal{S}) = \underset{|\bar{\mathcal{C}}|=|\mathcal{C}(q)|}{\operatorname{argmax}} \{ \mathbb{E}_{\mathcal{S}' \sim \mathbb{Q}_{\bar{\mathcal{C}}}}[h(|\mathcal{S} \cap \mathcal{S}'|)] \}. \tag{4.5}$$

Here $\mathbb{Q}_{\bar{\mathcal{C}}}$ is the uniform distribution over $\bar{\mathcal{C}}$, which is a subset of $\mathcal{C}$. Note that $\bar{\mathcal{C}}(q, \mathcal{S})$ is not necessarily unique and our following results hold for any $\bar{\mathcal{C}}(q, \mathcal{S})$.

**Theorem 4.4.** Under Condition 4.3, it holds that

$$\sup_{\mathcal{S} \in \mathcal{C}(q)} \{ \mathbb{E}_{\mathcal{S}' \sim \mathbb{Q}_{\bar{\mathcal{C}}(q,\mathcal{S})}}[h(|\mathcal{S} \cap \mathcal{S}'|)] \} \geq 1 + \log(1/\xi)/n. \tag{4.6}$$

Here the left-hand side only depends on $|\mathcal{C}(q)|$, and is a nonincreasing function of $|\mathcal{C}(q)|$.

*Proof.* We sketch the proof as follows. Recall that $\mathcal{C}(q)$ is defined in (4.1). For notational simplicity, we define

$$\mathbb{P}_{\mathcal{S} \in \mathcal{C}(q)} = \frac{1}{|\mathcal{C}(q)|} \sum_{\mathcal{S} \in \mathcal{C}(q)} \mathbb{P}_{\mathcal{S}} \tag{4.7}$$

as the uniform mixture of all $\mathbb{P}_{\mathcal{S}}$'s with $\mathcal{S} \in \mathcal{C}(q)$. The proof of Theorem 4.4 is based on the following key lemma, which characterizes the $\chi^2$-divergence between $\mathbb{P}_{\mathcal{S} \in \mathcal{C}(q)}$ and $\mathbb{P}_0$.

**Lemma 4.5.** Under Condition 4.3, for any query function $q$ it holds that

$$\chi^2 ( \mathbb{P}_{\mathcal{S} \in \mathcal{C}(q)}, \mathbb{P}_0 ) \geq \log(1/\xi)/n.$$

*Proof.* The proof is based on an application of Cauchy-Schwarz inequality to $|\mathbb{E}_{\mathbb{P}_{\mathcal{S}}}[q(\boldsymbol{X})] - \mathbb{E}_{\mathbb{P}_0}[q(\boldsymbol{X})]|$ in Definition 4.1. See §7.2 for a detailed proof. □

By Lemma 4.5, we prove (4.6) by the definition of $\chi^2$-divergence, the symmetry of $\mathcal{C}$ in Condition 4.3, and (4.4). We prove the claim that the left-hand side of (4.6) only depends on $|\mathcal{C}(q)|$ by explicitly constructing $\bar{\mathcal{C}}(q, \mathcal{S})$ in (4.5) and utilizing the symmetry of $\mathcal{C}$. We prove the claim on the monotonicity with respect to $|\mathcal{C}(q)|$ using the monotonicity of $h$ in Condition 4.3. See §7.3 for a detailed proof. □

Theorem 4.4 can be understood as follows. Since $h$ is nondecreasing, we can construct $\bar{\mathcal{C}}(q, \mathcal{S})$ in (4.5) explicitly. More specifically, we start with an empty set, and sequentially add in $\mathcal{S}'$ that has the top largest overlaps with $\mathcal{S}$ until this set has cardinality $|\mathcal{C}(q)|$. In another word, we first add $\mathcal{S}$ itself to $\bar{\mathcal{C}}(q, \mathcal{S})$, then all the $\mathcal{S}' \in \mathcal{C}$ with Hamming distance one from $\mathcal{S}$, and so on, until $|\bar{\mathcal{C}}(q, \mathcal{S})| = |\mathcal{C}(q)|$. Here the Hamming distance between $\mathcal{S}$ and $\mathcal{S}'$ is defined as $s^* - |\mathcal{S} \cap \mathcal{S}'|$. Theorem 4.4 gives a lower bound on $\mathbb{E}[h(|\mathcal{S} \cap \mathcal{S}'|)]$, where $\mathcal{S}'$ is uniformly drawn from $\bar{\mathcal{C}}(q, \mathcal{S})$. As we enlarge $|\mathcal{C}(q)| = |\bar{\mathcal{C}}(q, \mathcal{S})|$, $\bar{\mathcal{C}}(q, \mathcal{S})$ encompasses more sets that have larger Hamming distance from $\mathcal{S}$. Hence, the left-hand side of (4.6) decreases along $|\mathcal{C}(q)|$. This observation implies that, the smallest $|\mathcal{C}(q)|$ such that (4.6) fails is an upper bound of $\sup_{q \in \mathcal{Q}} |\mathcal{C}(q)|$, which follows from proof by contradiction. In this way, Theorem 4.4 connects the upper bound of $\sup_{q \in \mathcal{Q}} |\mathcal{C}(q)|$ with the combinatorial quantity on the left-hand side of (4.6). Therefore, it remains to calculate this combinatorial quantity for specific testing problems, which will be presented in §5 and §6.

The proof of Theorem 4.4 reflects the difference between our oracle complexity lower bound and classical information-theoretical lower bound. In particular, for the testing problems in §2, Le Cam's lemma states that if the divergence between $\mathbb{P}_{\mathcal{S} \in \mathcal{C}}$ and $\mathbb{P}_0$, e.g., $\chi^2(\mathbb{P}_{\mathcal{S} \in \mathcal{C}}, \mathbb{P}_0)$, is small, then the risk of any test is large. In contrast, as shown in Lemma 4.5 the oracle complexity lower bound relies on $\chi^2(\mathbb{P}_{\mathcal{S} \in \mathcal{C}(q)}, \mathbb{P}_0)$. In other words, the oracle complexity lower bound utilizes the local structure of the alternative distribution family $\{\mathbb{P}_{\mathcal{S}} : \mathcal{S} \in \mathcal{C}\}$, rather than its global structure as in Le Cam's method. From this perspective, we can view Theorem 4.2 as a localized refinement of Le Cam's lower bound, which incorporates computational complexity constraints.

As previously discussed, the proof of Theorem 4.2 is based upon a new construction of worst-case oracle. Corresponding to this new construction, the $\text{Var}[q(\boldsymbol{X})]$ term in Definition 4.1 is taken under $\mathbb{P}_0$, instead of $\mathbb{P}_{\mathcal{S}}$ as in Feldman et al. (2013, 2015). This allows us to establish the lower bound for $\chi^2(\mathbb{P}_{\mathcal{S} \in \mathcal{C}(q)}, \mathbb{P}_0)$ in Lemma 4.5, which is independent of specific choices of $q$. As shown in the proof of Lemma 4.5, this query-independent lower bound is made possible by cancelling the $\text{Var}[q(\boldsymbol{X})]$ terms within the upper bound of $|\mathbb{E}_{\mathbb{P}_{\mathcal{S}}}[q(\boldsymbol{X})] - \mathbb{E}_{\mathbb{P}_0}[q(\boldsymbol{X})]|$ and $\bar{\tau}_q$, since both of them are evaluated under $\mathbb{P}_0$. In contrast, if $\text{Var}[q(\boldsymbol{X})]$ in Definition 4.1 is taken under $\mathbb{P}_{\mathcal{S}}$, the lower bound of $\chi^2(\mathbb{P}_{\mathcal{S} \in \mathcal{C}(q)}, \mathbb{P}_0)$ has a dependency on $q$, which can not be eliminated when $q$ is real-valued. In particular, for specific settings of $\mathbb{P}_{\mathcal{S}}$ and $\mathbb{P}_0$, the resulting lower bound of $\chi^2(\mathbb{P}_{\mathcal{S} \in \mathcal{C}(q)}, \mathbb{P}_0)$ is much smaller than $\log(1/\xi)/n$, which leads to a more loose upper bound of $\sup_{q \in \mathcal{C}} |\mathcal{C}(q)|$ in Theorem 4.2.

## 5 Implication for Shifted Mean Detection

In the following, we employ Theorems 4.2 and 4.4 to explore the computational and statistical phase transition for the shifted mean detection problem in §2. We consider $\mathcal{C}$ being the class of sparse sets in §5.1 and the class of perfect matchings in §5.2. For each class, we first establish the computational lower bounds, then the information-theoretical lower bounds, and finally the matching upper bounds. In the following, we verify Condition 4.3. The next lemma specifies the $h$ function in (4.4).

**Lemma 5.1.** For any $\mathcal{S}_1, \mathcal{S}_2 \subseteq [d]$ with $|\mathcal{S}_1| = |\mathcal{S}_2| = s^*$, for the shifted mean detection problem in §2, it holds that

$$\mathbb{E}_{\mathbb{P}_0}\left[\frac{\mathrm{d}\mathbb{P}_{\mathcal{S}_1}}{\mathrm{d}\mathbb{P}_0}\frac{\mathrm{d}\mathbb{P}_{\mathcal{S}_2}}{\mathrm{d}\mathbb{P}_0}(\boldsymbol{X})\right] = \alpha^2 \exp\bigl(|\mathcal{S}_1 \cap \mathcal{S}_2|\beta^{*2}\bigr) + (1 - \alpha^2).$$

*Proof.* See §7.4 for a detailed proof. □

As shown in Addario-Berry et al. (2010), the classes of sparse sets and perfect matchings satisfy the symmetricity assumption in Condition 4.3. Together with Lemma 5.1, we conclude that Theorem 4.4 holds, which forms the basis of our following results.

### 5.1 Sparse Set Detection

In this section, we consider the sparse set detection problem in §2, in which $\mathcal{C} = \{\mathcal{S} \subseteq [d] : |\mathcal{S}| = s^*\}$. We start with the computational lower bound.

### 5.1.1 Computational Lower Bound

As previously discussed in §4, the oracle complexity lower bound boils down to the upper bound of $\sup_{q\in\mathcal{Q}}|\mathcal{C}(q)|$, which shows up in Theorem 4.2. The next lemma quantifies $\sup_{q\in\mathcal{Q}}|\mathcal{C}(q)|$ based upon the combinatorial characterization in Theorem 4.4. For notational simplicity we define

$$\zeta = d/(2s^{*2}), \quad \tau = \sqrt{\log(1/\xi)/n}, \quad \text{and} \quad \gamma = d/(2s^{*2})\cdot \log(1+\tau^2/\alpha^2)/(2\beta^{*2}). \tag{5.1}$$

**Lemma 5.2.** We consider the following settings: (i) $s^{*2}/d = o(1)$; (ii) $\lim_{d\to\infty} s^{*2}/d > 0$. For setting (i), we have

$$\sup_{q\in\mathcal{Q}}|\mathcal{C}(q)| \leq 2\exp\big\{-\log\zeta \cdot \big[\log(1+\tau^2/\alpha^2)/\beta^{*2} - 2\big]\big\}|\mathcal{C}|.$$

For setting (ii), it holds that

$$\sup_{q\in\mathcal{Q}}|\mathcal{C}(q)| \leq 2\exp\big[-\log\gamma \cdot \big(2\gamma s^{*2}/d - 1\big)\big]|\mathcal{C}|.$$

*Proof.* We first calculate the combinatorial quantity in (4.6) and its lower bound. Then we quantify the level set of this lower bound to establish the upper bound of $\sup_{q\in\mathcal{Q}}|\mathcal{C}(q)|$. In particular, settings (i) and (ii) respectively capture two different behaviors of the combinatorial quantity on the left-hand side of (4.6). See §7.5 for a detailed proof. □

Based on Lemma 5.2, the following theorem characterizes the computational lower bound in terms of $s^*$, $d$, $n$, $\beta^*$, and $\alpha$. Particularly, we provide sufficient conditions under which $T\cdot \sup_{q\in\mathcal{Q}}|\mathcal{C}(q)|/|\mathcal{C}| = o(1)$ when $T$ is polynomial in $d$. Hence, by Theorem 4.2, under these conditions the testing risk is at least $1 - 2\xi$, i.e., any test is asymptotically powerless for $\xi = o(1)$. Now we introduce several settings using the notation in (5.1). In detail, under setting (i) in Lemma 5.2, we consider

(a) $s^{*2}/d = o(1)$, $\lim_{d\to\infty} \zeta/d^\delta > 0$, $\lim_{d\to\infty} \tau^2/\alpha^2 > 0$, and $\beta^* = o(1)$;

(b) $s^{*2}/d = o(1)$, $\lim_{d\to\infty} \zeta/d^\delta > 0$, $\tau^2/\alpha^2 = o(1)$, and $\beta^{*2}n\alpha^2 = o(1)$;

(c) $s^{*2}/d = o(1)$, $\zeta/d^\delta = o(1)$, $\lim_{d\to\infty} \tau^2/\alpha^2 > 0$, and $\beta^{*2}\log d = o(1)$;

(d) $s^{*2}/d = o(1)$, $\zeta/d^\delta = o(1)$, $\tau^2/\alpha^2 = o(1)$, and $\beta^{*2}n\alpha^2 = o(1)$.

Here $\delta > 0$ is a constant that is sufficiently small. Under setting (ii) in Lemma 5.2, we consider

(a) $\lim_{d\to\infty} s^{*2}/d > 0$, $\tau^2/\alpha^2 = o(1)$, and $\beta^{*2}s^{*2}n\alpha^2/d = o(1)$;

(b) $\lim_{d\to\infty} s^{*2}/d > 0$, $\lim_{d\to\infty} \tau^2/\alpha^2 > 0$, and $\beta^{*2}s^{*2}\log d/d = o(1)$.

We use notation such as (i).(a) and (ii).(b) to refer to the above settings. The next theorem suggests that, under any of the above settings, any test that is based on an algorithm with polynomial oracle complexity is asymptotically powerless.

**Theorem 5.3.** For $T = O(d^\eta)$, where $\eta$ is any constant and $T \geq 1$, we have $T \cdot \sup_{q\in\mathcal{Q}}|\mathcal{C}(q)|/|\mathcal{C}| = o(1)$ under any of (i).(a) to (i).(d) or any of (ii).(a) to (ii).(b) defined above.

*Proof.* The proof follows from plugging Lemma 5.2 in Theorem 4.2. See §7.6 for a detailed proof. □

We will discuss the implication of Theorem 5.3 for computational and statistical phase transition in §5.1.4 after we establish the information-theoretical lower bound and the respective upper bounds.

### 5.1.2 Information-theoretical Lower Bound

The following proposition establishes the information-theoretical lower bound. Recall that, as defined in (2.1), $R(\phi)$ is the risk of the test $\phi$.

**Proposition 5.4.** We consider two cases: (i) $\beta^{*2}\alpha^2 n = o(1)$, $\beta^{*2}s^* = o(1)$, and $\beta^{*2}\alpha^2 n s^{*2}/d = o(1)$; (ii) $\beta^{*2}s^{*2}/d = o(1)$, $\beta^{*2}\alpha n = o(1)$, and $\beta^{*2}\alpha^2 n s^{*2}/d = o(1)$. For $s^*$, $d$, and $n$ sufficiently large, under setting (i) or (ii) we have $\inf_\phi R(\phi) \geq 1 - \epsilon$ with $\epsilon = o(1)$.

*Proof.* We use Le Cam's method by developing upper bounds for $\chi^2(\mathbb{P}^n_{\mathcal{S}\in\mathcal{C}}, \mathbb{P}^n_0)$ in both settings. See §7.7 for a detailed proof. □

As we will show in §5.1.4, such a lower bound is tight up to logarithmic factors. In fact, using a similar truncation argument for $\chi^2$-divergence in Butucea and Ingster (2013), we can eliminate these factors. See the proof of Theorem 2.2 therein for more details. However, since our major focus is on the computational lower bounds, we do not further pursue this direction in this paper.

### 5.1.3 Upper Bounds

In the sequel, we construct upper bounds under the oracle model. We consider the following settings.

(i) $\beta^{*2}s^{*2}/(d\log n) \to \infty$ and $\alpha n \geq C\log(1/\xi)$ with $C$ being a sufficiently large positive absolute constant. We consider an algorithm $\mathscr{A}$ with $\eta(\mathcal{Q}_\mathscr{A}) = 0$ that uses only one query function,

$$q(\boldsymbol{X}) = \mathbb{1}\bigl(1/\sqrt{d} \cdot \textstyle\sum_{j=1}^d X_j \geq \sqrt{2\log n}\bigr). \tag{5.2}$$

We define the test as

$$\mathbb{1}\bigl[z_q \geq 1 - \Phi\bigl(\sqrt{2\log n}\bigr) + \alpha/8\bigr], \tag{5.3}$$

where $z_q$ is the realization of $Z_q$ as in Definition 3.1, and $\Phi$ is the Gaussian cumulative density function. In the above and following settings, $b$ in Definition 3.1 equals one.

(ii) $\beta^{*2}n\alpha^2/[\log d + \log(1/\xi)] \to \infty$. We consider an algorithm $\mathscr{A}$ that uses the following sequence of queries,

$$q_t(\boldsymbol{X}) = \mathbb{1}(X_t \geq \beta^*/2), \tag{5.4}$$

where $t \in [T]$ and $T = d$. In other words, the query space $\mathcal{Q}_\mathscr{A}$ of $\mathscr{A}$ is discrete with $|\mathcal{Q}_\mathscr{A}| = d$, i.e., $\eta(\mathcal{Q}_\mathscr{A}) = \log d$. We define the test as

$$\mathbb{1}\bigl[\sup_{t\in[T]} z_{q_t} \geq 1 - \Phi(\beta^*/2) + \alpha\beta^*/(4\pi)\bigr]. \tag{5.5}$$

(iii) $\beta^{*2}s^{*2}n\alpha^2/[d\log(1/\xi)] \to \infty$ and $\alpha n \geq C\log(1/\xi)$. Here $C$ is a sufficiently large constant. We consider $\mathscr{A}$ with $\eta(\mathcal{Q}_{\mathscr{A}}) = 0$, which uses only one query function,

$$q(\boldsymbol{X}) = \mathbb{1}\big(\textstyle\sum_{j=1}^d X_j \geq \beta^* s^*/2\big). \tag{5.6}$$

We define the test as

$$\mathbb{1}\big\{z_q \geq 1 - \Phi\big[\beta^* s^*/\big(2\sqrt{d}\big)\big] + \alpha\beta^* s^*/\big(4\pi\sqrt{d}\big)\big\}. \tag{5.7}$$

(iv) $\beta^{*2}s^*/\log n \to \infty$, $\beta^{*2}n\alpha/(\log d \cdot \log n) \to \infty$, and $\log(1/\xi) \leq \min\{s^*\log d, \alpha n\}$. Let $C \geq 0$ be a constant that is sufficiently large. Under this setting, we consider two cases.

    (a) $s^* < n\alpha/(C\log d)$. We consider an algorithm $\mathscr{A}$ that uses the following sequence of query functions,

$$q_t(\boldsymbol{X}) = \mathbb{1}\big(\textstyle\sum_{j \in \mathcal{S}_t} X_j \geq \beta^* s^*/2\big), \tag{5.8}$$

where $t \in [T]$ with $T = \binom{d}{s^*}$ and $|\mathcal{S}_t| = s^*$, while $\bigcup_{t=1}^T \mathcal{S}_t = [d]$. In other words, the query space is discrete with $|\mathcal{Q}_{\mathscr{A}}| = \binom{d}{s^*}$, i.e., $\eta(\mathcal{Q}_{\mathscr{A}}) = \log\binom{d}{s^*}$. We define the test as

$$\mathbb{1}\big[\sup_{t \in [T]} z_{q_t} \geq 1 - \Phi\big(\beta^*\sqrt{s^*}/2\big) + \alpha/4\big]. \tag{5.9}$$

    (b) $s^* \geq n\alpha/(C\log d)$. Let $\bar{s}^* = 2n\alpha/(C\log d)$. We consider an algorithm $\mathscr{A}$ that uses

$$q_t(\boldsymbol{X}) = \mathbb{1}\big(\textstyle\sum_{j \in \bar{\mathcal{S}}_t} X_j \geq \beta^* \bar{s}^*/2\big), \tag{5.10}$$

where $t \in [T]$ with $T = \binom{d}{\bar{s}^*}$ and $|\bar{\mathcal{S}}_t| = \bar{s}^*$, while $\bigcup_{t=1}^T \bar{\mathcal{S}}_t = [d]$. We have $\eta(\mathcal{Q}_{\mathscr{A}}) = \log\binom{d}{\bar{s}^*}$. We define the test as

$$\mathbb{1}\big[\sup_{t \in [T]} z_{q_t} \geq 1 - \Phi\big(\beta^*\sqrt{\bar{s}^*}/2\big) + \alpha/4\big]. \tag{5.11}$$

Recall that $\xi$ is the tail probability of the oracle model in Definition 3.1. The next theorem establishes upper bounds for the risk of above tests. In particular, it suggests that those tests are asymptotically powerful for $\xi = o(1)$.

**Theorem 5.5.** *For settings (i)-(iv) defined above, the risk of each corresponding test is at most $2\xi$.*

*Proof.* See §7.8 for a detailed proof. $\square$

It is worth noting that, the above algorithms under the oracle model can be implemented using access to $\{\mathbf{x}_i\}_{i=1}^n$. In particular, instead of receiving response $z_q$ from the oracle for query $q$, using $n$ data points $\mathscr{A}$ can calculate $1/n \cdot \sum_{i=1}^n q(\mathbf{x}_i)$ as a replacement of $z_q$. By Bernstein's inequality and union bound, $1/n \cdot \sum_{i=1}^n q(\mathbf{x}_i)$ has the same tail behavior as $z_q$. Thus the same test based on $\mathscr{A}$ has small risk when $\mathscr{A}$ is given access to $\{\mathbf{x}_i\}_{i=1}^n$ instead of the oracle.

It is necessary to ensure that algorithms under the oracle model is implementable given $\{\mathbf{x}_i\}_{i=1}^n$. Otherwise, the oracle model may fail to faithfully characterize practical algorithms and capture the difficulty of underlying statistical problems. For example, suppose that in (3.1) of Definition 3.1, we set $\xi = 0$ and

$$\tau_q = \max\{2b/3 \cdot 1/n, \sqrt{2\operatorname{Var}[q(\boldsymbol{X})]/n}\}. \tag{5.12}$$

By following the same proof of Theorem 5.5, we can show that the risk of each of the aforementioned tests is exactly zero for setting (iv).(a), even if we replace $n$ with $n' = n/\log\binom{d}{s^*}$. However, in §5.1.4 we will show the test in (5.9) matches the information-theoretical lower bound in the original setting, i.e., with $n$ replaced by $n'$, the test in (5.9) violates the information-theoretical lower bound, which suggests, with sample size $n'$, any test should be asymptotically powerless. This is because the oracle model that satisfies $\xi = 0$ and (5.12) is unrealistic, in the sense that it is not implementable provided access to $\{\mathbf{x}_i\}_{i=1}^n$. In another word, the oracle provides much more information than the information $\{\mathbf{x}_i\}_{i=1}^n$ can provide, which makes the algorithm unrealistically powerful. This justifies our definition of oracle model, which explicitly incorporates query space capacity $\eta(\mathcal{Q}_\mathscr{A})$ as well as tail probability $\xi$ in comparison with the previous definition of oracle model (Feldman et al., 2013, 2015).

### 5.1.4 Computational and Statistical Phase Transition

To illustrate the phase transition, we define

$$p_{s^*} = \log s^*/\log d, \quad p_{\beta^*} = \log(1/\beta^*)/\log d, \quad p_n = \log n/\log d, \quad p_\alpha = \log(1/\alpha)/\log d. \tag{5.13}$$

That is to say, we denote $s^*$, $\beta^*$, $n$, and $\alpha$ as the (inverse) polynomial of $d$, e.g., $s^* = d^{p_{s^*}}$. We ignore the $\log(\log d)$ factors, the sufficiently small positive constant $\delta$, as well as $\log(1/\xi)$, for the simplicity of discussion. Then the lower bounds in §5.1.1 and §5.1.2 translate to the following.

(i) For $(p_n - 2p_\alpha)_+ + (2p_{s^*} - 1)_+ - 2p_{\beta^*} < 0$, any test based on an algorithm that has polynomial oracle complexity is asymptotically powerless;

(ii) For $(p_n - 2p_\alpha)_+ + (2p_{s^*} - 1)_+ - 2p_{\beta^*} < 0$, any test is asymptotically powerless if $p_{s^*} - 2p_{\beta^*} < 0$ or $p_n - p_\alpha - 2p_{\beta^*} < 0$.

Here $(a)_+$ is defined as $a \cdot \mathbb{1}(a > 0)$. Meanwhile, the upper bounds in §5.1.3 translate to the following two settings, which correspond to settings (i) and (ii) above.

(i) For $(p_n - 2p_\alpha)_+ + (2p_{s^*} - 1)_+ - 2p_{\beta^*} > 0$, there is a test based on an algorithm with polynomial oracle complexity that successfully distinguishes $H_0$ from $H_1$;

(ii) For $p_{s^*} - 2p_{\beta^*} > 0$ and $p_n - p_\alpha - 2p_{\beta^*} > 0$, there exists a test based upon an algorithm, which has exponential oracle complexity, that successfully distinguishes $H_0$ from $H_1$.

For setting (i), the lower bound subject to the computational constraint is attained by the algorithms under settings (i)-(iii) in §5.1.3. For setting (ii), the information-theoretical lower bound is achieved by the algorithm in setting (iv) in §5.1.3. Therefore, our computational and statistical phase transition is nearly sharp.

## 5.2 Perfect Matching Detection

In the following, we consider the sparse set detection problem in §2, where $\mathcal{C}$ is the set of all perfect matchings of a complete balanced bipartite graph, which has $2\sqrt{d}$ nodes. We first establish the lower bound on oracle complexity.

### 5.2.1 Computational Lower Bound

The next lemma quantifies $\sup_{q \in \mathcal{Q}} |\mathcal{C}(q)|$ based upon the combinatorial characterization in Theorem 4.4. We use the notation defined in (5.1).

**Lemma 5.6.** We assume that there exists a sufficiently small constant $\delta > 0$ such that

$$\log(1 + \tau^2/\alpha^2)/\beta^{*2} \geq 3d^\delta/2 + 1.$$

Then we have

$$\sup_{q \in \mathcal{Q}} |\mathcal{C}(q)| \leq 2\exp(-\delta \log d \cdot 3d^\delta/8)|\mathcal{C}|.$$

*Proof.* See §7.9 for a detailed proof. □

The following theorem quantifies the computational lower bound. In detail, we provide sufficient conditions under which $T \cdot \sup_{q \in \mathcal{Q}} |\mathcal{C}(q)|/|\mathcal{C}| = o(1)$ when $T$ is polynomial in $d$. Let $\delta$ be a sufficiently small positive constant. We consider the following settings.

(i) $\tau^2/\alpha^2 = o(1)$ and $\tau^2/(2d^\delta \alpha^2 \beta^{*2}) \to \infty$;

(ii) $\lim_{d \to \infty} \tau^2/\alpha^2 > 0$ and $\beta^* = o(d^{-\delta})$.

**Theorem 5.7.** For $T = O(d^\eta)$, where $\eta$ is any constant and $T \geq 1$, we have $T \cdot \sup_{q \in \mathcal{Q}} |\mathcal{C}(q)|/|\mathcal{C}| = o(1)$ under setting (i) or (ii) defined above.

*Proof.* See §7.10 for a detailed proof. □

Combining Theorems 5.7 and 4.2, we then conclude that, under setting (i) or (ii) defined above, any test based on an algorithm with polynomial oracle complexity is asymptotically powerless.

### 5.2.2 Information-theoretical Lower Bound

The following corollary of Proposition 5.4 establishes the information-theoretical lower bound.

**Corollary 5.8.** We consider (i) $\beta^{*2}\alpha^2 n = o(1)$ and $\beta^{*2}s^* = o(1)$; (ii) $\beta^* = o(1)$ and $\beta^{*2}\alpha n = o(1)$. For $\delta$ being a sufficiently small positive constant, under setting (i) or (ii) we have $\inf_\phi R(\phi) \geq 1 - \delta$ for $s^*, d, n$ sufficiently large.

*Proof.* Recall that for the perfect matching problem we have $s^* = \sqrt{d}$. Following the same proof of Proposition 5.4, we obtain the conclusion. It is worth noting that, in the proof of Proposition 5.4 we use the negative association property of sparse set, which also holds for perfect matching. See, e.g., Addario-Berry et al. (2010) for details. □

### 5.2.3 Upper Bounds

We consider the same algorithms and tests as in §5.1.3. For the detection of perfect matching, recall that $s^{*2} = d$. Therefore, $\beta^{*2}s^{*2}n\alpha^2/d = \beta^{*2}n\alpha^2 \to \infty$ in setting (ii) is implied by $\beta^{*2}n\alpha^2/\log d \to \infty$ in setting (iii). Hence, the tests for settings (i) and (iii) can be directly applied. For setting (iv), we can directly use the two tests that search over all $|\mathcal{S}_t| = s^*$ or $|\mathcal{S}_t| = \bar{s}^*$. Alternatively we can reduce the computational cost of the test under setting (iv).(a) by restricting the exhaustive search to all perfect matchings. Such a modification reduces $T$ from $\binom{d}{s^*}$ to $s^*!$. Then the capacity of query space reduces from $\log \binom{d}{s^*}$ to $\log(s^*!)$. By Stirling's approximation, we have that $\log \binom{d}{s^*}$ and $\log(s^*!)$ are roughly of the same order, since we have $s^* = \sqrt{d}$. Hence, the resulting improvement on the scaling of $\beta^*$ is only up to constants.

### 5.2.4 Computational and Statistical Phase Transition

Combining §5.2.1-§5.2.3, we obtain the following phase transition. In particular, using the notation in (5.13), the lower bounds in §5.2.1-§5.2.2 translate to the following two settings if we ignore $\log(\log d)$ factors and the sufficiently small constant $\delta > 0$.

(i) For $(p_n - 2p_\alpha)_+ - 2p_{\beta^*} < 0$, any test based on an algorithm with polynomial oracle complexity is asymptotically powerless;

(ii) For $(p_n - 2p_\alpha)_+ - 2p_{\beta^*} < 0$, any test is asymptotically powerless if we have $p_n - p_\alpha - 2p_{\beta^*} < 0$ or $p_{s^*} - 2p_{\beta^*} < 0$.

The upper bounds in §5.2.3 translate to the following two settings.

(i) For $(p_n - 2p_\alpha)_+ - 2p_{\beta^*} > 0$, there is a test based on an algorithm, which has polynomial oracle complexity, that successfully distinguishes $H_0$ from $H_1$;

(ii) For $p_{s^*} - 2p_{\beta^*} > 0$ and $p_n - p_\alpha - 2p_{\beta^*} > 0$, there exists a test based upon an algorithm, which has exponential oracle complexity, that successfully distinguishes $H_0$ from $H_1$.

The phase transition of perfect matching detection is almost the same as that of sparse set detection, except that it does not involve $(2p_{s^*} - 1)_+$. This is because we have $s^* = \sqrt{d}$, which by (5.13) implies $2p_{s^*} - 1 = 0$. Recall that by Theorem 4.4, the characterization of computational lower bound reduces to a combinatorial quantity that involves the overlaps between two elements uniformly drawn from $\mathcal{C}$. The similarity of computational phase transition between perfect matching and sparse set detection suggests that perfect matching and sparse set shares similar aforementioned combinatorial properties that involve overlapping pairs.

One byproduct of our computational and statistical phase transition is a lower bound for matrix permanent problems. We first consider $\alpha = 1$. As shown in Addario-Berry et al. (2010), the optimal test that matches the information-theoretical lower bound is the likelihood ratio test, which involves calculating the likelihood ratio

$$L(\mathbf{X}) = \frac{1/|\mathcal{C}| \cdot \sum_{\mathcal{S} \in \mathcal{C}} \mathrm{d}\mathbb{P}_\mathcal{S}^n}{\mathrm{d}\mathbb{P}_0^n}(\mathbf{x}_1, \ldots, \mathbf{x}_n) = 1/|\mathcal{C}| \cdot \sum_{\mathcal{S} \in \mathcal{C}} \exp\bigl(\beta^* \sum_{j \in \mathcal{S}} \sum_{i=1}^n x_{i,j} - s^* \beta^{*2} n/2\bigr). \quad (5.14)$$

Here $\mathbf{X} \in \mathbb{R}^{n \times d}$ is the data matrix and $\{\mathbf{x}_i\}_{i=1}^n$ are the data points, while $\mathbb{P}_0^n$ and $\mathbb{P}_\mathcal{S}^n$ are the product distributions of $\mathbb{P}_0$ and $\mathbb{P}_\mathcal{S}$. We define $\mathbf{M} \in \mathbb{R}^{\sqrt{d} \times \sqrt{d}}$ to be a matrix whose $(k, k')$-th entry is

$$M_{k,k'} = \sum_{i=1}^n x_{i,j}, \quad \text{where } j = (k-1)\sqrt{d} + k'. \tag{5.15}$$

Recall that $s^* = \sqrt{d}$. We denote by $\sigma$ the permutation over $[s^*]$. Then the right-hand side of (5.14) translates to

$$1/s^*! \cdot \exp(-s^* \beta^{*2} n/2) \cdot \sum_\sigma \overbrace{\left[\prod_{j=1}^{s^*} \exp(\beta^* M_{j,\sigma(j)})\right]}^{(i)}.$$

Here the summation is over all valid $\sigma$'s. As shown in Jerrum et al. (2004), term (i) is the permanent of $\overline{\mathbf{M}} \in \mathbb{R}^{\sqrt{d} \times \sqrt{d}}$ in which $\overline{M}_{i,j} = \exp(\beta^* M_{i,j})$, and can be approximated within any arbitrarily small error in polynomial time by Markov chain Monte Carlo (MCMC). Meanwhile, according to Feldman et al. (2013, 2015), MCMC is captured by the oracle model. The phase transition established earlier in §5.2.4 aligns with such an observation. More specifically, for $\alpha = 1$, i.e., $p_\alpha = 0$, the computational and information-theoretical lower bounds are the same, and the information-theoretical lower bound can be attained by a polynomial time algorithm under the oracle model.

Now we consider general $\alpha \leq 1$, for which the likelihood ratio takes the form

$$L(\mathbf{X}) = \frac{1/|\mathcal{C}| \cdot \sum_{\mathcal{S} \in \mathcal{C}} \mathrm{d}\mathbb{P}_\mathcal{S}^n}{\mathrm{d}\mathbb{P}_0^n}(\mathbf{x}_1, \ldots, \mathbf{x}_n) = 1/|\mathcal{C}| \cdot \underbrace{\sum_{\mathcal{S} \in \mathcal{C}} \left\{\prod_{i=1}^n [\alpha \exp(\beta^* \sum_{j \in \mathcal{S}} x_{i,j} - s^* \beta^{*2}/2) + (1-\alpha)]\right\}}_{(i)}$$

$$= 1/|\mathcal{C}| \cdot \sum_{\mathcal{I} \subseteq [n]} \left[\alpha^{|\mathcal{I}|}(1-\alpha)^{n-|\mathcal{I}|} \exp(-s^* \beta^{*2} |\mathcal{I}|/2) \underbrace{\sum_{\mathcal{S} \in \mathcal{C}} \exp(\beta^* \sum_{j \in \mathcal{S}} \sum_{i \in \mathcal{I}} x_{i,j})}_{(ii)}\right]. \tag{5.16}$$

Similar to the previous setting in which $\alpha = 1$, term (ii) in (5.16) is equivalent to the permanent of a matrix. The only difference is that $\sum_{i=1}^n x_{i,j}$ in (5.15) is now replaced using $\sum_{i \in \mathcal{I}} x_{i,j}$. To calculate term (i) in (5.16), a straightforward way is to calculate term (ii) using MCMC, which however leads to exponential running time, because we have to traverse all $\mathcal{I} \subseteq [n]$. A natural question is whether we can calculate term (i) in (5.16) tractably utilizing the shared structure of term (ii) across different $\mathcal{I} \subseteq [n]$ for $\alpha \leq 1$, which leads to a strictly generalization of the matrix permanent problem.

Our computational lower bound suggests that such a generalization of matrix permanent problem can not be solved efficiently under certain conditions. In detail, since the likelihood ratio test matches the information-theoretical lower bound, for

$$(p_n - 2p_\alpha)_+ - 2p_{\beta^*} < 0, \quad p_n - p_\alpha - 2p_{\beta^*} \geq 0, \quad \text{and } p_{s^*} - 2p_{\beta^*} \geq 0, \tag{5.17}$$

the likelihood ratio test is asymptotically powerful. If any algorithm efficiently solves the generalized matrix permanent problem under the oracle model, we can use it to efficiently compute the likelihood ratio. However, our computational lower bound suggests this is impossible for (5.17). That is to say, under the oracle model, if (5.17) holds, then no algorithm is able to efficiently solve the generalized matrix permanent problem, i.e., approximate term (i) in (5.16) up to an arbitrarily small error. This then rules out MCMC for the aforementioned generalization of matrix permanent problems, since it can be captured by the oracle model (Feldman et al., 2013, 2015).

## 6 Implication for Sparse Principal Component Detection

In this section, we explore the computational and statistical phase transition for the sparse principal component detection problem defined in §2. In the sequel, we first verify Condition 4.3. The following lemma from Berthet and Rigollet (2013b) specifies the $h$ function in (4.4).

**Lemma 6.1** (Berthet and Rigollet (2013b)). *For $\mathcal{S}_1, \mathcal{S}_2 \in [d]$ with $|\mathcal{S}_1| = |\mathcal{S}_2| = s^*$ and $\beta^* \in (0,1)$, it holds that*

$$\mathbb{E}_{\mathbb{P}_0}\left[\frac{d\mathbb{P}_{\mathcal{S}_1}}{d\mathbb{P}_0}\frac{d\mathbb{P}_{\mathcal{S}_2}}{d\mathbb{P}_0}(\boldsymbol{X})\right] = \left(1 - \frac{\beta^{*2}|\mathcal{S}_1 \cap \mathcal{S}_2|^2}{s^{*2}}\right)^{-1/2}.$$

*Proof.* The proof is the same as Lemma 5.1 in Berthet and Rigollet (2013b). We provide a detailed proof in §7.11 for completeness. □

Besides, the classes of sparse sets satisfy the symmetricity assumption in Condition 4.3. Together with Lemma 5.1, we conclude that Theorem 4.4 holds, which forms the basis of our following results.

### 6.1 Computational Lower Bound

The next lemma establishes the upper bound of $\sup_{q \in \mathcal{Q}} |\mathcal{C}(q)|$ based on the combinatorial characterization in Theorem 4.4. For notational simplicity we define

$$\tau = \sqrt{\log(1/\xi)/n}, \quad \text{and} \quad \bar{\gamma} = \left[1 - \frac{(1+2s^*)\beta^{*2}}{s^{*2} - s^{*2}\beta^{*2}}\right]^{-1/2}. \tag{6.1}$$

**Lemma 6.2.** *Let $d/s^{*2} = d^{2\delta}$. For $2d^{-\delta}(\bar{\gamma} - 1) \leq \tau^2$, we have*

$$\sup_{q \in \mathcal{Q}} |\mathcal{C}(q)| \leq 2\exp\left[-\delta \log d \cdot \left(\frac{s^*}{\beta^*}\left\{1 - \left[\frac{1+\tau^2}{1+2d^{-\delta}(\bar{\gamma}-1)}\right]^{-2}\right\}^{1/2} - 1\right)\right]|\mathcal{C}|. \tag{6.2}$$

*Proof.* See §7.12 for a detailed proof. □

The next theorem establishes sufficient conditions under which $T \cdot \sup_{q \in \mathcal{Q}}|\mathcal{C}(q)|/|\mathcal{C}| = o(1)$ when $T$ is polynomial in $d$. Let $\delta > 0$ be a sufficiently small constant.

**Theorem 6.3.** *Let $s^* = d^{1/8}$ and $T = O(d^\eta)$ with $\eta$ being any constant and $T \geq 1$. Assuming that $\beta^*\sqrt{n/[s^{*2}\log(1/\xi)]} = o(1)$, we have $T \cdot \sup_{q \in \mathcal{Q}}|\mathcal{C}(q)|/|\mathcal{C}| = o(1)$.*

*Proof.* See §7.13 for a detailed proof. □

Combining Theorems 4.3 and 6.3, we conclude that under the condition of Theorem 6.3, any test based on an algorithm with polynomial oracle complexity is asymptotically powerless for $\xi = o(1)$.

## 6.2 Information-theoretical Lower Bound

In the sequel we present a corollary of the information-theoretical lower bound established by Berthet and Rigollet (2013b).

**Corollary 6.4.** For $\beta^*\sqrt{n/(s^*\log d)} = o(1)$, we have $\inf_\phi R(\phi) \geq 1 - \epsilon$ with $\epsilon = o(1)$.

*Proof.* See Berthet and Rigollet (2013b) for a detail proof. □

## 6.3 Upper Bounds

In the sequel, we construct upper bounds under the oracle model. We consider the following settings.

(i) $\beta^{*2} n/[s^{*2} \log(d/\xi)] \to \infty$. We consider an algorithm $\mathscr{A}$, which employs the following sequence of query functions,

$$q_t(\boldsymbol{X}) = \mathbb{1}(X_t^2 \geq 1 + \beta^*/s^*), \tag{6.3}$$

where $t \in [T]$ and $T = d$. In other words, the query space $\mathcal{Q}_\mathscr{A}$ of $\mathscr{A}$ is discrete with $|\mathcal{Q}_\mathscr{A}| = d$, i.e., $\eta(\mathcal{Q}_\mathscr{A}) = \log d$. We define the test as

$$\mathbb{1}\{\sup_{t\in[T]} z_{q_t} \geq 2[1 - \Phi(\sqrt{1+\beta^*/s^*})] + \beta^*/(8\pi s^*)\}, \tag{6.4}$$

where $z_{q_t}$ is the realization of $Z_{q_t}$ in Definition 3.1, and $\Phi$ is the Gaussian cumulative density function. In the above and following settings, $b$ in Definition 3.1 equals one.

(ii) $\beta^{*2} n/[s^* \log d + \log(1/\xi)] \to \infty$. We consider an algorithm $\mathscr{A}$ that uses the following sequence of query functions,

$$q_t(\boldsymbol{X}) = \mathbb{1}\big[1/s^* \cdot \big(\sum_{j\in\mathcal{S}_t} X_j\big)^2 \geq 1 + \beta^*\big], \tag{6.5}$$

where $t \in [T]$ with $T = \binom{d}{s^*}$ and $|\mathcal{S}_t| = s^*$, while $\bigcup_{t=1}^T \mathcal{S}_t = [d]$. In other words, the query space $\mathcal{Q}_\mathscr{A}$ of $\mathscr{A}$ is discrete with $|\mathcal{Q}_\mathscr{A}| = \binom{d}{s^*}$, i.e., $\eta(\mathcal{Q}_\mathscr{A}) = \log\binom{d}{s^*}$. We define the test as

$$\mathbb{1}\{\sup_{t\in[T]} z_{q_t} \geq 2[1 - \Phi(\sqrt{1+\beta^*})] + \beta^*/(8\pi)\}. \tag{6.6}$$

The following theorem indicates that the tests under settings (i) and (ii) are asymptotically powerful when $\xi = o(1)$.

**Theorem 6.5.** For settings (i) and (ii), the risk of each corresponding test is at most $2\xi$.

*Proof.* See §7.14 for a detailed proof. □

## 6.4 Computational and Statistical Phase Transition

For the simplicity of discussion, we ignore the $\log(1/\xi)$ factor. The lower bounds in §6.1-§6.2 translate to the following.

(i) For $\beta^* = o(s^*\sqrt{1/n})$, any test based on an algorithm that has polynomial oracle complexity is asymptotically powerless;

(ii) For $\beta^* = o(\sqrt{s^* \log d/n})$, any test is asymptotically powerless.

Meanwhile, the upper bounds in §6.3 translate to the following corresponding to settings (i) and (ii) specified above.

(i) For $\beta^*/\sqrt{s^{*2} \log d/n} \to \infty$, there is a test based on an algorithm, which has polynomial oracle complexity, that successfully distinguishes $H_0$ from $H_1$;

(ii) For $\beta^*/\sqrt{s^* \log d/n} \to \infty$, there exists a test based upon an algorithm, which has exponential oracle complexity, that successfully distinguishes $H_0$ from $H_1$.

For setting (i), there exists a $\log d$ gap between the computational lower bound and the upper bound achievable with polynomial oracle complexity. This gap can be closed by the covariance thresholding algorithm in Deshpande and Montanari (2014), which can be formulated into the oracle model. This suggests that the computational lower bound for $\beta^* = o(s^*\sqrt{1/n})$ is tight. However, for succinctness we only present the algorithm defined in (6.3) and (6.4), which is essentially a diagonal thresholding algorithm. In addition, it is worth mentioning that the algorithms in (6.3)-(6.6) can be implemented using $\{\mathbf{x}_i\}_{i=1}^n$. In detail, we can simulate the oracle by answering each query with the sample average of that query over $\{\mathbf{x}_i\}_{i=1}^n$. See the previous discussion in §5.1.3 for more details.

## 7 Proof of Theoretical Results

In this section, we first present the detailed proof of the theoretical results in §4-§6. Then we present a useful auxiliary lemma and its proof.

### 7.1 Proof of Theorem 4.2

*Proof.* Let $\mathcal{Q}^T$ be the $T$-th cartesian power of $\mathcal{Q}$, which is the family of all possible queries. For any $\mathscr{A} \in \mathcal{A}(T)$ that makes queries $\{q_t\}_{t=1}^T \in \mathcal{Q}^T$, we consider two cases:

(i) There exists a sequence $\{\mathcal{S}^t\}_{t=1}^T$, which satisfies $\mathcal{S}^t \in \mathcal{C}(q_t)$ and $\mathcal{S}^t \notin \bigcup_{t' \neq t} \mathcal{C}(q_{t'})$ for all $t \in [T]$;

(ii) There does not exist such a sequence.

Here $\mathcal{C}(q_t)$ and $\mathcal{C}(q_{t'})$ are as defined in Definition 4.1.

For case (i), we construct the worst-case oracle $r \in \mathcal{R}[\xi, n, \eta(\mathcal{Q}_\mathscr{A})]$ as follows. As in Definition 3.1, $r$ returns a realization of $Z_{q_t}$ at the $t$-th iteration. Let $\mathcal{S}_1, \ldots, \mathcal{S}_m$ be the elements of $\mathcal{C} \setminus \bigcup_{t \in [T]} \mathcal{C}(q_t)$, where $m = |\mathcal{C} \setminus \bigcup_{t \in [T]} \mathcal{C}(q_t)|$. Under the null hypothesis, we set

$$\bar{\mathbb{P}}_0\big(\{Z_{q_t}\}_{t=1}^T = \{\mathbb{E}_{\mathbb{P}_\mathcal{S}}[q_t(\boldsymbol{X})]\}_{t=1}^T\big) = 2\xi/T, \quad \text{for} \ \ \mathcal{S} \in \{\mathcal{S}^t\}_{t=1}^T, \tag{7.1}$$

and

$$\bar{\mathbb{P}}_0\big(\{Z_{q_t}\}_{t=1}^T = \{\mathbb{E}_{\mathbb{P}_\mathcal{S}}[q_t(\boldsymbol{X})]\}_{t=1}^T\big) = (1-2\xi)/m, \quad \text{for} \ \ \mathcal{S} \in \{\mathcal{S}_j\}_{j=1}^m. \tag{7.2}$$

Under the alternative hypothesis, we set

$$\bar{\mathbb{P}}_\mathcal{S}\big(\{Z_{q_t}\}_{t=1}^T = \{\mathbb{E}_{\mathbb{P}_\mathcal{S}}[q_t(\boldsymbol{X})]\}_{t=1}^T\big) = 1, \quad \text{for} \ \ \mathcal{S} \in \mathcal{C}, \tag{7.3}$$

where two sequences are equal if and only if their entries are pairwise equal. Recall that $\mathcal{Q}_\mathscr{A} \subseteq \mathcal{Q}$ is the query space of $\mathscr{A}$. For any $q \in \mathcal{Q}_\mathscr{A} \setminus \{q_t\}_{t=1}^T$, we set

$$\bar{\mathbb{P}}_0\{Z_q = \mathbb{E}_{\mathbb{P}_0}[q(\boldsymbol{X})]\} = 1, \quad \text{and} \quad \bar{\mathbb{P}}_\mathcal{S}\{Z_q = \mathbb{E}_{\mathbb{P}_\mathcal{S}}[q(\boldsymbol{X})]\} = 1, \quad \text{for} \ \ \mathcal{S} \in \mathcal{C}. \tag{7.4}$$

Note that by the definition of $\mathcal{C}(q)$ in (4.1) and Definition 3.1, such a construction of oracle is valid, i.e., $\bar{r} \in \mathcal{R}[\xi, n, \eta(\mathcal{Q}_\mathscr{A})]$. To see this, first note that for any $\mathcal{S} \in \{\mathcal{S}_j\}_{j=1}^m$, we have

$$|\mathbb{E}_{\mathbb{P}_\mathcal{S}}[q_t(\boldsymbol{X})] - \mathbb{E}_{\mathbb{P}_0}[q_t(\boldsymbol{X})]| \leq \bar{\tau}_{q_t}, \quad \text{for} \ \ t \in [T],$$

since for any $\mathcal{S} \in \{\mathcal{S}_j\}_{j=1}^m$ we have $\mathcal{S} \notin \bigcup_{t \in [T]} \mathcal{C}(q_t)$ by definition. Meanwhile, we have

$$|\mathbb{E}_{\mathbb{P}_\mathcal{S}}[q_t(\boldsymbol{X})] - \mathbb{E}_{\mathbb{P}_0}[q_t(\boldsymbol{X})]| \leq \bar{\tau}_{q_t}, \quad \text{for} \ \ t \in [T] \ \text{and} \ \mathcal{S} \in \{\mathcal{S}^{t'}\}_{t'=1}^T \setminus \mathcal{S}^t,$$

since by the assumption of case (i) we have $\mathcal{S}^{t'} \notin \mathcal{C}(q_t)$ for all $t' \neq t$. By (7.1) and (7.2) we have that

$$\bar{\mathbb{P}}_0\{|Z_{q_t} - \mathbb{E}[q_t(\boldsymbol{X})]| \leq \bar{\tau}_{q_t}\} = m \cdot (1-2\xi)/m + (T-1) \cdot 2\xi/T = 1 - 2\xi/T, \quad \text{for} \ \ t \in [T],$$

which together with (7.4) and union bound implies

$$\bar{\mathbb{P}}_0\big(\bigcap_{q \in \mathcal{Q}_\mathscr{A}}\{|Z_q - \mathbb{E}_{\mathbb{P}_0}[q(\boldsymbol{X})]| \leq \tau_q\}\big) \geq \bar{\mathbb{P}}_0\big(\bigcap_{q \in \mathcal{Q}_\mathscr{A}}\{|Z_q - \mathbb{E}_{\mathbb{P}_0}[q(\boldsymbol{X})]| \leq \bar{\tau}_q\}\big) \geq 1 - 2\xi.$$

Here the first inequality holds because $\bar{\tau}_q$ is defined by setting $\eta(\mathcal{Q}_\mathscr{A}) = 0$ in $\tau_q$ as in Definition 4.1. Hence, (3.1) holds under the null hypothesis. In addition, under the alternative hypothesis, (3.1) is also satisfied according to (7.3) and (7.4).

Now we consider the behavior of test $\phi$ under case (i). For the response sequence $\{\mathbb{E}_{\mathbb{P}_\mathcal{S}}[q_t(\boldsymbol{X})]\}_{t=1}^T$, we assume $\phi$ accepts the null hypothesis for $k_1$ elements of $\{\mathcal{S}^t\}_{t=1}^T$ and rejects it for the rest $T - k_1$ elements. Meanwhile, we assume $\phi$ accepts the null hypothesis for $k_2$ elements of $\{\mathcal{S}_j\}_{j=1}^m$ and rejects it for the rest $m - k_2$ elements. Besides, for $\mathcal{S} \in \bigcup_{t \in [T]} \mathcal{C}(q_t) \setminus \{\mathcal{S}^t\}_{t=1}^T$, we assume that $\phi$ rejects the null hypothesis, since otherwise the type-II error increases, while the type-I error remains the same. Then from (7.1)-(7.3) we have

$$\bar{R}(\phi) = \bar{\mathbb{P}}_0(\phi = 1) + \frac{1}{|\mathcal{C}|} \sum_{\mathcal{S} \in \mathcal{C}} \bar{\mathbb{P}}_\mathcal{S}(\phi = 0) = 2\xi/T \cdot (T - k_1) + (1-2\xi)/m \cdot (m - k_2) + (k_1 + k_2)/|\mathcal{C}|$$

$$\geq \min_{k_1 \in \{0,\ldots,T\}}\{2\xi/T \cdot (T - k_1) + k_1/|\mathcal{C}|\} + \min_{k_2 \in \{0,\ldots,m\}}\{(1-2\xi)/m \cdot (m - k_2) + k_2/|\mathcal{C}|\}$$

$$= \min\{T/|\mathcal{C}|, 2\xi\} + \min\{m/|\mathcal{C}|, 1 - 2\xi\} \geq \min\{(T+m)/|\mathcal{C}|, 2\xi + m/|\mathcal{C}|, T/|\mathcal{C}| + 1 - 2\xi, 1\}$$

$$\geq \min\bigg\{1 - \frac{T \cdot [\sup_{q \in \mathcal{Q}}|\mathcal{C}(q)| - 1]}{|\mathcal{C}|}, 1 - \frac{T \cdot \sup_{q \in \mathcal{Q}}|\mathcal{C}(q)|}{|\mathcal{C}|} + 2\xi, T/|\mathcal{C}| + 1 - 2\xi, 1\bigg\},$$

where the last inequality is from $m = |\mathcal{C} \setminus \bigcup_{t \in [T]} \mathcal{C}(q_t)| \geq |\mathcal{C}| - T \cdot \sup_{q \in \mathcal{Q}} |\mathcal{C}(q)|$. Hence, for case (i) we have that, for any algorithm $\mathscr{A} \in \mathcal{A}(T)$, there exists an oracle $r \in \mathcal{R}[\xi, n, \eta(\mathcal{Q}_\mathscr{A})]$ such that

$$\inf_{\phi \in \mathcal{H}(\mathscr{A}, r)} \overline{R}(\phi) \geq \min\left\{1 - \frac{T \cdot \sup_{q \in \mathcal{Q}} |\mathcal{C}(q)|}{|\mathcal{C}|} + \min\{2\xi, T/|\mathcal{C}|\}, T/|\mathcal{C}| + 1 - 2\xi, 1\right\}. \tag{7.5}$$

For case (ii), we construct another worst-case oracle $r \in \mathcal{R}[\xi, n, \eta(\mathcal{Q}_\mathscr{A})]$ as follows. Under the null hypothesis we set

$$\overline{\mathbb{P}}_0\big(\{Z_{q_t}\}_{t=1}^T = \{\mathbb{E}_{\mathbb{P}_\mathcal{S}}[q_t(\boldsymbol{X})]\}_{t=1}^T\big) = 1/m, \quad \text{for } \mathcal{S} \in \{\mathcal{S}_j\}_{j=1}^m, \tag{7.6}$$

where $\{\mathcal{S}_j\}_{j=1}^m$ is defined in the same way as in case (i). Under the alternative hypothesis, we set

$$\overline{\mathbb{P}}_\mathcal{S}\big(\{Z_{q_t}\}_{t=1}^T = \{\mathbb{E}_{\mathbb{P}_\mathcal{S}}[q_t(\boldsymbol{X})]\}_{t=1}^T\big) = 1, \quad \text{for } \mathcal{S} \in \mathcal{C}. \tag{7.7}$$

For any $q \in \mathcal{Q}_\mathscr{A} \setminus \{q_t\}_{t=1}^T$, we set

$$\overline{\mathbb{P}}_0\{Z_q = \mathbb{E}_{\mathbb{P}_0}[q(\boldsymbol{X})]\} = 1, \quad \text{and} \quad \overline{\mathbb{P}}_\mathcal{S}\{Z_q = \mathbb{E}_{\mathbb{P}_\mathcal{S}}[q(\boldsymbol{X})]\} = 1, \quad \text{for } \mathcal{S} \in \mathcal{C}.$$

By (4.1) and Definition 3.1, this construction of oracle is valid, i.e., $r \in \mathcal{R}[\xi, n, \eta(\mathcal{Q}_\mathscr{A})]$. To see this, note that for any $\mathcal{S} \in \{\mathcal{S}_j\}_{j=1}^m$, by Definition 4.1 it holds that

$$|\mathbb{E}_{\mathbb{P}_\mathcal{S}}[q_t(\boldsymbol{X})] - \mathbb{E}_{\mathbb{P}_0}[q_t(\boldsymbol{X})]| \leq \overline{\tau}_{q_t}.$$

Then by the same argument as in case (i), we obtain $r \in \mathcal{R}[\xi, n, \eta(\mathcal{Q}_\mathscr{A})]$.

Now we consider the behavior of test $\phi$ under case (ii). For the response sequence $\{\mathbb{E}_{\mathbb{P}_\mathcal{S}}[q_t(\boldsymbol{X})]\}_{t=1}^T$, we assume $\phi$ accepts the null hypothesis for $k_3$ elements of $\{\mathcal{S}_j\}_{j=1}^m$ and rejects it for the rest $m - k_3$ elements. Besides, for $\mathcal{S} \in \bigcup_{t \in [T]} \mathcal{C}(q_t)$, we assume that $\phi$ rejects the null hypothesis, since otherwise the type-II error increases, while the type-I error remains the same. From (7.6) and (7.7) we have

$$\overline{R}(\phi) = \overline{\mathbb{P}}_0(\phi = 1) + \frac{1}{|\mathcal{C}|} \sum_{\mathcal{S} \in \mathcal{C}} \overline{\mathbb{P}}_\mathcal{S}(\phi = 0) = (m - k_3)/m + k_3/|\mathcal{C}| \geq \min\{m/|\mathcal{C}|, 1\}$$

$$\geq \min\left\{1 - \frac{(T-1) \cdot \sup_{q \in \mathcal{Q}} |\mathcal{C}(q)|}{|\mathcal{C}|}, 1\right\}, \tag{7.8}$$

where the first inequality is obtained by minimizing over $k_3 \in \{0, \ldots, m\}$. Here the last inequality is from the definition that $m = |\mathcal{C} \setminus \bigcup_{t \in [T]} \mathcal{C}(q_t)|$ and

$$\big|\bigcup_{t \in [T]} \mathcal{C}(q_t)\big| \leq (T-1) \cdot \sup_{q \in \mathcal{Q}} |\mathcal{C}(q)|,$$

which follows from proof by contradiction using the assumption of case (ii). In (7.8) the last equality is from Definition 4.1. Therefore, for case (ii) we have that, for any algorithm $\mathscr{A} \in \mathcal{A}(T)$, there exists an oracle $r \in \mathcal{R}[\xi, n, \eta(\mathcal{Q}_\mathscr{A})]$ such that

$$\inf_{\phi \in \mathcal{H}(\mathscr{A}, r)} \overline{R}(\phi) \geq \min\left\{1 - \frac{(T-1) \cdot \sup_{q \in \mathcal{Q}} |\mathcal{C}(q)|}{|\mathcal{C}|}, 1\right\}. \tag{7.9}$$

Finally, combining (7.5) for case (i) and (7.9) for case (ii), we have that for any $\mathscr{A} \in \mathcal{A}(T)$, there exists an oracle $r \in \mathcal{R}[\xi, n, \eta(\mathcal{Q}_{\mathscr{A}})]$ such that

$$\inf_{\phi \in \mathcal{H}(\mathscr{A}, r)} \overline{R}(\phi) \geq \min\left\{1 - \frac{T \cdot \sup_{q \in \mathcal{Q}}|\mathcal{C}(q)|}{|\mathcal{C}|} + \min\left\{2\xi, T/|\mathcal{C}|, \sup_{q \in \mathcal{Q}} |\mathcal{C}(q)|/|\mathcal{C}|\right\}, T/|\mathcal{C}| + 1 - 2\xi, 1\right\}.$$

Thus, we conclude the proof of Theorem 4.2. $\square$

## 7.2 Proof of Lemma 4.5

*Proof.* By the definition of distinguishable distribution in Definition 4.1, we have

$$|\mathbb{E}_{\mathbb{P}_{\mathcal{S}}}[q(\boldsymbol{X})] - \mathbb{E}_{\mathbb{P}_0}[q(\boldsymbol{X})]| > \bar{\tau}_q$$

for any query $q$ and $\mathcal{S} \in \mathcal{C}(q)$, which is defined in (4.1). Let $\mathbb{Q}$ be the uniform prior distribution over $\mathcal{C}(q)$. For notational simplicity, we define $\bar{q}(\mathbf{x}) = q(\mathbf{x}) - \mathbb{E}_{\mathbb{P}_0}[q(\boldsymbol{X})]$. Then we have

$$\bar{\tau}_q < \mathbb{E}_{\mathcal{S} \sim \mathbb{Q}}\{|\mathbb{E}_{\mathbb{P}_{\mathcal{S}}}[q(\boldsymbol{X})] - \mathbb{E}_{\mathbb{P}_0}[q(\boldsymbol{X})]|\} = \mathbb{E}_{\mathcal{S} \sim \mathbb{Q}}\left[\left|\int \bar{q}(\mathbf{x}) d\mathbb{P}_{\mathcal{S}}(\mathbf{x}) - \int \bar{q}(\mathbf{x}) d\mathbb{P}_0(\mathbf{x})\right|\right]$$

$$= \mathbb{E}_{\mathcal{S} \sim \mathbb{Q}}\left\{\left|\int \bar{q}(\mathbf{x})\left[\frac{d\mathbb{P}_{\mathcal{S}}}{d\mathbb{P}_0}(\mathbf{x}) - 1\right] d\mathbb{P}_0(\mathbf{x})\right|\right\} = \mathbb{E}_{\mathcal{S} \sim \mathbb{Q}}\left(\left|\mathbb{E}_{\mathbb{P}_0}\left\{\bar{q}(\boldsymbol{X})\left[\frac{d\mathbb{P}_{\mathcal{S}}}{d\mathbb{P}_0}(\boldsymbol{X}) - 1\right]\right\}\right|\right)$$

$$= \mathbb{E}_{\mathcal{S} \sim \mathbb{Q}}\left[\mathbb{E}_{\mathbb{P}_0}\left\{\bar{q}(\boldsymbol{X})\left[\frac{d\mathbb{P}_{\mathcal{S}}}{d\mathbb{P}_0}(\boldsymbol{X}) - 1\right]\right\} \cdot \text{sign}\left(\mathbb{E}_{\mathbb{P}_0}\left\{\bar{q}(\boldsymbol{X})\left[\frac{d\mathbb{P}_{\mathcal{S}}}{d\mathbb{P}_0}(\boldsymbol{X}) - 1\right]\right\}\right)\right]$$

$$= \mathbb{E}_{\mathbb{P}_0}\left[\bar{q}(\boldsymbol{X}) \cdot \mathbb{E}_{\mathcal{S} \sim \mathbb{Q}}\left\{\left[\frac{d\mathbb{P}_{\mathcal{S}}}{d\mathbb{P}_0}(\boldsymbol{X}) - 1\right] \cdot \text{sign}\left(\mathbb{E}_{\mathbb{P}_0}\left\{\bar{q}(\boldsymbol{X})\left[\frac{d\mathbb{P}_{\mathcal{S}}}{d\mathbb{P}_0}(\boldsymbol{X}) - 1\right]\right\}\right)\right\}\right]. \quad (7.10)$$

Meanwhile, by Cauchy-Schwarz inequality we have

$$\mathbb{E}_{\mathbb{P}_0}\left[\bar{q}(\boldsymbol{X}) \cdot \mathbb{E}_{\mathcal{S} \sim \mathbb{Q}}\left\{\left[\frac{d\mathbb{P}_{\mathcal{S}}}{d\mathbb{P}_0}(\boldsymbol{X}) - 1\right] \cdot \text{sign}\left(\mathbb{E}_{\mathbb{P}_0}\left\{\bar{q}(\boldsymbol{X})\left[\frac{d\mathbb{P}_{\mathcal{S}}}{d\mathbb{P}_0}(\boldsymbol{X}) - 1\right]\right\}\right)\right\}\right] \quad (7.11)$$

$$\leq \underbrace{\mathbb{E}_{\mathbb{P}_0}[\bar{q}(\boldsymbol{X})^2]^{1/2}}_{(i)} \cdot \underbrace{\mathbb{E}_{\mathbb{P}_0}\left(\mathbb{E}_{\mathcal{S} \sim \mathbb{Q}}\left\{\left[\frac{d\mathbb{P}_{\mathcal{S}}}{d\mathbb{P}_0}(\boldsymbol{X}) - 1\right] \cdot \text{sign}\left(\mathbb{E}_{\mathbb{P}_0}\left\{\bar{q}(\boldsymbol{X})\left[\frac{d\mathbb{P}_{\mathcal{S}}}{d\mathbb{P}_0}(\boldsymbol{X}) - 1\right]\right\}\right)\right\}^2\right)^{1/2}}_{(ii)}.$$

For term (i), by the definition that $\bar{q}(\mathbf{x}) = q(\mathbf{x}) - \mathbb{E}_{\mathbb{P}_0}[q(\boldsymbol{X})]$, we have

$$\mathbb{E}_{\mathbb{P}_0}[\bar{q}(\boldsymbol{X})^2]^{1/2} = \sqrt{\text{Var}[q(\boldsymbol{X})]}, \quad (7.12)$$

where the variance is taken under $\mathbb{P}_0$. For notational simplicity, we define

$$z(\mathcal{S}) = \text{sign}\left(\mathbb{E}_{\mathbb{P}_0}\left\{\bar{q}(\boldsymbol{X})\left[\frac{d\mathbb{P}_{\mathcal{S}}}{d\mathbb{P}_0}(\boldsymbol{X}) - 1\right]\right\}\right) \in \{-1, 1\}.$$

For term (ii), by the definition of $\mathbb{Q}$ we have

$$\mathbb{E}_{\mathbb{P}_0}\left(\mathbb{E}_{\mathcal{S}\sim\mathbb{Q}}\left\{\left[\frac{d\mathbb{P}_\mathcal{S}}{d\mathbb{P}_0}(\boldsymbol{X})-1\right]\cdot z(\mathcal{S})\right\}^2\right)$$

$$=\int z(\mathcal{S}_1)z(\mathcal{S}_2)\cdot\mathbb{E}_{\mathbb{P}_0}\left\{\left[\frac{d\mathbb{P}_{\mathcal{S}_1}}{d\mathbb{P}_0}(\boldsymbol{X})-1\right]\cdot\left[\frac{d\mathbb{P}_{\mathcal{S}_2}}{d\mathbb{P}_0}(\boldsymbol{X})-1\right]\right\}d\mathbb{Q}(\mathcal{S}_1)d\mathbb{Q}(\mathcal{S}_2)$$

$$\leq\int\left|\mathbb{E}_{\mathbb{P}_0}\left\{\left[\frac{d\mathbb{P}_{\mathcal{S}_1}}{d\mathbb{P}_0}(\boldsymbol{X})-1\right]\cdot\left[\frac{d\mathbb{P}_{\mathcal{S}_2}}{d\mathbb{P}_0}(\boldsymbol{X})-1\right]\right\}\right|d\mathbb{Q}(\mathcal{S}_1)d\mathbb{Q}(\mathcal{S}_2)$$

$$=\int\left|\mathbb{E}_{\mathbb{P}_0}\left[\frac{d\mathbb{P}_{\mathcal{S}_1}}{d\mathbb{P}_0}\frac{d\mathbb{P}_{\mathcal{S}_2}}{d\mathbb{P}_0}(\boldsymbol{X})-1\right]\right|d\mathbb{Q}(\mathcal{S}_1)d\mathbb{Q}(\mathcal{S}_2) = \frac{1}{|\mathcal{C}(q)|^2}\sum_{\mathcal{S}_1,\mathcal{S}_2\in\mathcal{C}(q)}\left\{\mathbb{E}_{\mathbb{P}_0}\left[\frac{d\mathbb{P}_{\mathcal{S}_1}}{d\mathbb{P}_0}\frac{d\mathbb{P}_{\mathcal{S}_2}}{d\mathbb{P}_0}(\boldsymbol{X})\right]-1\right\}$$

$$=\mathbb{E}_{\mathbb{P}_0}\left\{\left[\frac{d\mathbb{P}_{\mathcal{S}\in\mathcal{C}(q)}}{d\mathbb{P}_0}(\boldsymbol{X})-1\right]^2\right\}=\chi^2\big(\mathbb{P}_{\mathcal{S}\in\mathcal{C}(q)},\mathbb{P}_0\big), \tag{7.13}$$

where $\mathbb{P}_{\mathcal{S}\in\mathcal{C}(q)}$ is defined in (4.7). Here the third last equality is from Condition 4.3 that

$$\mathbb{E}_{\mathbb{P}_0}\left[\frac{d\mathbb{P}_{\mathcal{S}_1}}{d\mathbb{P}_0}\frac{d\mathbb{P}_{\mathcal{S}_2}}{d\mathbb{P}_0}(\boldsymbol{X})\right]=h(|\mathcal{S}_1\cap\mathcal{S}_2|)\geq h(0)\geq 1$$

for all $\mathcal{S}_1,\mathcal{S}_2\in\mathcal{C}(q)\subseteq\mathcal{C}$. By Definition 3.1 we have

$$\bar{\tau}_q\geq\sqrt{\text{Var}[q(\boldsymbol{X})]\cdot\log(1/\xi)/n}, \tag{7.14}$$

where the variance is taken under $\mathbb{P}_0$. By plugging (7.12) and (7.13) into (7.11) and then into (7.10), together with (7.14), we obtain

$$\chi^2\big(\mathbb{P}_{\mathcal{S}\in\mathcal{C}(q)},\mathbb{P}_0\big)\geq\log(1/\xi)/n.$$

Thus we conclude the proof. $\square$

### 7.3  Proof of Theorem 4.4

*Proof.* Let $\mathbb{Q}_{\mathcal{C}(q)}$ be the uniform distribution over $\mathcal{C}(q)$. By the definition of $\chi^2$-divergence, we have

$$\chi^2\big(\mathbb{P}_{\mathcal{S}\in\mathcal{C}(q)},\mathbb{P}_0\big)=\mathbb{E}_{\mathbb{P}_0}\left\{\left[\frac{d\mathbb{P}_{\mathcal{S}\in\mathcal{C}(q)}}{d\mathbb{P}_0}(\boldsymbol{X})-1\right]^2\right\}=\frac{1}{|\mathcal{C}(q)|^2}\sum_{\mathcal{S}_1,\mathcal{S}_2\in\mathcal{C}(q)}\mathbb{E}_{\mathbb{P}_0}\left[\frac{d\mathbb{P}_{\mathcal{S}_1}}{d\mathbb{P}_0}\frac{d\mathbb{P}_{\mathcal{S}_2}}{d\mathbb{P}_0}(\boldsymbol{X})\right]-1$$

$$=\frac{1}{|\mathcal{C}(q)|^2}\sum_{\mathcal{S}_1,\mathcal{S}_2\in\mathcal{C}(q)}h(|\mathcal{S}_1\cap\mathcal{S}_2|)-1\leq\sup_{\mathcal{S}\in\mathcal{C}(q)}\left[\frac{1}{|\mathcal{C}(q)|}\sum_{\mathcal{S}'\in\mathcal{C}(q)}h(|\mathcal{S}\cap\mathcal{S}'|)\right]-1$$

$$=\sup_{\mathcal{S}\in\mathcal{C}(q)}\left\{\mathbb{E}_{\mathcal{S}'\sim\mathbb{Q}_{\mathcal{C}(q)}}[h(|\mathcal{S}\cap\mathcal{S}'|)]\right\}-1\leq\sup_{\mathcal{S}\in\mathcal{C}(q)}\left\{\mathbb{E}_{\mathcal{S}'\sim\mathbb{Q}_{\overline{\mathcal{C}}(q,\mathcal{S})}}[h(|\mathcal{S}\cap\mathcal{S}'|)]\right\}-1,$$

where the third equality is from Condition 4.3 and the last inequality is from (4.5). By Lemma 4.5, we obtain (4.6).

Next we prove the second claim that the left-hand side of (4.6) only depends on $|\mathcal{C}(q)|$, and the third claim that the left-hand side of (4.6) is a nonincreasing function of $|\mathcal{C}(q)|$. For $j\in\{0,\ldots,s^*\}$ we define

$$\mathcal{C}_j(\mathcal{S})=\{\mathcal{S}':|\mathcal{S}\cap\mathcal{S}'|=s^*-j,\ \mathcal{S}'\in\mathcal{C}\}.$$

Let $k(q, \mathcal{S}) \in \{0, \ldots, s^*\}$ be the smallest integer such that $\sum_{j=0}^{k(q,\mathcal{S})} |\mathcal{C}_j(\mathcal{S})| \geq |\mathcal{C}(q)|$. By the definition of $\bar{\mathcal{C}}(q, \mathcal{S})$ in (4.5), it holds that

$$\bar{\mathcal{C}}(q, \mathcal{S}) = \mathcal{C}_0(\mathcal{S}) \cup \mathcal{C}_1(\mathcal{S}) \cup \cdots \cup \mathcal{C}_{k(q,\mathcal{S})-1}(\mathcal{S}) \cup \mathcal{C}'_{k(q,\mathcal{S})}(\mathcal{S}).$$

To ensure $|\bar{\mathcal{C}}(q, \mathcal{S})| = |\mathcal{C}(q)|$, here $\mathcal{C}'_{k(q,\mathcal{S})}(\mathcal{S})$ is any subset of $\mathcal{C}_{k(q,\mathcal{S})}(\mathcal{S})$ with cardinality

$$|\mathcal{C}(q)| - \sum_{j=0}^{k(q,\mathcal{S})-1} |\mathcal{C}_j(\mathcal{S})|.$$

Let $\mathbb{Q}_\mathcal{C}$ be the uniform distribution over $\mathcal{C}$. For any $\mathcal{S} \in \mathcal{C}(q)$, we have

$$\frac{|\mathcal{C}(q)|}{|\mathcal{C}|} \cdot \mathbb{E}_{\mathcal{S}' \sim \mathbb{Q}_{\bar{\mathcal{C}}(q,\mathcal{S})}}[h(|\mathcal{S} \cap \mathcal{S}'|)] = \mathbb{E}_{\mathcal{S}' \sim \mathbb{Q}_\mathcal{C}}\{h(|\mathcal{S} \cap \mathcal{S}'|) \cdot \mathbb{1}[|\mathcal{S} \cap \mathcal{S}'| \geq s^* - k(q, \mathcal{S}) + 1]\} \quad (7.15)$$
$$+ \frac{|\mathcal{C}(q)| - \sum_{j=0}^{k(q,\mathcal{S})-1} |\mathcal{C}_j(\mathcal{S})|}{|\mathcal{C}|} \cdot h[s^* - k(q, \mathcal{S})].$$

According to the symmetricity assumption in Condition 4.3, $k(q, \mathcal{S})$ is the same across all $\mathcal{S} \in \mathcal{C}$ and only depends on $|\mathcal{C}(q)|$, since it can be equivalently defined as the the smallest integer such that

$$\mathbb{Q}_{\mathcal{S}' \sim \mathbb{Q}_\mathcal{C}}[|\mathcal{S} \cap \mathcal{S}'| \geq s^* - k(q, \mathcal{S})] \geq |\mathcal{C}(q)|/|\mathcal{C}|.$$

Similarly, the right-hand side of (7.15) is the same for all $\mathcal{S} \in \mathcal{C}$ and only depends on $|\mathcal{C}(q)|$. Therefore, we obtain the second claim. For any $\mathcal{S} \in \mathcal{C}$, when $|\mathcal{C}(q)|$ increases $\bar{\mathcal{C}}(q, \mathcal{S})$ encompasses more elements $\mathcal{S}'$ with smaller $|\mathcal{S} \cap \mathcal{S}'|$. Meanwhile, by Condition 4.3, $h$ is an increasing function, which implies the third claim. $\square$

### 7.4 Proof of Lemma 5.1

*Proof.* By the definition of $\mathbb{P}_{\mathcal{S}_1}$, $\mathbb{P}_{\mathcal{S}_2}$, and $\mathbb{P}_0$, we have

$$\frac{\mathrm{d}\mathbb{P}_{\mathcal{S}_1}}{\mathrm{d}\mathbb{P}_0}(\mathbf{x}) = \frac{\alpha \exp\left(\sum_{j \in [d]} x_j^2/2\right)}{\exp\left[\sum_{j \in \mathcal{S}_1}(x_j - \beta^*)^2/2 + \sum_{j \notin \mathcal{S}_1} x_j^2/2\right]} + (1 - \alpha)$$
$$= \alpha \exp\left(\sum_{j \in \mathcal{S}_1} \beta^* x_j - s^* \beta^{*2}/2\right) + (1 - \alpha),$$

and the same holds for $\mathrm{d}\mathbb{P}_{\mathcal{S}_2}/\mathrm{d}\mathbb{P}_0$. Therefore, we have

$$\mathbb{E}_{\mathbb{P}_0}\left[\frac{\mathrm{d}\mathbb{P}_{\mathcal{S}_1}}{\mathrm{d}\mathbb{P}_0} \frac{\mathrm{d}\mathbb{P}_{\mathcal{S}_2}}{\mathrm{d}\mathbb{P}_0}(\mathbf{X})\right] = \alpha^2 \exp\left(-s^* \beta^{*2}\right) \int (2\pi)^{-d/2} \exp\left(\sum_{j \in \mathcal{S}_1} \beta^* x_j + \sum_{j \in \mathcal{S}_2} \beta^* x_j - \sum_{j \in [d]} x_j^2/2\right) \mathrm{d}\mathbf{x}$$
$$+ \alpha(1-\alpha) \exp\left(-s^* \beta^{*2}/2\right) \int (2\pi)^{-d/2} \exp\left(\sum_{j \in \mathcal{S}_1} \beta^* x_j - \sum_{j \in [d]} x_j^2/2\right) \mathrm{d}\mathbf{x}$$
$$+ \alpha(1-\alpha) \exp\left(-s^* \beta^{*2}/2\right) \int (2\pi)^{-d/2} \exp\left(\sum_{j \in \mathcal{S}_2} \beta^* x_j - \sum_{j \in [d]} x_j^2/2\right) \mathrm{d}\mathbf{x}$$
$$+ (1 - \alpha)^2. \quad (7.16)$$

Note that we have

$$\sum_{j\in\mathcal{S}_1}\beta^*x_j + \sum_{j\in\mathcal{S}_2}\beta^*x_j = \sum_{j\in\mathcal{S}_1-\mathcal{S}_2}\beta^*x_j + \sum_{j\in\mathcal{S}_2-\mathcal{S}_1}\beta^*x_j + 2\sum_{j\in\mathcal{S}_1\cap\mathcal{S}_2}\beta^*x_j.$$

Hence, for the first term on the right-hand side of (7.16) we have

$$\int \exp\bigl(\sum_{j\in\mathcal{S}_1-\mathcal{S}_2}\beta^*x_j + \sum_{j\in\mathcal{S}_2-\mathcal{S}_1}\beta^*x_j + 2\sum_{j\in\mathcal{S}_1\cap\mathcal{S}_2}\beta^*x_j - \sum_{j\in[d]}x_j^2/2\bigr)\mathrm{d}\mathbf{x}$$
$$= \underbrace{\int \exp\bigl(-\sum_{j\in[d]-\mathcal{S}_1\cup\mathcal{S}_2}x_j^2/2\bigr)\mathrm{d}\mathbf{x}}_{\text{(i)}} \cdot \underbrace{\int \exp\bigl[-\sum_{j\in\mathcal{S}_1-\mathcal{S}_2}(\beta^*x_j - x_j^2/2)\bigr]\mathrm{d}\mathbf{x}}_{\text{(ii)}}$$
$$\cdot \underbrace{\int \exp\bigl[-\sum_{j\in\mathcal{S}_2-\mathcal{S}_1}(\beta^*x_j - x_j^2/2)\bigr]\mathrm{d}\mathbf{x}}_{\text{(iii)}} \cdot \underbrace{\int \exp\bigl[-\sum_{j\in\mathcal{S}_1\cap\mathcal{S}_2}(2\beta^*x_j - x_j^2/2)\bigr]\mathrm{d}\mathbf{x}}_{\text{(iv)}}.$$

Note that term (i) is $(2\pi)^{(d-|\mathcal{S}_1\cup\mathcal{S}_2|)/2}$ since it is the integration of the Gaussian density function. For term (ii) note that $\beta^*x_j - x_j^2/2 = -(x_j-\beta^*)^2/2 + \beta^{*2}/2$. Thus term (ii) equals $(2\pi)^{|\mathcal{S}_1-\mathcal{S}_2|/2}\exp(|\mathcal{S}_1-\mathcal{S}_2|\beta^{*2}/2)$. Using a similar calculation for terms (iii) and (iv) as well as the second and third terms in (7.16), we obtain

$$\mathbb{E}_{\mathbb{P}_0}\left[\frac{\mathrm{d}\mathbb{P}_{\mathcal{S}_1}}{\mathrm{d}\mathbb{P}_0}\frac{\mathrm{d}\mathbb{P}_{\mathcal{S}_2}}{\mathrm{d}\mathbb{P}_0}(\mathbf{X})\right]$$
$$= \alpha^2\exp\bigl(-s^*\beta^{*2} + |\mathcal{S}_1-\mathcal{S}_2|\beta^{*2}/2 + |\mathcal{S}_2-\mathcal{S}_1|\beta^{*2}/2 + 2|\mathcal{S}_1\cap\mathcal{S}_2|\beta^{*2}\bigr) + 2\alpha(1-\alpha) + (1-\alpha)^2$$
$$= \alpha^2\exp\bigl(|\mathcal{S}_1\cap\mathcal{S}_2|\beta^{*2}\bigr) + (1-\alpha^2),$$

which concludes the proof of Lemma 5.1. $\square$

### 7.5 Proof of Lemma 5.2

*Proof.* By Lemma 5.1, we have

$$h(|\mathcal{S}\cap\mathcal{S}'|) = \alpha^2\exp\bigl(|\mathcal{S}_1\cap\mathcal{S}_2|\beta^{*2}\bigr) + (1-\alpha^2).$$

For notational simplicity, we abbreviate $k(q,\mathcal{S})$ as $k$ and $\bar{\mathcal{C}}(q,\mathcal{S})$ as $\bar{\mathcal{C}}$. Following the notation in the proof of Theorem 4.4, for any $\mathcal{S} \in \mathcal{C}(q)$ we have

$$\mathbb{E}_{\mathcal{S}'\sim\mathbb{Q}_{\bar{\mathcal{C}}}}[h(|\mathcal{S}\cap\mathcal{S}'|)] = \frac{\alpha^2\cdot\bigl\{\sum_{j=0}^{k-1}|\mathcal{C}_j(\mathcal{S})|\exp\bigl[(s^*-j)\beta^{*2}\bigr] + |\mathcal{C}_k'(\mathcal{S})|\exp\bigl[(s^*-k)\cdot\beta^{*2}\bigr]\bigr\}}{\sum_{j=0}^{k-1}|\mathcal{C}_j(\mathcal{S})| + |\mathcal{C}_k'(\mathcal{S})|} + (1-\alpha^2)$$
$$\leq \frac{\alpha^2\cdot\sum_{j=0}^{k-1}|\mathcal{C}_j(\mathcal{S})|\exp\bigl[(s^*-j)\beta^{*2}\bigr]}{\sum_{j=0}^{k-1}|\mathcal{C}_j(\mathcal{S})|} + (1-\alpha^2). \tag{7.17}$$

Note that $|\mathcal{C}_j(\mathcal{S})| = \binom{s^*}{s^*-j}\binom{d-s^*}{j}$. Hence, for any $j \in \{0,\ldots,k-1\}$ ($k \leq s^*$) we have

$$|\mathcal{C}_{j+1}(\mathcal{S})|/|\mathcal{C}_j(\mathcal{S})| = \frac{(s^*-j)(d-s^*-j)}{(j+1)^2} \geq \frac{(s^*-k+1)(d-2s^*+1)}{s^{*2}}. \tag{7.18}$$

In the rest of proof, we consider two settings: (i) $s^{*2}/d = o(1)$; (ii) $\lim_{d\to\infty} s^{*2}/d > 0$.

For setting (i), i.e., $s^{*2}/d = o(1)$, first note that $k \leq s^*$. Then on the right-hand side of (7.18) it holds that

$$(s^* - k + 1)(d - 2s^* + 1)/s^{*2} \geq (d - 2s^* + 1)/s^{*2} \geq d/(2s^{*2}), \tag{7.19}$$

where the last inequality is from the fact that $d$ and $s^*$ are sufficiently large. Let the right-hand side of (7.19) be $\zeta$. By applying Lemma 7.1 to the right-hand side of (7.17), we have

$$\mathbb{E}_{\mathcal{S}' \sim \mathbb{Q}_{\bar{\mathcal{C}}}}[h(|\mathcal{S} \cap \mathcal{S}'|)] \leq \frac{\alpha^2 \cdot \sum_{j=0}^{k-1} \zeta^j \exp\big[(s^* - j)\beta^{*2}\big]}{\sum_{j=0}^{k-1} \zeta^j} + (1 - \alpha^2)$$

$$\leq \frac{\alpha^2 \cdot \exp\big[(s^* - k + 1)\beta^{*2}\big] \cdot (1 - \zeta^{-1})}{1 - \zeta^{-1} \exp(\beta^{*2})} + (1 - \alpha^2). \tag{7.20}$$

Here we use $\zeta^{-1} \exp(\beta^{*2}) = o(1)$. For notational simplicity, we denote $\sqrt{\log(1/\xi)/n}$ by $\tau$ hereafter. Since (7.20) holds for any $\mathcal{S} \in \mathcal{C}(q)$ and by Theorem 4.4 we have

$$\sup_{\mathcal{S} \in \mathcal{C}(q)} \big\{\mathbb{E}_{\mathcal{S}' \sim \mathbb{Q}_{\bar{\mathcal{C}}}}[h(|\mathcal{S} \cap \mathcal{S}'|)]\big\} \geq 1 + \tau^2,$$

by calculation we have

$$s^* - k + 1 \geq \log(1 + \tau^2/\alpha^2)/\beta^{*2} - \log\left[\frac{1 - \zeta^{-1}}{1 - \zeta^{-1} \exp(\beta^{*2})}\right]/\beta^{*2}. \tag{7.21}$$

Note that by Taylor expansion we have

$$\log\left[\frac{1 - \zeta^{-1}}{1 - \zeta^{-1} \exp(\beta^{*2})}\right] = \log\left\{1 + \frac{[\exp(\beta^{*2}) - 1]\zeta^{-1}}{1 - \zeta^{-1} \exp(\beta^{*2})}\right\}$$

$$= O\left\{\frac{[\exp(\beta^{*2}) - 1]\zeta^{-1}}{1 - \zeta^{-1} \exp(\beta^{*2})}\right\} = O(\zeta^{-1}\beta^{*2}), \tag{7.22}$$

where we use the fact that $\zeta^{-1} \exp(\beta^{*2}) = o(1)$. Therefore, from (7.21) we have that, for a sufficiently large $d$,

$$s^* - k + 2 \geq \log(1 + \tau^2/\alpha^2)/\beta^{*2}. \tag{7.23}$$

Now we derive the upper bound of $|\mathcal{C}(q)| = |\bar{\mathcal{C}}|$. Following the definition of $|\bar{\mathcal{C}}|$ and $\mathcal{C}_j(\mathcal{S})$, by (7.18), (7.19), and (7.23), we have

$$|\mathcal{C}(q)| = |\bar{\mathcal{C}}| \leq \sum_{j=0}^{k} |\mathcal{C}_j(\mathcal{S})| \leq \zeta^{-s^*} |\mathcal{C}_{s^*}(\mathcal{S})| \sum_{j=0}^{k} \zeta^j \leq \frac{\zeta^{-(s^* - k)}|\mathcal{C}|}{1 - \zeta^{-1}}$$

$$\leq 2\exp\big\{-\log\zeta \cdot \big[\log(1 + \tau^2/\alpha^2)/\beta^{*2} - 2\big]\big\}|\mathcal{C}|,$$

where in the last inequality we invoke the fact that $\zeta^{-1} = o(1)$.

For setting (ii), i.e., $\lim_{d\to\infty} s^{*2}/d > 0$, we define

$$\bar{k} = s^* + 1 - \log(1+\tau^2/\alpha^2)/(2\beta^{*2}). \tag{7.24}$$

For notational simplicity, let

$$\gamma = d/(2s^{*2}) \cdot \log(1+\tau^2/\alpha^2)/(2\beta^{*2}). \tag{7.25}$$

Then by our assumption, $\gamma$ is lower bounded by a sufficiently large constant larger than one. By the definition of $\bar{k}$ in (7.24), for $j \leq \bar{k} - 1$ we have

$$|\mathcal{C}_{j+1}(\mathcal{S})|/|\mathcal{C}_j(\mathcal{S})| = \frac{(s^*-j)(d-s^*-j)}{(j+1)^2} \geq \frac{(s^*-\bar{k}+1)(d-2s^*+1)}{s^{*2}} \geq \gamma. \tag{7.26}$$

Here the last inequality follows from (7.25) and the fact that $(d-2s^*+1)/s^{*2} \geq d/(2s^{*2})$, since $d$ and $s^*$ are sufficiently large. Meanwhile, in the following proof we denote

$$f(k) = \frac{\alpha^2 \cdot \sum_{j=0}^{k-1} |\mathcal{C}_j(\mathcal{S})| \exp[(s^*-j)\beta^{*2}]}{\sum_{j=0}^{k-1} |\mathcal{C}_j(\mathcal{S})|}. \tag{7.27}$$

Following the same derivation of (7.20) with $\zeta$ replaced by $\gamma$ and $k$ replaced by $\bar{k}$, we have

$$f(\bar{k}) + (1-\alpha^2) \leq \frac{\alpha^2 \cdot \exp[(s^*-\bar{k}+1)\beta^{*2}] \cdot (1-\gamma^{-1})}{1-\gamma^{-1}\exp(\beta^{*2})} + (1-\alpha^2) \tag{7.28}$$

$$= \alpha^2 \exp\left\{(s^*-\bar{k}+1)\beta^{*2} + \log\left[1+\gamma^{-1} \cdot \frac{\exp(\beta^{*2})-1}{1-\gamma^{-1}\exp(\beta^{*2})}\right]\right\} + (1-\alpha^2).$$

Note that on the right-hand side of (7.28), we have

$$\log\left[1+\gamma^{-1} \cdot \frac{\exp(\beta^{*2})-1}{1-\gamma^{-1}\exp(\beta^{*2})}\right] \leq C\gamma^{-1}[\exp(\beta^{*2})-1] \leq C'\gamma^{-1}\beta^{*2}, \tag{7.29}$$

where $C$ and $C'$ are absolute constants. Here we invoke the facts that $\log(1+x) \leq x$, $\beta^* = o(1)$, and $\exp(x) \leq 1 + 2x$ for $x \in [0, 0.5)$. Then by applying (7.24) and (7.29), from (7.28) we have

$$f(\bar{k}) + (1-\alpha^2) \leq \alpha^2 \exp[(s^*-\bar{k}+1+C'\gamma^{-1})\beta^{*2}] + (1-\alpha^2)$$
$$\leq \alpha^2 \exp[\log(1+\tau^2/\alpha^2)] + (1-\alpha^2) = 1 + \tau^2,$$

where we use the fact that $\gamma$ is lower bounded by a sufficiently large constant. Meanwhile, note that $f(k)$ in (7.27) is nonincreasing for $k \leq \bar{k}$. Hence, if $f(k) + (1-\alpha^2) \geq 1 + \tau^2$, we have $k \leq \bar{k}$. By the same derivation of (7.20) with $\zeta$ replaced by $\gamma$, we have

$$\frac{\alpha^2 \cdot \exp[(s^*-k+1)\beta^{*2}] \cdot (1-\gamma^{-1})}{1-\gamma^{-1}\exp(\beta^{*2})} + (1-\alpha^2) \geq \mathbb{E}_{\mathcal{S}'\sim\mathbb{Q}_{\bar{c}}}[h(|\mathcal{S}\cap\mathcal{S}'|)] \geq 1+\tau^2.$$

Following the same derivation of (7.21) and (7.22) with $\zeta$ replaced by $\gamma$, we obtain

$$s^* - k + 2 \geq \log(1+\tau^2/\alpha^2)/\beta^{*2}. \tag{7.30}$$

Combining (7.30) and (7.24) we have

$$\bar{k} - k \geq \log(1 + \tau^2/\alpha^2)/(2\beta^{*2}) - 1 = 2\gamma s^{*2}/d - 1. \tag{7.31}$$

Now we establish the upper bound of $|\mathcal{C}(q)| = |\bar{\mathcal{C}}|$. From (7.26) we have

$$|\bar{\mathcal{C}}| \leq \sum_{j=0}^{k} |\mathcal{C}_j(\mathcal{S})| \leq \gamma^{-\bar{k}} |\mathcal{C}_{\bar{k}}(\mathcal{S})| \sum_{j=0}^{k} \gamma^j \leq \frac{\gamma^{-(\bar{k}-k)}|\mathcal{C}|}{1 - \gamma^{-1}} \leq 2\gamma^{-(\bar{k}-k)}|\mathcal{C}|. \tag{7.32}$$

Here the last inequality follows from the fact that $\gamma$ is lower bounded by a sufficiently large constant. Combining (7.32) and (7.31) we have

$$|\mathcal{C}(q)| = |\bar{\mathcal{C}}| \leq 2\exp\bigl[-\log\gamma \cdot \bigl(2\gamma s^{*2}/d - 1\bigr)\bigr]|\mathcal{C}|.$$

Thus we conclude the proof of Lemma 5.2. □

## 7.6 Proof of Theorem 5.3

*Proof.* We employ the notation in the proof of Lemma 5.2. Let $\delta > 0$ be a sufficiently small constant. We assume $T = O(d^\eta)$, where $\eta$ is a constant, and $T \geq 1$. Recall that in (5.1) we define

$$\zeta = d/(2s^{*2}), \quad \tau = \sqrt{\log(1/\xi)/n}, \quad \text{and} \quad \gamma = d/(2s^{*2}) \cdot \log(1 + \tau^2/\alpha^2)/(2\beta^{*2}).$$

Here $\xi \in [0, 1/4)$. Under setting (i) in Lemma 5.2, i.e., $s^{*2}/d = o(1)$, we consider the following cases.

(a) $\lim_{d \to \infty} \zeta/d^\delta > 0$ and $\lim_{d \to \infty} \tau^2/\alpha^2 > 0$. As long as $\beta^* = o(1)$, we have

$$\frac{T \cdot \sup_{q \in \mathcal{Q}} |\mathcal{C}(q)|}{|\mathcal{C}|} \leq T \cdot 2\exp\bigl\{-\log\zeta \cdot \bigl[\log(1 + \tau^2/\alpha^2)/\beta^{*2} - 2\bigr]\bigr\}$$

$$= O\bigl(d^\eta \cdot 2\exp\bigl\{-\log\zeta \cdot \bigl[\log(1 + \tau^2/\alpha^2)/\beta^{*2} - 2\bigr]\bigr\}\bigr) = o(1), \tag{7.33}$$

where the last equality holds because $\log(1 + \tau^2/\alpha^2)/\beta^{*2} \to \infty$.

(b) $\lim_{d \to \infty} \zeta/d^\delta > 0$ and $\tau^2/\alpha^2 = o(1)$. In (7.33) we have

$$\log(1 + \tau^2/\alpha^2)/\beta^{*2} \geq \tau^2/(2\alpha^2\beta^{*2}) = \log(T/\zeta)/(2\beta^{*2}n\alpha^2) \tag{7.34}$$

for $d$ sufficiently large, where we use $\log(1 + x) \geq x/2$ for $x$ sufficiently small. Then as long as $\beta^{*2}n\alpha^2 = o(1)$, similar to (7.33) we have

$$\frac{T \cdot \sup_{q \in \mathcal{Q}} |\mathcal{C}(q)|}{|\mathcal{C}|} = O\bigl(d^\eta \cdot 2\exp\bigl\{-\log\zeta \cdot \bigl[\log(T/\zeta)/(2\beta^{*2}n\alpha^2) - 2\bigr]\bigr\}\bigr) = o(1).$$

(c) $\zeta/d^\delta = o(1)$ and $\lim_{d \to \infty} \tau^2/\alpha^2 > 0$. As long as $\beta^{*2}\log d = o(1)$, we have

$$\bigl[\log(1 + \tau^2/\alpha^2)/\beta^{*2}\bigr]/\log d \to \infty.$$

Hence, we have that (7.33) holds.

(d) $\zeta/d^\delta = o(1)$ and $\tau^2/\alpha^2 = o(1)$. As long as $\beta^{*2}n\alpha^2 = o(1)$, by (7.34) we have

$$\frac{T \cdot \sup_{q \in \mathcal{Q}} |\mathcal{C}(q)|}{|\mathcal{C}|} \leq T \cdot 2\exp\bigl\{-\log\zeta \cdot \bigl[\log(1/\xi)/(2\beta^{*2}n\alpha^2) - 2\bigr]\bigr\}$$
$$= O\bigl[T\zeta^2 \cdot (1/\xi)^{-\log\zeta/(2\beta^{*2}n\alpha^2)}\bigr] = O\bigl[d^{2\delta}\xi \cdot (1/\xi)^{1-\log\zeta/(\beta^{*2}n\alpha^2)}\bigr] = o(1).$$

Now we consider two cases under setting (ii) in Lemma 5.2, i.e., $\lim_{d\to\infty} s^{*2}/d > 0$:

(a) $\tau^2/\alpha^2 = o(1)$. First note that by (5.1) we have

$$\gamma = d/(2s^{*2}) \cdot \log(1 + \tau^2/\alpha^2)/(2\beta^{*2}) \geq d/(2s^{*2}) \cdot \log(1/\xi)/(4\beta^{*2}n\alpha^2), \quad (7.35)$$

where we use $\log(1+x) \geq x/2$ for $x$ sufficiently small. Then we have

$$\frac{T \cdot \sup_{q \in \mathcal{Q}} |\mathcal{C}(q)|}{|\mathcal{C}|} \leq T \cdot 2\exp\bigl[-\log\gamma \cdot (2\gamma s^{*2}/d - 1)\bigr] \quad (7.36)$$
$$= O\bigl(T \cdot \exp\bigl\{-\log\gamma \cdot \bigl[\log(1/\xi)/(4\beta^{*2}n\alpha^2) - 1\bigr]\bigr\}\bigr) = O\bigl[\xi\gamma \cdot (1/\xi)^{1-\log\gamma/(4\beta^{*2}n\alpha^2)}\bigr].$$

As long as $\beta^{*2}s^{*2}n\alpha^2/d = o(1)$, by (7.35) we have $\gamma \to \infty$. Moreover, as long as $\beta^{*2}n\alpha^2 = o(1)$ the right-hand side of (7.36) is $o(1)$.

(b) $\lim_{d\to\infty} \tau^2/\alpha^2 > 0$. As long as $\beta^{*2}s^{*2}\log d/d = o(1)$, by (5.1) we have

$$\gamma/\log d = d/(2s^{*2}) \cdot \log(1 + \tau^2/\alpha^2)/(2\beta^{*2}\log d) \to \infty. \quad (7.37)$$

Since $T = O(d^\eta)$, we have

$$\frac{T \cdot \sup_{q \in \mathcal{Q}} |\mathcal{C}(q)|}{|\mathcal{C}|} \leq T \cdot 2\exp\bigl[-\log\gamma \cdot (2\gamma s^{*2}/d - 1)\bigr]$$
$$= O\bigl\{\exp\bigl[\eta\log d - \log\gamma \cdot (2\gamma s^{*2}/d - 1)\bigr]\bigr\} = o(1).$$

Here the last equality is from (7.37) and our assumption for setting (ii) that $\lim_{d\to\infty} s^{*2}/d > 0$.

Thus we conclude the proof of Theorem 5.3. □

## 7.7 Proof of Proposition 5.4

*Proof.* We define $\mathbb{P}^n_{\mathcal{S} \in \mathcal{C}} = 1/|\mathcal{C}| \cdot \sum_{\mathcal{S} \in \mathcal{C}} \mathbb{P}^n_{\mathcal{S}}$, where the superscript $n$ denotes the product distribution. Let $Z = |\mathcal{S} \cap \mathcal{S}'|$, where $\mathcal{S}$ and $\mathcal{S}'$ are uniformly drawn from $\mathcal{C}$. In the sequel we consider two settings.

For $\beta^{*2}\alpha^2 n = o(1)$, $\beta^{*2}s^* = o(1)$, and $\beta^{*2}\alpha^2 n s^{*2}/d = o(1)$, note that

$$\chi^2(\mathbb{P}^n_{\mathcal{S} \in \mathcal{C}}, \mathbb{P}^n_0) = \mathbb{E}_{\mathbb{P}^n_0}\left\{\left[\frac{d\mathbb{P}^n_{\mathcal{S} \in \mathcal{C}}}{d\mathbb{P}^n_0}(\boldsymbol{X}_1, \ldots, \boldsymbol{X}_n) - 1\right]^2\right\} \quad (7.38)$$
$$= \frac{1}{|\mathcal{C}|^2}\sum_{\mathcal{S}_1, \mathcal{S}_2 \in \mathcal{C}} \mathbb{E}_{\mathbb{P}^n_0}\left[\frac{d\mathbb{P}^n_{\mathcal{S}_1}}{d\mathbb{P}^n_0}\frac{d\mathbb{P}^n_{\mathcal{S}_2}}{d\mathbb{P}^n_0}(\boldsymbol{X}_1, \ldots, \boldsymbol{X}_n)\right] - 1 = \frac{1}{|\mathcal{C}|^2}\sum_{\mathcal{S}_1, \mathcal{S}_2 \in \mathcal{C}} \mathbb{E}_{\mathbb{P}_0}\left[\frac{d\mathbb{P}_{\mathcal{S}_1}}{d\mathbb{P}_0}\frac{d\mathbb{P}_{\mathcal{S}_2}}{d\mathbb{P}_0}(\boldsymbol{X})\right]^n - 1$$
$$= \frac{1}{|\mathcal{C}|^2}\sum_{\mathcal{S}_1, \mathcal{S}_2 \in \mathcal{C}} h(|\mathcal{S}_1 \cap \mathcal{S}_2|)^n - 1 = \mathbb{E}_Z[h(Z)^n] - 1 = \mathbb{E}_Z\bigl\{\bigl[\alpha^2\exp(\beta^{*2}Z) + (1-\alpha^2)\bigr]^n\bigr\} - 1,$$

where the third last equality is from Condition 4.3 and the last equality is from Lemma 5.1. On the right-hand of (7.38) we have

$$\big[\alpha^2 \exp(\beta^{*2} Z) + (1-\alpha^2)\big]^n \leq \big(2\alpha^2 \beta^{*2} Z + 1\big)^n$$
$$= \exp\big[n \log\big(2\alpha^2 \beta^{*2} Z + 1\big)\big] \leq \exp\big(2n\alpha^2 \beta^{*2} Z\big),$$

where we use the fact that $Z \leq s^*$, $\beta^{*2} s^* = o(1)$, and $\alpha = O(1)$. Furthermore, we have

$$\mathbb{E}_Z\big[\exp(2n\alpha^2 \beta^{*2} Z)\big] = \mathbb{E}_{\mathcal{S}\sim\mathbb{Q}_\mathcal{C}}\big\{\textstyle\prod_{j=1}^{s^*} \exp\big[2n\alpha^2 \beta^{*2} \mathbb{1}(j \in \mathcal{S})\big]\big\} \leq \prod_{j=1}^{s^*} \mathbb{E}_{\mathcal{S}\sim\mathbb{Q}_\mathcal{C}}\big\{\exp\big[2n\alpha^2 \beta^{*2} \mathbb{1}(j \in \mathcal{S})\big]\big\}$$
$$= \big[s^*/d \cdot \exp(2n\alpha^2 \beta^{*2}) + (1 - s^*/d)\big]^{s^*}, \tag{7.39}$$

where the inequality is from the negative association property. See, e.g., Addario-Berry et al. (2010) for details. Then from (7.39) we have

$$\mathbb{E}_Z\big[\exp(2n\alpha^2 \beta^{*2} Z)\big] \leq \big(s^*/d \cdot 4n\alpha^2 \beta^{*2} + 1\big)^{s^*} = \exp\big[s^* \log\big(s^*/d \cdot 4n\alpha^2 \beta^{*2} + 1\big)\big]$$
$$\leq \exp\big(s^{*2}/d \cdot 4n\alpha^2 \beta^{*2}\big) \leq s^{*2}/d \cdot 8n\alpha^2 \beta^{*2} + 1, \tag{7.40}$$

where we use the fact that $\beta^{*2} \alpha^2 n = o(1)$ and $\beta^{*2} \alpha^2 n s^{*2}/d = o(1)$. Plugging it into (7.38) we obtain $\chi^2(\mathbb{P}^n_{\mathcal{S}\in\mathcal{C}}, \mathbb{P}^n_0) = o(1)$, which further implies $\mathrm{TV}(\mathbb{P}^n_{\mathcal{S}\in\mathcal{C}}, \mathbb{P}^n_0) \leq \sqrt{\chi^2(\mathbb{P}^n_{\mathcal{S}\in\mathcal{C}}, \mathbb{P}^n_0)} = o(1)$. Here $\mathrm{TV}(\cdot,\cdot)$ is the total variation distance. See, e.g., Tsybakov (2008) for the relationship between $\chi^2$-divergence and total variation distance.

For $\beta^{*2} s^{*2}/d = o(1)$, $\beta^{*2} \alpha n = o(1)$, and $\beta^{*2} \alpha^2 n s^{*2}/d = o(1)$, for notational simplicity we define several events. Let $Y$ be the latent variable associated with the mixture in the alternative hypothesis, i.e., $\boldsymbol{X}$ follows $N(\boldsymbol{\theta}, \mathbf{I}_d)$ for $Y = 1$ and $N(\mathbf{0}, \mathbf{I}_d)$ for $Y = 0$. Let $Y_1, \ldots, Y_n$ be the independent copies of $Y$. We define

$$\mathcal{E}_a = \{\textstyle\sum_{i=1}^n Y_i = a\}, \quad \text{and} \quad \mathcal{E}'_a = \{\textstyle\sum_{i=1}^n Y_i > a\}.$$

Let $\mathbb{P}^{\mathcal{E}_a}_\mathcal{S}$ be the joint distribution of $\boldsymbol{X}_1, \ldots, \boldsymbol{X}_n$, which are the independent copies of $\boldsymbol{X}$, conditioning on event $\mathcal{E}_a$. We define $\mathbb{P}^{\mathcal{E}_a}_{\mathcal{S}\in\mathcal{C}} = 1/|\mathcal{C}| \cdot \sum_{\mathcal{S}\in\mathcal{C}} \mathbb{P}^{\mathcal{E}_a}_\mathcal{S}$. Note that

$$\mathbb{P}^n_{\mathcal{S}\in\mathcal{C}} = \mathbb{E}_{K\sim B(n,\alpha)}\big(\mathbb{P}^{\mathcal{E}_K}_{\mathcal{S}\in\mathcal{C}}\big),$$

where $B(n,\alpha)$ denotes the binomial distribution. Hence, for total variation distance we have

$$\mathrm{TV}(\mathbb{P}^n_{\mathcal{S}\in\mathcal{C}}, \mathbb{P}^n_0) = \mathrm{TV}\big[\mathbb{E}_{K\sim B(n,\alpha)}\big(\mathbb{P}^{\mathcal{E}_K}_{\mathcal{S}\in\mathcal{C}}\big), \mathbb{P}^n_0\big] \leq \mathbb{E}_{K\sim B(n,\alpha)}\big[\mathrm{TV}\big(\mathbb{P}^{\mathcal{E}_K}_{\mathcal{S}\in\mathcal{C}}, \mathbb{P}^n_0\big)\big] \tag{7.41}$$
$$\leq 2\mathbb{P}(\mathcal{E}'_{Cn\alpha}) + \sum_{k \leq Cn\alpha} \mathbb{P}(\mathcal{E}_k) \, \mathrm{TV}\big(\mathbb{P}^{\mathcal{E}_k}_{\mathcal{S}\in\mathcal{C}}, \mathbb{P}^n_0\big),$$

where the first inequality follows from the convexity of total variation distance and Jensen's inequality. The second inequality follows from the law of total expectation as well as the fact that total variation distance is upper bounded by two. In (7.41), $C$ is a sufficiently large absolute constant. Note that

$$\mathrm{TV}\big(\mathbb{P}^{\mathcal{E}_k}_{\mathcal{S}\in\mathcal{C}}, \mathbb{P}^n_0\big) \leq \sqrt{\chi^2\big(\mathbb{P}^{\mathcal{E}_k}_{\mathcal{S}\in\mathcal{C}}, \mathbb{P}^n_0\big)}. \tag{7.42}$$

Let $\mathcal{B}_k = \{\mathcal{S} \subseteq [n] : |\mathcal{S}| = k\}$. For a random matrix $\mathbf{X} \in \mathbb{R}^{n \times d}$ with independent entries, we denote by $\mathbf{X} \sim \mathbb{P}_{\mathcal{S}_1, \mathcal{S}_2}$ if $X_{i,j} \sim N(\beta^*, 1)$ for $i \in \mathcal{S}_1, j \in \mathcal{S}_2$ and $X_{i,j} \sim N(0,1)$ otherwise. Besides, with slight abuse of notation, we denote by $\mathbf{X} \sim \mathbb{P}_0$ if $X_{i,j} \sim N(0,1)$ for all $i \in [n], j \in [d]$. Then we have

$$\chi^2\big(\mathbb{P}^{\mathcal{E}_k}_{\mathcal{S} \in \mathcal{C}}, \mathbb{P}^n_0\big) = \mathbb{E}_{\mathbb{P}_0}\bigg\{\bigg[\frac{1}{|\mathcal{C}||\mathcal{B}_k|} \sum_{\mathcal{S}_1 \in \mathcal{C}} \sum_{\mathcal{S}_2 \in \mathcal{B}_k} \frac{\mathrm{d}\mathbb{P}_{\mathcal{S}_1, \mathcal{S}_2}}{\mathrm{d}\mathbb{P}_0}(\mathbf{X}) - 1\bigg]^2\bigg\} \tag{7.43}$$

$$= \frac{1}{|\mathcal{C}|^2|\mathcal{B}_k|^2} \sum_{\mathcal{S}_1, \mathcal{S}'_1 \in \mathcal{C}} \sum_{\mathcal{S}_2, \mathcal{S}'_2 \in \mathcal{B}_k} \mathbb{E}_{\mathbb{P}_0}\bigg[\frac{\mathrm{d}\mathbb{P}_{\mathcal{S}_1, \mathcal{S}_2}}{\mathrm{d}\mathbb{P}_0} \frac{\mathrm{d}\mathbb{P}_{\mathcal{S}'_1, \mathcal{S}'_2}}{\mathrm{d}\mathbb{P}_0}(\mathbf{X})\bigg] - 1$$

$$= \mathbb{E}_{Z_1, Z_2}\big[\exp\big(\beta^{*2} Z_1 Z_2\big)\big] - 1.$$

Here $Z_1 = |\mathcal{S}_1 \cap \mathcal{S}'_1|$, where $\mathcal{S}_1$ and $\mathcal{S}'_1$ are uniformly drawn from $\mathcal{C}$, while $Z_2 = |\mathcal{S}_2 \cap \mathcal{S}'_2|$, where $\mathcal{S}_2$ and $\mathcal{S}'_2$ are uniformly drawn from $\mathcal{B}_k$. Also, the last equality is obtained by plugging in the Gaussian probability density function. By a similar derivation of (7.39) and (7.40), we have

$$\mathbb{E}_{Z_1, Z_2}\big[\exp\big(\beta^{*2} Z_1 Z_2\big)\big] \leq \mathbb{E}_{Z_2}\big\{\big[s^*/d \cdot \exp(\beta^{*2} Z_2) + (1 - s^*/d)\big]^{s^*}\big\}$$

$$\leq \mathbb{E}_{Z_2}\big[\exp\big(2s^{*2}/d \cdot \beta^{*2} Z_2\big)\big] \leq \big(4k/n \cdot s^{*2}/d \cdot \beta^{*2} + 1\big)^k$$

$$\leq 8k^2/n \cdot s^{*2}/d \cdot \beta^{*2} + 1.$$

Plugging it into the right-hand side of (7.43) and then into (7.41) and (7.42), we obtain

$$\mathrm{TV}(\mathbb{P}^n_{\mathcal{S} \in \mathcal{C}}, \mathbb{P}^n_0) \leq 2\mathbb{P}(\mathcal{E}'_{Cn\alpha}) + \sum_{k \leq Cn\alpha} \mathbb{P}(\mathcal{E}_k) \mathrm{TV}\big(\mathbb{P}^{\mathcal{E}_k}_{\mathcal{S} \in \mathcal{C}}, \mathbb{P}^n_0\big)$$

$$\leq 2\exp[-\alpha n \cdot C^2/(2+C)] + \sum_{k \leq Cn\alpha} \mathbb{P}(\mathcal{E}_k)\sqrt{8C^2 \cdot k^2/n \cdot s^{*2}/d \cdot \beta^{*2}}$$

$$\leq 2\exp[-\alpha n \cdot C^2/(2+C)] + \sqrt{8C^2 \cdot (\alpha n)^2/n \cdot s^{*2}/d \cdot \beta^{*2}}. \tag{7.44}$$

Here the second inequality is based on the multiplicative Hoeffding's inequality (Angluin and Valiant, 1977). Recall that we assume that $\alpha n \to \infty$. Also, according to our assumption, the second term on the right-hand side of (7.44) is $o(1)$. Hence, we have $\mathrm{TV}(\mathbb{P}^n_{\mathcal{S} \in \mathcal{C}}, \mathbb{P}^n_0) = o(1)$.

In summary, in both settings we obtain $\mathrm{TV}(\mathbb{P}^n_{\mathcal{S} \in \mathcal{C}}, \mathbb{P}^n_0) = o(1)$, which by Theorem 2.2 in Tsybakov (2008) implies

$$\inf_\phi R(\phi) \geq 1 - \mathrm{TV}(\mathbb{P}^n_{\mathcal{S} \in \mathcal{C}}, \mathbb{P}^n_0) = 1 - o(1)$$

for $s^*, d, n$ sufficiently large. Thus we conclude the proof of Proposition 5.4. $\square$

## 7.8 Proof of Theorem 5.5

*Proof.* For notational simplicity, we denote by $\boldsymbol{X} \sim \mathbb{P}'_\mathcal{S}$ if $\boldsymbol{X} \sim N(\boldsymbol{\theta}, \mathbf{I}_d)$ where $\theta_j = \beta$ for $j \in \mathcal{S}$ and $\theta_j = 0$ for $j \notin \mathcal{S}$. We consider the following settings.

**Setting (i):** For $\beta^{*2} s^{*2}/(d \log n) \to \infty$, recall that we consider the query function and test in (5.2) and (5.3). Note that we have

$$\mathbb{E}_{\mathbb{P}_0}[q(\boldsymbol{X})] = \mathbb{P}_0\big(1/\sqrt{d} \cdot \textstyle\sum_{j=1}^d X_j \geq \sqrt{2 \log n}\big) = 1 - \Phi\big(\sqrt{2 \log n}\big) \leq \exp(-\log n) = 1/n, \tag{7.45}$$

since $1/\sqrt{d} \cdot \sum_{j=1}^{d} X_j \sim N(0,1)$ under $\mathbb{P}_0$. Also, for any $\mathcal{S} \in \mathcal{C}$ we have

$$\mathbb{E}_{\mathbb{P}_\mathcal{S}}[q(\boldsymbol{X})] = \mathbb{P}_\mathcal{S}\big(1/\sqrt{d} \cdot \sum_{j=1}^{d} X_j \geq \sqrt{2\log n}\big) \tag{7.46}$$
$$= \alpha \mathbb{P}'_\mathcal{S}\big(1/\sqrt{d} \cdot \sum_{j=1}^{d} X_j \geq \sqrt{2\log n}\big) + (1-\alpha)\mathbb{E}_{\mathbb{P}_0}[q(\boldsymbol{X})] \leq \alpha + 1/n \leq 2\alpha,$$

where the second last inequality is from (7.45) and the inequality is from the fact that $\alpha n \geq C$ with $C \geq 1$ sufficiently large, which is specified in §2. Meanwhile, we have

$$\mathbb{E}_{\mathbb{P}_\mathcal{S}}[q(\boldsymbol{X})] - \mathbb{E}_{\mathbb{P}_0}[q(\boldsymbol{X})] \geq \alpha\big[\mathbb{P}'_\mathcal{S}\big(\sum_{j=1}^{d} X_j \geq \sqrt{2\log n}\big) - \mathbb{P}_0\big(\sum_{j=1}^{d} X_j \geq \sqrt{2\log n}\big)\big]$$
$$\geq \alpha/2 - \alpha/n \geq \alpha/4, \tag{7.47}$$

where the second inequality is obtained using the facts that $1/\sqrt{d} \cdot \sum_{j=1}^{d} X_j \sim N(\beta^* s^*/\sqrt{d}, 1)$ under $\mathbb{P}_0$, $\beta^{*2} s^{*2}/(d\log n) \to \infty$, and (7.45). Recall that we have $\eta(\mathcal{Q}_\mathscr{A}) = 0$, since the algorithm only uses one query. For $\mathbb{P}_0$, by (7.45) we have

$$\sqrt{2\operatorname{Var}[q(\boldsymbol{X})] \cdot [\eta(\mathcal{Q}_\mathscr{A}) + \log(1/\xi)]/n} \leq \sqrt{\mathbb{E}_{\mathbb{P}_0}[q(\boldsymbol{X})] \cdot 2\log(1/\xi)/n} \leq \sqrt{2\log(1/\xi)}/n \leq \alpha/8,$$

since we assume $\alpha n \geq C\log(1/\xi)$, where $C$ is sufficiently large, and it implies that $\alpha n \geq C\sqrt{\log(1/\xi)}$ by $\xi \in [0, 1/4)$. For $\mathbb{P}_\mathcal{S}$, by (7.46) we have

$$\sqrt{2\operatorname{Var}[q(\boldsymbol{X})] \cdot [\eta(\mathcal{Q}_\mathscr{A}) + \log(1/\xi)]/n} \leq \sqrt{\mathbb{E}_{\mathbb{P}_\mathcal{S}}[q(\boldsymbol{X})] \cdot 2\log(1/\xi)/n} \leq \sqrt{4\log(1/\xi) \cdot \alpha/n} \leq \alpha/8,$$

since $\alpha n \geq C\log(1/\xi)$ with $C$ sufficiently large. For the same reason, we have $2/(3n) \cdot \log(1/\xi) \leq \alpha/8$. Then we have

$$\overline{R}(\phi) = \overline{\mathbb{P}}_0\big[Z_q \geq 1 - \Phi\big(\sqrt{2\log n}\big) + \alpha/8\big] + \frac{1}{|\mathcal{C}|}\sum_{\mathcal{S} \in \mathcal{C}} \overline{\mathbb{P}}_\mathcal{S}\big[Z_q < 1 - \Phi\big(\sqrt{2\log n}\big) + \alpha/8\big] \tag{7.48}$$
$$\leq \overline{\mathbb{P}}_0\{|Z_q - \mathbb{E}_{\mathbb{P}_0}[q(\boldsymbol{X})]| > \alpha/8\} + \frac{1}{|\mathcal{C}|}\sum_{\mathcal{S} \in \mathcal{C}} \overline{\mathbb{P}}_\mathcal{S}\{|Z_q - \mathbb{E}_{\mathbb{P}_\mathcal{S}}[q(\boldsymbol{X})]| > \alpha/8\} \leq 2\xi,$$

where the first and second inequalities are from (7.45)-(7.47) and Definition 3.1.

**Setting (ii):** For $\beta^{*2} n\alpha^2/[\log d + \log(1/\xi)] \to \infty$, recall the query functions and test are defined in (5.4) and (5.5). We have

$$\mathbb{E}_{\mathbb{P}_0}[q_t(\boldsymbol{X})] = \mathbb{P}_0(X_t \geq \beta^*/2) = 1 - \Phi(\beta^*/2),$$

while for any $\mathcal{S} \in \mathcal{C}$ we have

$$\mathbb{E}_{\mathbb{P}_\mathcal{S}}[q_t(\boldsymbol{X})] = \mathbb{P}_\mathcal{S}(X_t \geq \beta^*/2) = \alpha \mathbb{P}'_\mathcal{S}(X_t \geq \beta^*/2) + (1-\alpha)\mathbb{E}_{\mathbb{P}_0}[q_t(\boldsymbol{X})].$$

Meanwhile, since $\beta^* = o(1)$, for $t \in \mathcal{S}$ we have

$$\mathbb{E}_{\mathbb{P}_\mathcal{S}}[q_t(\boldsymbol{X})] - \mathbb{E}_{\mathbb{P}_0}[q_t(\boldsymbol{X})] = \alpha[\mathbb{P}'_\mathcal{S}(X_t \geq \beta^*/2) - \mathbb{P}_0(X_t \geq \beta^*/2)] = \alpha \mathbb{P}_0(|X_t| \leq \beta^*/2) \tag{7.49}$$
$$= \alpha \int_{-\beta^*/2}^{\beta^*/2} \frac{\exp(-x^2/2)}{\sqrt{2\pi}} \mathrm{d}x \geq \frac{\alpha\beta^*}{\sqrt{2\pi}} \exp\big(-\beta^{*2}/8\big) \geq \alpha\beta^*/(2\pi),$$

where in the last inequality we use the fact that $\exp(-\beta^{*2}/8) \leq 1/\sqrt{2\pi}$ since $\beta^*$ is sufficiently small. Recall we have $\eta(\mathcal{Q}_{\mathscr{A}}) = \log d$. For all $t \in [T]$, since $\beta^{*2}n\alpha^2/[\log d + \log(1/\xi)] \to \infty$, for $\mathbb{P}_0$ we have

$$\sqrt{2\operatorname{Var}[q_t(\boldsymbol{X})] \cdot [\eta(\mathcal{Q}_{\mathscr{A}}) + \log(1/\xi)]/n} = \sqrt{\mathbb{E}_{\mathbb{P}_0}[q_t(\boldsymbol{X})]\{1 - \mathbb{E}_{\mathbb{P}_0}[q_t(\boldsymbol{X})]\} \cdot 2\log(d/\xi)/n}$$
$$\leq \sqrt{2[\log d + \log(1/\xi)]/n} \leq \alpha\beta^*/(4\pi), \tag{7.50}$$

while for $\mathbb{P}_{\mathcal{S}}$ similarly we have

$$\sqrt{2\operatorname{Var}[q_t(\boldsymbol{X})] \cdot [\eta(\mathcal{Q}_{\mathscr{A}}) + \log(1/\xi)]/n} = \sqrt{\mathbb{E}_{\mathbb{P}_{\mathcal{S}}}[q_t(\boldsymbol{X})]\{1 - \mathbb{E}_{\mathbb{P}_{\mathcal{S}}}[q_t(\boldsymbol{X})]\} \cdot 2\log(d/\xi)/n}$$
$$\leq \sqrt{2[\log d + \log(1/\xi)]/n} \leq \alpha\beta^*/(4\pi). \tag{7.51}$$

Also, we have $2/(3n) \cdot \log(d/\xi)/n \leq \alpha\beta^*/(4\pi)$, since $\beta^{*2}n\alpha^2/[\log d + \log(1/\xi)] \to \infty$, $\beta^* = o(1)$, and $\alpha = o(1)$. Thus, we have

$$\overline{R}(\phi) = \overline{\mathbb{P}}_0\big[\sup_{t\in[T]}Z_{q_t} \geq 1 - \Phi(\beta^*/2) + \alpha\beta^*/(4\pi)\big] \tag{7.52}$$
$$+ \frac{1}{|\mathcal{C}|}\sum_{\mathcal{S}\in\mathcal{C}}\overline{\mathbb{P}}_{\mathcal{S}}\big[\sup_{t\in[T]}Z_{q_t} < 1 - \Phi(\beta^*/2) + \alpha\beta^*/(4\pi)\big]$$
$$\leq \overline{\mathbb{P}}_0\big\{\sup_{t\in[T]}|Z_{q_t} - \mathbb{E}_{\mathbb{P}_0}[q_t(\boldsymbol{X})]| > \alpha\beta^*/(4\pi)\big\}$$
$$+ \frac{1}{|\mathcal{C}|}\sum_{\mathcal{S}\in\mathcal{C}}\overline{\mathbb{P}}_{\mathcal{S}}\big\{\sup_{t\in[T]}|Z_{q_t} - \mathbb{E}_{\mathbb{P}_{\mathcal{S}}}[q_t(\boldsymbol{X})]| > \alpha\beta^*/(4\pi)\big\} \leq 2\xi,$$

where the first and second inequalities are from (7.49)-(7.51) and Definition 3.1.

**Setting (iii):** For $\beta^{*2}s^{*2}n\alpha^2/[d\log(1/\xi)] \to \infty$, the queries and test are defined in (5.6) and (5.7). In the following, we consider two cases: $\beta^*s^*/\sqrt{d} = O(1)$ and $\beta^*s^*/\sqrt{d} \to \infty$. First, for $\beta^*s^*/\sqrt{d} = O(1)$ we have

$$\mathbb{E}_{\mathbb{P}_{\mathcal{S}}}[q(\boldsymbol{X})] - \mathbb{E}_{\mathbb{P}_0}[q(\boldsymbol{X})] = \alpha\big[\mathbb{P}'_{\mathcal{S}}\big(\textstyle\sum_{j=1}^d X_j \geq \beta^*s^*/2\big) - \mathbb{P}_0\big(\textstyle\sum_{j=1}^d X_j \geq \beta^*s^*/2\big)\big]$$
$$= \alpha\mathbb{P}_0\big(\big|\textstyle\sum_{j=1}^d X_j\big| \leq \beta^*s^*/2\big) = \alpha\int_{-\beta^*s^*/(2\sqrt{d})}^{\beta^*s^*/(2\sqrt{d})} \frac{\exp(-x^2/2)}{\sqrt{2\pi}}\mathrm{d}x$$
$$\geq \frac{\alpha\beta^*s^*}{\sqrt{2\pi d}}\exp\big[-\beta^{*2}s^{*2}/(8d)\big] \geq C\alpha\beta^*s^*/\sqrt{d}, \tag{7.53}$$

where in the last inequality we use the fact that $\exp\big[-\beta^{*2}s^{*2}/(8d)\big] \geq C\sqrt{2\pi}$, where $C$ is an absolute constant, since $\beta^*s^*/\sqrt{d} = O(1)$. Recall that we have $\eta(\mathcal{Q}_{\mathscr{A}}) = 0$. Since $\beta^{*2}s^{*2}n\alpha^2/[d\log(1/\xi)] \to \infty$, for $\mathbb{P}_0$ we have

$$\sqrt{2\operatorname{Var}[q(\boldsymbol{X})] \cdot [\eta(\mathcal{Q}_{\mathscr{A}}) + \log(1/\xi)]/n} \leq \sqrt{2\log(1/\xi)/n} \leq C/2 \cdot \alpha\beta^*s^*/\sqrt{d},$$

while similarly for $\mathbb{P}_{\mathcal{S}}$ we have

$$\sqrt{2\operatorname{Var}[q(\boldsymbol{X})] \cdot [\eta(\mathcal{Q}_{\mathscr{A}}) + \log(1/\xi)]/n} \leq \sqrt{2\log(1/\xi)/n} \leq C/2 \cdot \alpha\beta^*s^*/\sqrt{d}.$$

Also, we have $2/(3n) \cdot \log(1/\xi) \leq 2/3 \cdot \sqrt{\log(1/\xi)/n} \leq C/2 \cdot \alpha \beta^* s^*/\sqrt{d}$. Here $C$ is the same constant as in (7.53). Therefore, by the same derivation of (7.48), the risk is at most $2\xi$. In the second setting, for $\beta^* s^*/\sqrt{d} \to \infty$ we have

$$\mathbb{E}_{\mathbb{P}_{\mathcal{S}}}[q(\boldsymbol{X})] - \mathbb{E}_{\mathbb{P}_0}[q(\boldsymbol{X})] = \alpha \big[\mathbb{P}'_{\mathcal{S}}\big(\textstyle\sum_{j=1}^d X_j \geq \beta^* s^*/2\big) - \mathbb{P}_0\big(\textstyle\sum_{j=1}^d X_j \geq \beta^* s^*/2\big)\big] \geq \alpha/4.$$

Here we use $\mathbb{P}'_{\mathcal{S}}\big(\sum_{j=1}^d X_j \geq \beta^* s^*/2\big) \geq 1/2$ and $\mathbb{P}_0\big(\sum_{j=1}^d X_j \geq \beta^* s^*/2\big) = o(1)$, since $1/\sqrt{d} \cdot \sum_{j=1}^d X_j$ follows $N(0,1)$ under $\mathbb{P}_0$ and $N(\beta^* s^*/\sqrt{d}, 1)$ under $\mathbb{P}'_{\mathcal{S}}$. Note that we have

$$\mathbb{E}_{\mathbb{P}_0}[q(\boldsymbol{X})] \leq \mathbb{E}_{\mathbb{P}_{\mathcal{S}}}[q(\boldsymbol{X})] = \alpha \mathbb{P}'_{\mathcal{S}}\big(\textstyle\sum_{j=1}^d X_j \geq \beta^* s^*/2\big) + (1-\alpha)\mathbb{P}_0\big(\textstyle\sum_{j=1}^d X_j \geq \beta^* s^*/2\big)$$
$$\leq \alpha + \exp\big[-\beta^{*2} s^{*2}/(8d)\big] \leq 1/2.$$

In the second last inequality, we use $1/\sqrt{d} \cdot \sum_{j=1}^d X_j \sim N(0,1)$ under $\mathbb{P}_0$, while in the last inequality we use $\alpha = o(1)$ and $\beta^* s^*/\sqrt{d} \to \infty$. Recall that $\eta(\mathcal{Q}_{\mathscr{A}}) = 0$. Then for both $\mathbb{P}_0$ and $\mathbb{P}_{\mathcal{S}}$ we have

$$\sqrt{2 \operatorname{Var}[q(\boldsymbol{X})] \cdot [\eta(\mathcal{Q}_{\mathscr{A}}) + \log(1/\xi)]/n} \leq \sqrt{\big\{\alpha + \exp\big[-\beta^{*2} s^{*2}/(8d)\big]\big\} \cdot 2 \log(1/\xi)/n}.$$

In the sequel, we prove that the right-hand side is upper bounded by $\alpha/8$. Since $\alpha n \geq C \log(1/\xi)$ where $C$ is sufficiently large, we have $2\alpha \log(1/\xi)/n \leq \alpha^2/128$. Meanwhile, since $\beta^{*2} s^{*2} n \alpha^2/[d \log(1/\xi)] \to \infty$ and $\beta^* s^*/\sqrt{d} \to \infty$, we have

$$\exp\big[-\beta^{*2} s^{*2}/(8d)\big] \cdot 2 \log(1/\xi) \leq 8d/\big(\beta^{*2} s^{*2}\big) \cdot 2 \log(1/\xi) \leq n\alpha^2/128,$$

where the first inequality follows from the fact that $\exp(-x) \leq 1/x$ for $x > 0$. In summary, we have

$$\sqrt{\big\{\alpha + \exp\big[-\beta^{*2} s^{*2}/(8d)\big]\big\} \cdot 2 \log(1/\xi)/n} \leq \alpha/8.$$

Following the same derivation of (7.48), we obtain that the risk is at most $2\xi$.

**Setting (iv):** For $\beta^{*2} s^*/\log n \to \infty$ and $\beta^{*2} n\alpha/(\log d \cdot \log n) \to \infty$, we consider two cases:

(a) $C \beta^{*2} s^*/\log n < \beta^{*2} n\alpha/(\log d \cdot \log n)$ for a sufficiently large constant $C$, i.e., $s^* < n\alpha/(C \log d)$. We consider the query functions and test in (5.8) and (5.9). For $\mathbb{P}_0$, we have $\sum_{j \in \mathcal{S}_t} X_j/\sqrt{s^*} \sim N(0,1)$. Since $\beta^{*2} s^*/\log n \to \infty$, we have

$$\mathbb{E}_{\mathbb{P}_0}[q_t(\boldsymbol{X})] = \mathbb{P}_0\big(\textstyle\sum_{j \in \mathcal{S}_t} X_j \geq \beta^* s^*/2\big) = 1 - \Phi\big(\beta^* \sqrt{s^*}/2\big) \qquad (7.54)$$
$$\leq \exp\big(-\beta^{*2} s^*/8\big) \leq \exp(-\log n) = 1/n.$$

Meanwhile, for $\mathbb{P}_{\mathcal{S}}$ with $\mathcal{S} = \mathcal{S}_t$, we have $\sum_{j \in \mathcal{S}_t} X_j/\sqrt{s^*} \sim N\big(\beta^* \sqrt{s^*}, 1\big)$. Then we have

$$\mathbb{E}_{\mathbb{P}_{\mathcal{S}}}[q_t(\boldsymbol{X})] = \mathbb{P}_{\mathcal{S}}\big(\textstyle\sum_{j \in \mathcal{S}_t} X_j \geq \beta^* s^*/2\big) = \alpha \mathbb{P}'_{\mathcal{S}}\big(\textstyle\sum_{j \in \mathcal{S}_t} X_j \geq \beta^* s^*/2\big) + (1-\alpha) \mathbb{E}_{\mathbb{P}_0}[q_t(\boldsymbol{X})]$$
$$\leq \alpha + (1-\alpha)/n \leq \alpha + 1/n \leq 2\alpha. \qquad (7.55)$$

Here we use the assumption that $\alpha n \geq C'$ where $C'$ is sufficiently large. Recall that $\eta(\mathcal{Q}_{\mathscr{A}}) = \log \binom{d}{s^*}$. We have

$$[\eta(\mathcal{Q}_{\mathscr{A}}) + \log(1/\xi)]/n \leq [2s^* \log d + \log(1/\xi)]/n \leq (\alpha n)/(128n) < \alpha/128,$$

since $s^* < n\alpha/(C\log d)$, where $C$ is sufficiently large, and $\log(1/\xi) < s^*\log d$. For $\mathbb{P}_0$, we have

$$\sqrt{2\operatorname{Var}[q_t(\boldsymbol{X})]\cdot[\eta(\mathcal{Q}_{\mathscr{A}})+\log(1/\xi)]/n} \leq \sqrt{2[s^*\log d + \log(1/\xi)]/n^2}$$
$$\leq \sqrt{4s^*\log d/n^2} \leq \sqrt{\alpha/(32n)} < \alpha/4$$

according to (7.54). For $\mathbb{P}_{\mathcal{S}}$, we have

$$\sqrt{\operatorname{Var}[q_t(\boldsymbol{X})]\cdot[\eta(\mathcal{Q}_{\mathscr{A}})+\log(1/\xi)]/n} \leq \sqrt{2\alpha[s^*\log d + \log(1/\xi)]/n}$$
$$\leq \sqrt{8\alpha s^*\log d/n} \leq \sqrt{\alpha^2/16} = \alpha/4$$

according to (7.55). Meanwhile, since $\sum_{j\in\mathcal{S}_t}X_j/\sqrt{s^*} \sim N(\beta^*\sqrt{s^*},1)$, for $\mathcal{S}=\mathcal{S}_t$ we have

$$\mathbb{E}_{\mathbb{P}_{\mathcal{S}}}[q(\boldsymbol{X})] - \mathbb{E}_{\mathbb{P}_0}[q(\boldsymbol{X})] = \alpha\big[\mathbb{P}'_{\mathcal{S}}\big(\textstyle\sum_{j\in\mathcal{S}_t}X_j \geq \beta^*s^*/2\big) - \mathbb{P}_0\big(\sum_{j\in\mathcal{S}_t}X_j \geq \beta^*s^*/2\big)\big]$$
$$= \alpha\mathbb{P}_0\big(\big|\textstyle\sum_{j\in\mathcal{S}_t}X_j\big| \leq \beta^*s^*/2\big) = \alpha\int_{-\beta^*\sqrt{s^*}/2}^{\beta^*\sqrt{s^*}/2}\frac{\exp(-x^2/2)}{\sqrt{2\pi}}\mathrm{d}x \geq \alpha/2.$$

Here the last inequality holds because $\beta^*\sqrt{s^*} \to \infty$. By following a similar derivation of (7.52) we obtain that the risk is at most $2\xi$.

(b) $C\beta^{*2}s^*/\log n \geq \beta^{*2}n\alpha/(\log d \cdot \log n)$, i.e., $s^* \geq n\alpha/(C\log d)$, where $C$ is the same constant in case (a). We consider the query functions and test in (5.10) and (5.11). By the same derivation of case (a) with $s^*$ replaced by $\bar{s}^*$, we obtain the same conclusion.

Combining settings (i)-(iv), we obtain the conclusion of Theorem 5.5. $\square$

## 7.9 Proof of Lemma 5.6

*Proof.* For notational simplicity, we abbreviate $k(q,\mathcal{S})$ as $k$ and $\bar{\mathcal{C}}(q,\mathcal{S})$ as $\bar{\mathcal{C}}$. We define $\sqrt{\log(1/\xi)/n}$ as $\tau$. Similar to (7.17) in the proof of Lemma 5.2, we have

$$\mathbb{E}_{\mathcal{S}'\sim\mathbb{Q}_{\bar{c}}}[h(|\mathcal{S}\cap\mathcal{S}'|)] \leq \frac{\alpha^2\cdot\sum_{j=0}^{k-1}|\mathcal{C}_j(\mathcal{S})|\exp\big[(s^*-j)\beta^{*2}\big]}{\sum_{j=0}^{k-1}|\mathcal{C}_j(\mathcal{S})|} + (1-\alpha^2).$$

Note that for perfect matching we have

$$|\mathcal{C}_j(\mathcal{S})| = \binom{s^*}{s^*-j}\sum_{\ell=0}^{j}\frac{(-1)^\ell j!}{\ell!}, \tag{7.56}$$

which is obtained using the number of derangements. Thus, for any $j \in \{2,\ldots,k-1\}$ ($k \leq s^*$) we have

$$|\mathcal{C}_{j+1}(\mathcal{S})|/|\mathcal{C}_j(\mathcal{S})| = \frac{\sum_{\ell=0}^{j+1}(-1)^\ell/\ell!}{\sum_{\ell=0}^{j}(-1)^\ell/\ell!}\cdot(s^*-j) \geq 2(s^*-k+1)/3, \tag{7.57}$$

since $\left[\sum_{\ell=0}^{j+1}(-1)^\ell/\ell!\right]/\left[\sum_{\ell=0}^{j}(-1)^\ell/\ell!\right] \geq 2/3$ for any $j \geq 2$, where the inequality is tight when $j = 2$. Meanwhile, by our assumption there exists a sufficiently small constant $\delta > 0$ such that

$$\log(1 + \tau^2/\alpha^2)/\beta^{*2} \geq 3d^\delta/2 + 1. \tag{7.58}$$

We define $\bar{k} = s^* + 1 - 3d^{\delta/2}/2$. Using the same derivation of (7.22) in the proof of Lemma 5.2 with $\zeta^{-1}$ replaced by $d^{-\delta/2}$, we have

$$\log\left[\frac{1 - d^{-\delta/2}}{1 - d^{-\delta/2}\exp(\beta^{*2})}\right]\bigg/\beta^{*2} = O(d^{-\delta/2}) \leq 1.$$

Then from (7.58) we have

$$s^* - \bar{k} + 1 = 3d^{\delta/2}/2 \leq \log(1 + \tau^2/\alpha^2)/\beta^{*2} - 1$$
$$\leq \log(1 + \tau^2/\alpha^2)/\beta^{*2} - \log\left[\frac{1 - d^{-\delta/2}}{1 - d^{-\delta/2}\exp(\beta^{*2})}\right]\bigg/\beta^{*2},$$

from which we obtain

$$\frac{\alpha^2 \cdot \exp\left[(s^* - \bar{k} + 1)\beta^{*2}\right] \cdot (1 - d^{-\delta/2})}{1 - d^{-\delta/2}\exp(\beta^{*2})} + (1 - \alpha^2) \leq 1 + \tau^2. \tag{7.59}$$

Meanwhile, in the following proof we use the notation

$$f(k) = \frac{\alpha^2 \cdot \sum_{j=0}^{k-1}|\mathcal{C}_j(\mathcal{S})|\exp\left[(s^* - j)\beta^{*2}\right]}{\sum_{j=0}^{k-1}|\mathcal{C}_j(\mathcal{S})|}. \tag{7.60}$$

Note that we have

$$f(\bar{k}) + (1 - \alpha^2) = \frac{\alpha^2 \cdot \left\{\exp(s^*\beta^{*2}) + \sum_{j=2}^{\bar{k}-1}|\mathcal{C}_j(\mathcal{S})|\exp\left[(s^* - j)\beta^{*2}\right]\right\}}{1 + \sum_{j=2}^{\bar{k}-1}|\mathcal{C}_j(\mathcal{S})|} + (1 - \alpha^2) \tag{7.61}$$

$$\leq \frac{\alpha^2 \cdot \left\{\exp(s^*\beta^{*2}) + \sqrt{s^*(s^* - 1)/2}\cdot\exp\left[(s^* - 1)\beta^{*2}\right] + \sum_{j=2}^{\bar{k}-1}|\mathcal{C}_j(\mathcal{S})|\exp\left[(s^* - j)\beta^{*2}\right]\right\}}{1 + \sqrt{s^*(s^* - 1)/2} + \sum_{j=2}^{\bar{k}-1}|\mathcal{C}_j(\mathcal{S})|} + (1 - \alpha^2)$$

$$\leq \frac{\alpha^2 \cdot \sum_{j=0}^{\bar{k}-1}d^{\delta j/2}\exp\left[(s^* - j)\beta^{*2}\right]}{\sum_{j=0}^{\bar{k}-1}d^{\delta j/2}} + (1 - \alpha^2) \leq \frac{\alpha^2 \cdot \exp\left[(s^* - \bar{k} + 1)\beta^{*2}\right] \cdot (1 - d^{-\delta/2})}{1 - d^{-\delta/2}\exp(\beta^{*2})} + (1 - \alpha^2)$$

$$\leq 1 + \tau^2,$$

where the first equality in (7.61) holds because $|\mathcal{C}_0(\mathcal{S})| = 1$ and $|\mathcal{C}_1(\mathcal{S})| = 0$ according to (7.56). The first inequality in (7.61) holds because

$$f(\bar{k}) \leq f(3) = \frac{\alpha^2 \cdot \left\{\exp(s^*\beta^{*2}) + |\mathcal{C}_2(\mathcal{S})|\exp\left[(s^* - 2)\beta^{*2}\right]\right\}}{1 + |\mathcal{C}_2(\mathcal{S})|} \leq \alpha^2\exp\left[(s^* - 1)\beta^{*2}\right],$$

where the first inequality is from the definition of $\bar{k}$ and the last inequality can be verified by noting that $|\mathcal{C}_2(\mathcal{S})| = s^*(s^* - 1)/2$ and $\beta^* \in (0, 1]$. Meanwhile, the second inequality in (7.61) follows from

Lemma 7.1, (7.57), the definition of $\bar k$, and $\sqrt{s^*(s^*-1)/2}\geq d^{\delta/2}$. In addition, the last inequality in (7.61) follows from (7.59). Note that for any $k\leq \bar k$, by the same derivation of (7.61) we have

$$\mathbb{E}_{\mathcal{S}'\sim\mathbb{Q}_{\bar{\mathcal{C}}}}[h(|\mathcal{S}\cap\mathcal{S}'|)] \leq f(k) + (1-\alpha^2) \leq \frac{\alpha^2 \cdot \exp\left[(s^*-k+1)\beta^{*2}\right]\cdot(1-d^{-\delta/2})}{1-d^{-\delta/2}\exp(\beta^{*2})} + (1-\alpha^2). \quad (7.62)$$

Since (7.62) holds for any $\mathcal{S}\in\mathcal{C}(q)$ and

$$\sup_{\mathcal{S}\in\mathcal{C}(q)}\left\{\mathbb{E}_{\mathcal{S}'\sim\mathbb{Q}_{\bar{\mathcal{C}}}}[h(|\mathcal{S}\cap\mathcal{S}'|)]\right\} \geq 1+\tau^2, \quad (7.63)$$

we have $f(k)+(1-\alpha^2)\geq 1+\tau^2$. Therefore we have $k\leq \bar k$, since $f(k)$ in (7.60) is nonincreasing for $k\leq \bar k$. Combining (7.62) and (7.63) we have

$$s^* - k + 1 \geq \log(1+\tau^2/\alpha^2)/\beta^{*2} - \log\left[\frac{1-d^{-\delta/2}}{1-d^{-\delta/2}\exp(\beta^{*2})}\right]\Big/\beta^{*2} \geq \log(1+\tau^2/\alpha^2)/\beta^{*2} - 1$$
$$\geq 3d^\delta/2 + 1 = d^{\delta/2}(s^* - \bar k + 1) + 1,$$

which implies

$$\bar k - k \geq (d^{\delta/2}-1)(s^* - \bar k + 1) = 3(d^{\delta/2}-1)d^{\delta/2}/2 \geq 3d^\delta/4.$$

Following the same derivation of (7.32) in the proof of Lemma 5.2 with $\gamma$ replaced by $d^\delta$, we obtain

$$|\mathcal{C}(q)| = |\bar{\mathcal{C}}| \leq 2d^{-\delta(\bar k - k)/2}|\mathcal{C}| \leq 2\exp(-\delta\log d \cdot 3d^\delta/8)|\mathcal{C}|.$$

Thus we conclude the proof of Lemma 5.6. $\square$

### 7.10 Proof of Theorem 5.7

*Proof.* We use the notation in the proof of Lemma 5.6. We assume $T = O(d^\eta)$, where $\eta$ is a constant, and $T\geq 1$. Let $\delta > 0$ be a sufficiently small constant. We consider the following settings.

(i) $\tau^2/\alpha^2 = o(1)$. As long as $\tau^2/(2d^\delta\alpha^2\beta^{*2})\to\infty$, we have

$$\log(1+\tau^2/\alpha^2)/\beta^{*2} \geq \tau^2/(2\alpha^2\beta^{*2}) \geq 3d^\delta/2 + 1,$$

which implies the assumption of Lemma 5.6. Therefore, by Lemma 5.6 we have

$$\frac{T\cdot \sup_{q\in\mathcal{Q}}|\mathcal{C}(q)|}{|\mathcal{C}|} = O[d^\eta \exp(-\delta\log d \cdot 3d^\delta/8)] = o(1). \quad (7.64)$$

(ii) $\lim_{d\to\infty}\tau^2/\alpha^2 > 0$. As long as $\beta^* = o(d^{-\delta})$, it holds that $\log(1+\tau^2/\alpha^2)/\beta^{*2}\geq 3d^\delta/2+1$. By Lemma 5.6 we obtain (7.64).

Thus, we obtain the conclusion of Theorem 5.7. $\square$

## 7.11 Proof of Lemma 6.1

*Proof.* For $\text{supp}(\mathbf{v}^*) = \mathcal{S}$, we use the notation $\mathbf{v}^* = \mathbf{v}_\mathcal{S}$ and
$$\boldsymbol{\Sigma}_\mathcal{S} = \mathbf{I}_d + \beta^* \mathbf{v}_\mathcal{S} \mathbf{v}_\mathcal{S}^\top.$$

By the definition of $\mathbb{P}_{\mathcal{S}_1}$, $\mathbb{P}_{\mathcal{S}_2}$, and $\mathbb{P}_0$ we have
$$\frac{d\mathbb{P}_{\mathcal{S}_1}}{d\mathbb{P}_0}(\mathbf{x}) = \frac{\exp(-\mathbf{x}^\top \boldsymbol{\Sigma}_{\mathcal{S}_1}^{-1} \mathbf{x}/2)}{\det(\boldsymbol{\Sigma}_{\mathcal{S}_1})^{1/2} \exp(-\mathbf{x}^\top \mathbf{x}/2)}, \tag{7.65}$$

and the same holds for $d\mathbb{P}_{\mathcal{S}_2}/d\mathbb{P}_0$. By Sylvester's determinant theorem, i.e., $\det(\mathbf{I}_d + \mathbf{u}\mathbf{u}^\top) = 1 + \mathbf{u}^\top \mathbf{u}$ for any $\mathbf{u} \in \mathbb{R}^d$, we have
$$\det(\boldsymbol{\Sigma}_{\mathcal{S}_1}) = \det(\mathbf{I}_d + \beta^* \mathbf{v}_{\mathcal{S}_1} \mathbf{v}_{\mathcal{S}_1}^\top) = 1 + \beta^* \mathbf{v}_{\mathcal{S}_1}^\top \mathbf{v}_{\mathcal{S}_1} = 1 + \beta^*, \tag{7.66}$$

where the last equality follows from $\|\mathbf{v}_{\mathcal{S}_1}\|_2 = 1$. Moreover, by Sherman-Morrison formula, i.e.,
$$(\mathbf{B} + \mathbf{u}\mathbf{u}^\top)^{-1} = \mathbf{B}^{-1} - (\mathbf{B}^{-1}\mathbf{u}\mathbf{u}^\top \mathbf{B}^{-1})/(1 - \mathbf{u}^\top \mathbf{B}^{-1}\mathbf{u})$$

for any $\mathbf{B} \in \mathbb{R}^{d \times d}, \mathbf{u} \in \mathbb{R}^d$, we have
$$\boldsymbol{\Sigma}_{\mathcal{S}_1}^{-1} = (\mathbf{I}_d + \beta^* \mathbf{v}_{\mathcal{S}_1} \mathbf{v}_{\mathcal{S}_1}^\top)^{-1} = \mathbf{I}_d - \frac{\beta^* \mathbf{v}_{\mathcal{S}_1} \mathbf{v}_{\mathcal{S}_1}^\top}{1 + \beta^*}, \tag{7.67}$$

and the same holds for $\boldsymbol{\Sigma}_{\mathcal{S}_2}$. Substituting (7.66) and (7.67) into (7.65), we obtain
$$\frac{d\mathbb{P}_{\mathcal{S}_1}}{d\mathbb{P}_0}(\mathbf{x}) = \frac{1}{\sqrt{1+\beta^*}} \exp\left[\frac{\beta^*}{2(1+\beta^*)}(\mathbf{x}^\top \mathbf{v}_{\mathcal{S}_1})^2\right].$$

Similarly, we can calculate $d\mathbb{P}_{\mathcal{S}_2}/d\mathbb{P}_0(\mathbf{x})$. Therefore, we have
$$\frac{d\mathbb{P}_{\mathcal{S}_1}}{d\mathbb{P}_0} \frac{d\mathbb{P}_{\mathcal{S}_2}}{d\mathbb{P}_0}(\mathbf{x}) = \frac{1}{1+\beta^*} \exp\left\{\frac{\beta^*}{2(1+\beta^*)}\left[(\mathbf{x}^\top \mathbf{v}_{\mathcal{S}_1})^2 + (\mathbf{x}^\top \mathbf{v}_{\mathcal{S}_2})^2\right]\right\}$$
$$= \frac{1}{1+\beta^*} \exp\left[\frac{\beta^*}{2(1+\beta^*)} \mathbf{x}^\top \mathbf{A} \mathbf{x}\right],$$

where $\mathbf{A} = \mathbf{v}_{\mathcal{S}_1} \mathbf{v}_{\mathcal{S}_1}^\top + \mathbf{v}_{\mathcal{S}_2} \mathbf{v}_{\mathcal{S}_2}^\top$. Note that $\mathbf{A}$ is the summation of two rank one matrices. Thus, its top two leading eigenvalues are
$$\lambda_1(\mathbf{A}) = 1 + \mathbf{v}_{\mathcal{S}_1}^\top \mathbf{v}_{\mathcal{S}_2} \quad \text{and} \quad \lambda_2(\mathbf{A}) = 1 - \mathbf{v}_{\mathcal{S}_1}^\top \mathbf{v}_{\mathcal{S}_2}, \tag{7.68}$$

and the rest eigenvalues are all zero. Let the eigenvalue decomposition of $\mathbf{A}$ be $\mathbf{A} = \mathbf{U}\boldsymbol{\Lambda}\mathbf{U}^\top$, where $\boldsymbol{\Lambda} = \text{diag}\{\lambda_1(\mathbf{A}), \lambda_2(\mathbf{A}), \ldots, 0\}$. We have
$$\mathbb{E}_{\mathbb{P}_0}\left[\frac{d\mathbb{P}_{\mathcal{S}_1}}{d\mathbb{P}_0} \frac{d\mathbb{P}_{\mathcal{S}_2}}{d\mathbb{P}_0}(\mathbf{X})\right] = \frac{1}{1+\beta^*} \int (2\pi)^{-d/2} \exp\left[\frac{\beta^*}{2(1+\beta^*)} \mathbf{x}^\top \mathbf{A} \mathbf{x} - \frac{\mathbf{x}^\top \mathbf{x}}{2}\right] d\mathbf{x}$$
$$= \frac{1}{1+\beta^*} \int (2\pi)^{-d/2} \exp\left[\frac{\beta^*}{2(1+\beta^*)} \bar{\mathbf{x}}^\top \boldsymbol{\Lambda} \bar{\mathbf{x}} - \frac{\bar{\mathbf{x}}^\top \bar{\mathbf{x}}}{2}\right] |\det(\mathbf{U})| d\bar{\mathbf{x}}$$
$$= \frac{1}{1+\beta^*} \int (2\pi)^{-d/2} \exp\left\{\frac{\beta^*}{2(1+\beta^*)}\left[\lambda_1(\mathbf{A})\bar{x}_1^2 + \lambda_2(\mathbf{A})\bar{x}_2^2\right] - \frac{\bar{\mathbf{x}}^\top \bar{\mathbf{x}}}{2}\right\} d\bar{\mathbf{x}},$$

where $\bar{\mathbf{x}} = \mathbf{U}^\top \mathbf{x}$. Here we use the fact that $\bar{\mathbf{x}}^\top \bar{\mathbf{x}} = \mathbf{x}^\top \mathbf{U}\mathbf{U}^\top \mathbf{x} = \mathbf{x}^\top \mathbf{x}$. Meanwhile, note that

$$\int (2\pi)^{-d/2} \exp\left[\frac{\beta^*}{2(1+\beta^*)}\lambda_1(\mathbf{A})\bar{x}_1^2 + \lambda_2(\mathbf{A})\bar{x}_2^2 - \frac{\bar{\mathbf{x}}^\top \bar{\mathbf{x}}}{2}\right] d\bar{\mathbf{x}}$$

$$= \underbrace{\int (2\pi)^{-1/2} \exp\left[\frac{\beta^*}{2(1+\beta^*)}\lambda_1(\mathbf{A})\bar{x}_1^2 - \frac{\bar{x}_1^2}{2}\right] d\bar{\mathbf{x}}}_{\text{(i)}} \cdot \underbrace{\int (2\pi)^{-1/2} \exp\left[\frac{\beta^*}{2(1+\beta^*)}\lambda_2(\mathbf{A})\bar{x}_2^2 - \frac{\bar{x}_2^2}{2}\right] d\bar{\mathbf{x}}}_{\text{(ii)}}$$

$$\cdot \underbrace{\int (2\pi)^{-(d-2)/2} \exp\left(-\sum_{j=3}^d \bar{x}_j^2/2\right) d\bar{\mathbf{x}}}_{\text{(iii)}}.$$

By calculation, we have that term (i) equals $[1 - \lambda_1(\mathbf{A})\beta^*/(1+\beta^*)]^{-1/2}$. Similarly, term (ii) equals $[1 - \lambda_2(\mathbf{A})\beta^*/(1+\beta^*)]^{-1/2}$. In addition, term (iii) is one. Therefore, we have

$$\mathbb{E}_{\mathbb{P}_0}\left[\frac{d\mathbb{P}_{\mathcal{S}_1}}{d\mathbb{P}_0}\frac{d\mathbb{P}_{\mathcal{S}_2}}{d\mathbb{P}_0}(\boldsymbol{X})\right] = \frac{1}{1+\beta^*}\left[1 - \frac{\lambda_1(\mathbf{A})\beta^*}{(1+\beta^*)}\right]^{-1/2}\left[1 - \frac{\lambda_2(\mathbf{A})\beta^*}{(1+\beta^*)}\right]^{-1/2}$$

$$= \frac{1}{1+\beta^*}\left[1 - \frac{(1+|\mathcal{S}_1 \cap \mathcal{S}_2|/s^*)\beta^*}{(1+\beta^*)}\right]^{-1/2}\left[1 - \frac{(1-|\mathcal{S}_1 \cap \mathcal{S}_2|/s^*)\beta^*}{(1+\beta^*)}\right]^{-1/2}$$

$$= \left(1 - \frac{\beta^{*2}|\mathcal{S}_1 \cap \mathcal{S}_2|^2}{s^{*2}}\right)^{-1/2},$$

where the second equality is from the fact that $\lambda_1(\mathbf{A}) = 1 + |\mathcal{S}_1 \cap \mathcal{S}_2|/s^*$ and $\lambda_2(\mathbf{A}) = 1 - |\mathcal{S}_1 \cap \mathcal{S}_2|/s^*$ according to (7.68). Thus, we conclude the proof of Lemma 6.1. $\square$

## 7.12 Proof of Lemma 6.2

*Proof.* By Lemma 6.1 we have

$$h(|\mathcal{S} \cap \mathcal{S}'|) = \left(1 - \frac{\beta^{*2}|\mathcal{S}_1 \cap \mathcal{S}_2|^2}{s^{*2}}\right)^{-1/2}. \tag{7.69}$$

Following the notation in the proof of Lemma 5.2, we denote $k(q, \mathcal{S})$ as $k$ and $\bar{\mathcal{C}}(q, \mathcal{S})$ as $\bar{\mathcal{C}}$. Following the notation in the proof of Theorem 4.4, for any $\mathcal{S} \in \mathcal{C}(q)$ we have

$$\mathbb{E}_{\mathcal{S}' \sim \mathbb{Q}_{\bar{\mathcal{C}}}}[h(|\mathcal{S} \cap \mathcal{S}'|)] = \frac{\sum_{j=0}^{k-1} |\mathcal{C}_j(\mathcal{S})|\left[1 - \beta^{*2}(s^* - j)^2/s^{*2}\right]^{-1/2} + |\mathcal{C}'_k(\mathcal{S})|\left[1 - \beta^{*2}(s^* - k)^2/s^{*2}\right]^{-1/2}}{\sum_{j=0}^{k-1} |\mathcal{C}_j(\mathcal{S})| + |\mathcal{C}'_k(\mathcal{S})|}$$

$$\leq \frac{\sum_{j=0}^{k-1} |\mathcal{C}_j(\mathcal{S})|\left[1 - \beta^{*2}(s^* - j)^2/s^{*2}\right]^{-1/2}}{\sum_{j=0}^{k-1} |\mathcal{C}_j(\mathcal{S})|}.$$

Recall that we set $d/s^{*2} = d^{2\delta}$. Note that for any $j \in \{0, \ldots k-1\}$, we have

$$\frac{|\mathcal{C}_{j+1}(\mathcal{S})|}{|\mathcal{C}_j(\mathcal{S})|} = \frac{(s^* - j)(d - s^* - j)}{(j+1)^2} \geq \frac{d - 2s^* + 1}{s^{*2}} = d^{2\delta} - 2/s^* + 1/s^{*2} \geq d^\delta, \tag{7.70}$$

where we use the fact that $1/s^* = o(1)$. By Lemma 7.1, we have

$$\mathbb{E}_{\mathcal{S}'\sim\mathbb{Q}_{\bar{c}}}[h(|\mathcal{S}\cap\mathcal{S}'|)] \leq \frac{\sum_{j=0}^{k-1}|\mathcal{C}_j|\left[1-(s^*-j)^2\beta^{*2}/s^{*2}\right]^{-1/2}}{\sum_{j=0}^{k-1}|\mathcal{C}_j|} \leq \frac{\sum_{j=0}^{k-1}\left[1-(s^*-j)^2\beta^{*2}/s^{*2}\right]^{-1/2}d^{\delta j}}{\sum_{j=0}^{k-1}d^{\delta j}}$$

$$= \frac{\sum_{j=0}^{k-1}\left[1-(s^*-j)^2\beta^{*2}/s^{*2}\right]^{-1/2}d^{\delta(j-k+1)}}{\sum_{j=0}^{k-1}d^{\delta(j-k+1)}}. \tag{7.71}$$

According to (7.69) we have

$$\frac{h[s^*-(j-1)]}{h(s^*-j)} = \left[\frac{s^{*2}-(s^*-j+1)^2\beta^{*2}}{s^{*2}-(s^*-j)^2\beta^{*2}}\right]^{-1/2}$$

$$\geq \left[\frac{s^{*2}-(s^*+1)^2\beta^{*2}}{s^{*2}-s^{*2}\beta^{*2}}\right]^{-1/2} = \left[1-\frac{(1+2s^*)\beta^{*2}}{s^{*2}-s^{*2}\beta^{*2}}\right]^{-1/2}.$$

Let the right-hand side be $\bar{\gamma}$. We have $\bar{\gamma} > 1$ and

$$\bar{\gamma} = 1 + O\left[\frac{(1+2s^*)\beta^{*2}}{s^{*2}-s^{*2}\beta^{*2}}\right] = 1 + O(\beta^{*2}/s^*). \tag{7.72}$$

since $\beta^{*2}/s^* = o(1)$. Thus, we have $h(j) \leq \bar{\gamma}^{k-1-j}h(k-1)$, which together with (7.71) implies

$$\mathbb{E}_{\mathcal{S}'\sim\mathbb{Q}_{\bar{c}}}[h(|\mathcal{S}\cap\mathcal{S}'|)] \leq \left[1-\frac{(s^*-k+1)^2\beta^{*2}}{s^{*2}}\right]^{-1/2}\frac{\sum_{j=0}^{k-1}\bar{\gamma}^{k-1-j}d^{\delta(j-k+1)}}{\sum_{j=0}^{k-1}d^{\delta(j-k+1)}} \tag{7.73}$$

$$= \left[1-\frac{(s^*-k+1)^2\beta^{*2}}{s^{*2}}\right]^{-1/2}\frac{\sum_{j=0}^{k-1}(\bar{\gamma}^{-1}d^{\delta})^{(j-k+1)}}{\sum_{j=0}^{k-1}d^{\delta(j-k+1)}}.$$

On the right-hand side, by (7.72) we have

$$\frac{\sum_{j=0}^{k-1}(\bar{\gamma}^{-1}d^{\delta})^{(j-k+1)}}{\sum_{j=0}^{k-1}d^{\delta(j-k+1)}} = \frac{\sum_{j=0}^{k-1}(\bar{\gamma}^{-1}d^{\delta})^{-j}}{\sum_{j=0}^{k-1}d^{-\delta j}} = \frac{1-(\bar{\gamma}d^{-\delta})^k}{1-\bar{\gamma}d^{-\delta}}\bigg/\frac{1-(d^{-\delta})^k}{1-d^{-\delta}} \leq \frac{1-d^{-\delta}}{1-\bar{\gamma}d^{-\delta}}. \tag{7.74}$$

Plugging (7.74) into (7.73), we obtain

$$\mathbb{E}_{\mathcal{S}'\sim\mathbb{Q}_{\bar{c}}}[h(|\mathcal{S}\cap\mathcal{S}'|)] \leq \left[1-\frac{(s^*-k+1)^2\beta^{*2}}{s^{*2}}\right]^{-1/2}\frac{1-d^{-\delta}}{1-\bar{\gamma}d^{-\delta}}$$

$$\leq \left[1-\frac{(s^*-k+1)^2\beta^{*2}}{s^{*2}}\right]^{-1/2}[1+2d^{-\delta}(\bar{\gamma}-1)],$$

where we use the fact that $d^{-\delta} = o(1)$. Since (7.74) holds for any $\mathcal{S} \in \mathcal{C}(q)$ and

$$\sup_{\mathcal{S}\in\mathcal{C}(q)}\left\{\mathbb{E}_{\mathcal{S}'\sim\mathbb{Q}_{\bar{c}}}[h(|\mathcal{S}\cap\mathcal{S}'|)]\right\} \geq 1 + \tau^2,$$

by calculation we have

$$s^* - k + 1 \geq \frac{s^*}{\beta^*}\left\{1 - \left[\frac{1+\tau^2}{1+2d^{-\delta}(\bar{\gamma}-1)}\right]^{-2}\right\}^{1/2}.$$

Therefore, from (7.70) we have

$$|\bar{\mathcal{C}}| = \sum_{j=0}^{k}|\mathcal{C}_j(\mathcal{S})| \leq d^{-\delta s^*}|\mathcal{C}_{s^*}(\mathcal{S})|\sum_{j=0}^{k}d^{\delta j} \leq \frac{d^{-\delta(s^*-k)}|\mathcal{C}|}{1-d^{-\delta}}$$

$$\leq 2\exp\left[-\delta\log d \cdot \left(\frac{s^*}{\beta^*}\left\{1-\left[\frac{1+\tau^2}{1+2d^{-\delta}(\bar{\gamma}-1)}\right]^{-2}\right\}^{1/2}-1\right)\right]|\mathcal{C}|.$$

Here in the last inequality we use the fact that $d^{-\delta} = o(1)$. We conclude the proof of Lemma 6.2. □

## 7.13 Proof of Theorem 6.3

*Proof.* By Taylor expansion, we have $(1-x)^{-1/2} - 1 = x/2 + o(x^2)$, $[1-(1+x)^{-2}]^{1/2} = \sqrt{2x} + o(x^{3/2})$ for $x > 0$. Since $\beta^* = o(1)$, by (6.1) we have $\bar{\gamma} - 1 \asymp \beta^{*2}/s^*$, where $a \asymp b$ denotes that $a$ and $b$ are of the same order. Using the notation in Lemma 6.2, we have $\delta = 3/8$. Then we obtain

$$\left\{1-\left[\frac{1+\tau^2}{1+2d^{-\delta}(\bar{\gamma}-1)}\right]^{-2}\right\}^{1/2} = \left\{1-\left[1+\overbrace{\frac{\tau^2 - 2d^{-\delta}(\bar{\gamma}-1)}{1+2d^{-\delta}(\bar{\gamma}-1)}}^{\kappa}\right]^{-2}\right\}^{1/2} = \sqrt{2\kappa} + o(\kappa^{3/2}). \quad (7.75)$$

Here we use the fact $2d^{-\delta}(\bar{\gamma}-1) = o(\tau^2)$. To see this, recall that we assume $\beta^* = o\bigl[s^*\sqrt{\log(1/\xi)/n}\bigr]$, which implies

$$\bar{\gamma} - 1 \asymp \beta^{*2}/s^* = o[s^*\log(1/\xi)/n] = o(d^{\delta}\tau^2),$$

since $\tau = \sqrt{\log(1/\xi)/n}$, $s^* = d^{1/8}$, and $\delta = 3/8$. Hence we have $\kappa \asymp \tau^2 = o(1)$. Plugging it into (7.75) and then back into (6.2), we obtain that $T \cdot \sup_{q\in\mathcal{Q}}|\mathcal{C}(q)|/|\mathcal{C}| = o(1)$ for $s^*\tau/\beta^* \to \infty$ and $T = o(d^\eta)$. Thus, we conclude the proof of Theorem 6.3. □

## 7.14 Proof of Theorem 6.5

*Proof.* Recall that $\boldsymbol{X} \sim \mathbb{P}_\mathcal{S}$ if $\boldsymbol{X} \sim N(\boldsymbol{0}, \mathbf{I}_d + \beta^*\boldsymbol{v}^*\boldsymbol{v}^{*\top})$ in which $\text{supp}(\boldsymbol{v}^*) = \mathcal{S}$. In the following, we consider two settings.

**Setting (i):** For $\beta^{*2}n/[s^{*2}\log(d/\xi)] \to \infty$, we consider the query functions and test defined in (6.3) and (6.4). Note that under $\mathbb{P}_0$, we have $X_t^2 \sim \chi_1^2$ for any $t \in [d]$, since $X_t \sim N(0,1)$ under $\mathbb{P}_0$. Hence, it holds that

$$\mathbb{E}_{\mathbb{P}_0}[q_t(\boldsymbol{X})] = \mathbb{P}_0(X_t^2 \geq 1+\beta^*/s^*) = 2\bigl[1-\Phi\bigl(\sqrt{1+\beta^*/s^*}\bigr)\bigr]. \quad (7.76)$$

Under $\mathbb{P}_\mathcal{S}$ and $t \in \mathcal{S}$, we have $X_t^2 \sim (1+\beta^*/s^*)\chi_1^2$, since $X_t \sim N(0, 1+\beta^*/s^*)$. Therefore, for $t \in \mathcal{S}$ we have

$$\mathbb{E}_{\mathbb{P}_\mathcal{S}}[q_t(\boldsymbol{X})] = \mathbb{P}_\mathcal{S}(X_t^2 \geq 1+\beta^*/s^*) = 2[1-\Phi(1)]. \quad (7.77)$$

Meanwhile, for $t \in \mathcal{S}$ we have

$$\mathbb{E}_{\mathbb{P}_\mathcal{S}}[q_t(\boldsymbol{X})] - \mathbb{E}_{\mathbb{P}_0}[q_t(\boldsymbol{X})] = 2\big[\Phi\big(\sqrt{1+\beta^*/s^*}\big) - \Phi(1)\big] = \int_1^{\sqrt{1+\beta^*/s^*}} \sqrt{2/\pi} \cdot \exp(-x^2/2)\mathrm{d}x$$
$$\geq \sqrt{2/\pi} \cdot \big(\sqrt{1+\beta^*/s^*} - 1\big) \cdot \exp[-(1+\beta^*/s^*)/2] \geq \beta^*/(4\pi s^*). \tag{7.78}$$

Here the last inequality is from $\sqrt{1+\beta^*/s^*} - 1 \geq \beta^*/(2s^*)$ and $\beta^*/s^* = o(1)$. Note that for $\mathbb{P}_0$, by (7.76) we have

$$\sqrt{\mathbb{E}_{\mathbb{P}_0}[q_t(\boldsymbol{X})]\{1 - \mathbb{E}_{\mathbb{P}_0}[q_t(\boldsymbol{X})]\} \cdot \log(d/\xi)/n} \leq \sqrt{\log(d/\xi)/n} \leq \beta^*/(8\pi s^*),$$

since $\beta^{*2} n/[s^{*2} \log(d/\xi)] \to \infty$. For $\mathbb{P}_\mathcal{S}$, we have

$$\sqrt{\mathbb{E}_{\mathbb{P}_\mathcal{S}}[q_t(\boldsymbol{X})]\{1 - \mathbb{E}_{\mathbb{P}_\mathcal{S}}[q_t(\boldsymbol{X})]\} \cdot \log(d/\xi)/n} \leq \sqrt{\log(d/\xi)/n} \leq \beta^*/(8\pi s^*)$$

for the same reason. Then we have

$$\overline{R}(\phi) = \bar{\mathbb{P}}_0\big\{\sup_{t \in [T]} Z_{q_t} \geq 2\big[1 - \Phi\big(\sqrt{1+\beta^*/s^*}\big)\big] + \beta^*/(8\pi s^*)\big\}$$
$$+ \frac{1}{|\mathcal{C}|} \sum_{\mathcal{S} \in \mathcal{C}} \bar{\mathbb{P}}_\mathcal{S}\big\{\sup_{t \in [T]} Z_{q_t} < 2\big[1 - \Phi\big(\sqrt{1+\beta^*/s^*}\big)\big] + \beta^*/(8\pi s^*)\big\}$$
$$\leq \bar{\mathbb{P}}_0\big\{\sup_{t \in [T]} |Z_{q_t} - \mathbb{E}_{\mathbb{P}_0}[q_t(\boldsymbol{X})]| > \beta^*/(8\pi s^*)\big\}$$
$$+ \frac{1}{|\mathcal{C}|} \sum_{\mathcal{S} \in \mathcal{C}} \bar{\mathbb{P}}_\mathcal{S}\big\{\sup_{t \in [T]} |Z_{q_t} - \mathbb{E}_{\mathbb{P}_\mathcal{S}}[q_t(\boldsymbol{X})]| > \beta^*/(8\pi s^*)\big\} \leq 2\xi,$$

where the first and second inequalities are from (7.76)-(7.78) and Definition 3.1.

**Setting (ii):** For $\beta^{*2} n/[s^* \log d + \log(1/\xi)] \to \infty$, we consider the query functions and test in (6.5) and (6.6). For $\mathbb{P}_0$, we have $1/s^* \cdot \big(\sum_{j \in \mathcal{S}_t} X_j\big)^2 \sim \chi_1^2$, since $1/\sqrt{s^*} \cdot \sum_{j \in \mathcal{S}_t} X_j \sim N(0,1)$. Thus we have

$$\mathbb{E}_{\mathbb{P}_0}[q_t(\boldsymbol{X})] = \mathbb{P}_0\big[1/s^* \cdot \big(\textstyle\sum_{j \in \mathcal{S}_t} X_j\big)^2 \geq 1 + \beta^*\big] = 2\big[1 - \Phi\big(\sqrt{1+\beta^*}\big)\big]. \tag{7.79}$$

For $\mathbb{P}_\mathcal{S}$ with $\mathcal{S} = \mathcal{S}_t$, we have $1/s^* \cdot \big(\sum_{j \in \mathcal{S}_t} X_j\big)^2 \sim (1+\beta^*)\chi_1^2$, since $1/\sqrt{s^*} \cdot \sum_{j \in \mathcal{S}_t} X_j \sim N(0, 1+\beta^*)$. Therefore, for $\mathcal{S} = \mathcal{S}_t$ we have

$$\mathbb{E}_{\mathbb{P}_\mathcal{S}}[q_t(\boldsymbol{X})] = \mathbb{P}_\mathcal{S}\big[1/s^* \cdot \big(\textstyle\sum_{j \in \mathcal{S}} X_j\big)^2 \geq 1 + \beta^*\big] = 2[1 - \Phi(1)]. \tag{7.80}$$

Meanwhile, for $\mathcal{S} = \mathcal{S}_t$ we have

$$\mathbb{E}_{\mathbb{P}_\mathcal{S}}[q_t(\boldsymbol{X})] - \mathbb{E}_{\mathbb{P}_0}[q_t(\boldsymbol{X})] = \mathbb{P}_\mathcal{S}\big[1/s^* \cdot \big(\textstyle\sum_{j \in \mathcal{S}_t} X_j\big)^2 \geq 1 + \beta^*\big] - \mathbb{P}_0\big[1/s^* \cdot \big(\textstyle\sum_{j \in \mathcal{S}_t} X_j\big)^2 \geq 1 + \beta^*\big]$$
$$= 2\big[\Phi\big(\sqrt{1+\beta^*}\big) - \Phi(1)\big] = \int_1^{\sqrt{1+\beta^*}} \sqrt{2/\pi} \cdot \exp(-x^2/2)\mathrm{d}x$$
$$\geq \big(\sqrt{1+\beta^*} - 1\big) \cdot \sqrt{2/\pi} \cdot \exp[-(1+\beta^*)/2] \geq \beta^*/(4\pi), \tag{7.81}$$

where the last inequality holds because $\sqrt{1+\beta^*} - 1 \geq \beta^*/2$. Recall that $T = \binom{d}{s^*}$. Note that for $\mathbb{P}_0$, by (7.79) we have

$$\sqrt{\mathbb{E}_{\mathbb{P}_0}[q_t(\boldsymbol{X})]\{1 - \mathbb{E}_{\mathbb{P}_0}[q_t(\boldsymbol{X})]\} \cdot \log(T/\xi)/n} \leq \sqrt{[s^* \log d + \log(1/\xi)]/n} \leq \beta^*/(8\pi),$$

since $\beta^{*2}n/[s^*\log d + \log(1/\xi)] \to \infty$. Similarly, for $\mathbb{P}_\mathcal{S}$ we have

$$\sqrt{\mathbb{E}_{\mathbb{P}_\mathcal{S}}[q_t(\boldsymbol{X})]\{1-\mathbb{E}_{\mathbb{P}_\mathcal{S}}[q_t(\boldsymbol{X})]\} \cdot \log(T/\xi)/n} \leq \sqrt{[s^*\log d + \log(1/\xi)]/n} \leq \beta^*/(8\pi)$$

for the same reason. Then we have

$$\begin{aligned}
\overline{R}(\phi) &= \overline{\mathbb{P}}_0\big\{\sup_{t\in[T]}Z_{q_t} \geq 2\big[1-\Phi(\sqrt{1+\beta^*})\big] + \beta^*/(8\pi)\big\} \\
&\quad + \frac{1}{|\mathcal{C}|}\sum_{\mathcal{S}\in\mathcal{C}}\overline{\mathbb{P}}_\mathcal{S}\big\{\sup_{t\in[T]}Z_{q_t} < 2\big[1-\Phi(\sqrt{1+\beta^*})\big] + \beta^*/(8\pi)\big\} \\
&\leq \overline{\mathbb{P}}_0\big\{\sup_{t\in[T]}|Z_{q_t} - \mathbb{E}_{\mathbb{P}_0}[q_t(\boldsymbol{X})]| > \beta^*/(8\pi)\big\} \\
&\quad + \frac{1}{|\mathcal{C}|}\sum_{\mathcal{S}\in\mathcal{C}}\overline{\mathbb{P}}_\mathcal{S}\big\{\sup_{t\in[T]}|Z_{q_t} - \mathbb{E}_{\mathbb{P}_\mathcal{S}}[q_t(\boldsymbol{X})]| > \beta^*/(8\pi)\big\} \leq 2\xi,
\end{aligned}$$

where the first and second inequalities are from (7.79)-(7.81) and Definition 3.1. Hence, by combining settings (i) and (ii), we conclude the proof of Theorem 6.5. □

## 7.15 Auxiliary Result

In the sequel, we provide an auxiliary lemma on the weighted average of nonincreasing functions.

**Lemma 7.1.** Let $\{a_i\}_{i=0}^k$ and $\{b_i\}_{i=0}^k$ be two sequences satisfying $b_{i+1}/b_i = \kappa$ and $a_{i+1}/a_i \geq \kappa$ with $\kappa > 1$. For any nonincreasing function $h$, it holds that

$$\frac{\sum_{i=0}^k h(i)a_i}{\sum_{i=0}^k a_i} \leq \frac{\sum_{i=0}^k h(i)b_i}{\sum_{i=0}^k b_i}.$$

*Proof.* Let $\bar{a}_i = a_i/\sum_{i=0}^k a_i$ and $\bar{b}_i = b_i/\sum_{i=0}^k b_i$. Note that $\sum_{i=0}^k \bar{a}_i = 1$ and $\sum_{i=0}^k \bar{b}_i = 1$. We have

$$\frac{\sum_{i=0}^k h(i)a_i}{\sum_{i=0}^k a_i} = \sum_{i=0}^k h(i)\bar{a}_i, \quad \text{and} \quad \frac{\sum_{i=0}^k h(i)a_i}{\sum_{i=0}^k b_i} = \sum_{i=0}^k h(i)\bar{b}_i.$$

Without loss of generality, we assume $a_0 = b_0$, which implies $a_i \geq b_i$ for all $i \in \{0,\ldots,k\}$. Therefore, we have

$$\bar{a}_0 = \frac{a_0}{\sum_{i=0}^k a_i} \leq \frac{b_0}{\sum_{i=0}^k b_i} = \bar{b}_0.$$

Moreover, we have $a_i/a_{i-1} \geq b_i/b_{i-1}$, which immediately implies $\bar{a}_i/\bar{a}_{i-1} \geq \bar{b}_i/\bar{b}_{i-1}$. Hence, we have $\bar{a}_i/\bar{b}_i \geq \bar{a}_{i-1}/\bar{b}_{i-1}$. In other words, $\{\bar{a}_i/\bar{b}_i\}_{i=0}^k$ is nondecreasing. Because $\sum_{i=0}^k \bar{a}_i = 1$ and $\sum_{i=0}^k \bar{b}_i = 1$, there is an $\ell \in \{0,\ldots,k\}$ such that $\bar{a}_i \leq \bar{b}_i$ for $0 \leq i \leq \ell$ and $\bar{a}_i > \bar{b}_i$ for $\ell+1 \leq i \leq k$. Also we have

$$\sum_{i=0}^\ell (\bar{b}_i - \bar{a}_i) = \sum_{\ell+1}^k (\bar{a}_i - \bar{b}_i), \tag{7.82}$$

since $\sum_{i=0}^{k} \bar{a}_i = 1$ and $\sum_{i=0}^{k} \bar{b}_i = 1$. Therefore, we obtain

$$\frac{\sum_{i=0}^{k} h(i)a_i}{\sum_{i=0}^{k} a_i} - \frac{\sum_{i=0}^{k} h(i)b_i}{\sum_{i=0}^{k} b_i} = \sum_{i=0}^{k} h(i)(\bar{a}_i - \bar{b}_i) = \sum_{i=0}^{\ell} h(i)(\bar{a}_i - \bar{b}_i) + \sum_{i=\ell+1}^{k} h(i)(\bar{a}_i - \bar{b}_i)$$
$$\leq h(\ell)\sum_{i=0}^{\ell}(\bar{a}_i - \bar{b}_i) + h(\ell+1)\sum_{i=\ell+1}^{k}(\bar{a}_i - \bar{b}_i)$$
$$\leq -h(\ell)\sum_{i=\ell+1}^{k}(\bar{a}_i - \bar{b}_i) + h(\ell+1)\sum_{i=\ell+1}^{k}(\bar{a}_i - \bar{b}_i) \leq 0.$$

The second line is from the fact that $h$ is nonincreasing, and the last line is from (7.82). Hence, we complete the proof. □

## Acknowledgement

The authors sincerely thank Vitaly Feldman, Santosh Vempala, Will Perkins, Lev Reyzin, Hao Lu, and Jiacheng Zhang for helpful discussions and valuable suggestions.